\documentclass[acmsmall, table, xcdraw, numbers ,square, comma, sort&compress]{acmart}

\usepackage{amsmath,amsthm,amsfonts}
\usepackage{mathrsfs,mathtools}
\usepackage{psfrag,graphics,epsfig,graphicx}
\usepackage{adjustbox}
\usepackage{multirow}
\usepackage{rotating}
\usepackage{url}
\usepackage{setspace}
\usepackage{color}
\usepackage{array}
\usepackage{subcaption}
\usepackage{xcolor}
\usepackage{hhline}
\usepackage{ctable}
\usepackage{nccmath}
\usepackage{eucal}
\usepackage{hyperref}
\usepackage[english]{babel}
\usepackage[utf8]{inputenc}
\usepackage{algorithm}
\usepackage[noend]{algpseudocode}
\usepackage{resizegather}

\usepackage{bm}

\usepackage{enumitem}
\usepackage{textcomp}
\usepackage{chngpage} 

\let\oldReturn\Return
\renewcommand{\Return}{\State\oldReturn}
\renewcommand{\algorithmicrequire}{\textbf{Input: }}  
\renewcommand{\algorithmicensure}{\textbf{Output: }} 

\theoremstyle{plain}
\newtheorem{defn}{Definition}
\newtheorem{thm}{Theorem}
\newtheorem{lem}{Lemma}

\newtheorem{cor}{Corollary}

\newtheorem{fact}{Fact}

\theoremstyle{remark}
\newtheorem{rmk}{Remark}

\DeclareMathOperator*{\argmax}{arg\,max}

\DeclarePairedDelimiter\abs{\lvert}{\rvert}

\newcolumntype{L}[1]{>{\raggedright\let\newline\\\arraybackslash\hspace{0pt}}m{#1}}
\newcolumntype{C}[1]{>{\centering\let\newline\\\arraybackslash\hspace{0pt}}m{#1}}
\newcolumntype{R}[1]{>{\raggedleft\let\newline\\\arraybackslash\hspace{0pt}}m{#1}}
\newcolumntype{H}{>{\lrbox0}c<{\endlrbox}@{}} 

\renewcommand{\comment}[1]{}

\newcommand{\vars}{\texttt}
\newcommand{\quotes}[1]{``#1''}
\newcommand{\quotess}[1]{`#1'}
\newcommand{\simnot}{\mathord{\sim}}
\newcommand{\unaryminus}{\scalebox{0.75}[1.0]{\( - \)}}
\newcommand{\ntimes}{{\mkern-0mu\times\mkern-0mu}}
\makeatletter
\newcommand{\subalign}[1]{%
  \vcenter{%
    \Let@ \restore@math@cr \default@tag
    \baselineskip\fontdimen10 \scriptfont\tw@
    \advance\baselineskip\fontdimen12 \scriptfont\tw@
    \lineskip\thr@@\fontdimen8 \scriptfont\thr@@
    \lineskiplimit\lineskip
    \ialign{\hfil$\m@th\scriptstyle##$&$\m@th\scriptstyle{}##$\crcr
      #1\crcr
    }%
  }
}
\newcommand\footnoteref[1]{\protected@xdef\@thefnmark{\ref{#1}}\@footnotemark}
\ifcase \@ptsize \relax
  \newcommand{\miniscule}{\@setfontsize\miniscule{5}{6}}
\or
  \newcommand{\miniscule}{\@setfontsize\miniscule{5}{6}}
\or
  \newcommand{\miniscule}{\@setfontsize\miniscule{5}{6}}
\fi
\makeatother

\AtBeginDocument{%
  \providecommand\BibTeX{{%
    \normalfont B\kern-0.5em{\scshape i\kern-0.25em b}\kern-0.8em\TeX}}}
    
\begin{document}
\title{Achieving Transparency Report Privacy in Linear Time}

\author{Chien-Lun Chen}
\email{chienlun@usc.edu}
\author{Leana Golubchik}
\email{leana@usc.edu}
\affiliation{%
  \institution{University of Southern California}
}

\author{Ranjan Pal}
\email{palr@umich.edu}
\affiliation{%
  \institution{University of Michigan}
 }


\begin{abstract}
An accountable algorithmic transparency report (ATR) should \emph{ideally} investigate the (a) \emph{transparency} of the underlying algorithm, and (b) \emph{fairness} of the algorithmic decisions, and at the same time preserve data subjects' \emph{privacy}. However, a provably formal study of the impact to data subjects' privacy caused by the utility of releasing an ATR (that investigates transparency and fairness), is yet to be addressed in the literature. The far-fetched benefit of such a study lies in the methodical characterization of privacy-utility trade-offs for release of ATRs in public, and their consequential application-specific impact on the dimensions of society, politics, and economics.  
In this paper, we first investigate and demonstrate potential privacy hazards brought on by the deployment of transparency and fairness measures in released ATRs. \emph{To preserve data subjects' privacy, we then propose a linear-time optimal-privacy scheme}, built upon standard linear fractional programming (LFP) theory, for announcing ATRs, subject to constraints controlling the tolerance of privacy perturbation on the utility of transparency schemes. Subsequently, we quantify the privacy-utility trade-offs induced by our scheme, and analyze the impact of privacy perturbation on fairness measures in ATRs. To the best of our knowledge, this is the first analytical work that simultaneously addresses trade-offs between the triad of privacy, utility, and fairness, applicable to algorithmic transparency reports.

\end{abstract}

\comment{

}


\maketitle

\section{Introduction}\label{sec:intro}
In the era of big data and machine learning (ML), automated data processing algorithms are widely adopted in many fields for classification, prediction, or decision-making tasks due to huge volumes of input data and successful performance of ML approaches. 
Ongoing concerns and social uproar about the transparency and fairness of such decision-making have been raised by the media, government agencies, foundations, and academics over the past decade \cite{berk2017fairness, o2016weapons}. On a technical note, it has been shown in example studies that ML algorithms can be biased when (i) a dataset used to train ML models reflects society's historical biases \cite{torralba2011unbiased}, e.g., only a few female presidential nominees in the U.S. history, or (ii) because ML algorithms have much better understanding of the majority groups and poor understanding of the minority groups \cite{barocas2016big}, and so on. 
Thus, as we rapidly move forward to a data-driven age where a significant amount of day-day decision making in personal and professional spheres might be automation-driven, it would make great sense to often know the reasons behind certain decisions in order to understand if they are being treated fairly. Unfortunately, most decision processes today are often opaque, making it difficult to rationalize why certain decisions are made and whether they favor or disfavor certain individuals or groups.

Providing an algorithmic transparency report (ATR) by data controllers and third party regulatory agencies to decision-facing individuals is one way to investigate whether decisions made in a \emph{blackbox} are fair and transparent \cite{fink2018opening, diakopoulos2014algorithmic, rader2018explanations} - an immediate application area of considerable social impact being explainable AI for medical diagnoses  \cite{holzinger2017we, samek2019explainable} to enable medical personnel better understand and interpret diagnostic reports, and justify vulnerabilities of deployed AI models through domain expertise. 
This is a popular topic in research and there have been works in the last decade that have developed methodologies to reduce opaqueness in decision making \cite{guidotti2018survey,dovsilovic2018explainable} and improve on its fairness relative to certain protected attributes\footnote{Protected attributes form a subset of attributes, to which any decision process should not show preference, in any instance. It may contain public attributes (gender, race, etc.) and/or private/sensitive attributes (health conditions, gene, etc.).}
\cite{mehrabi2019survey}. The notion of transparency has also made its way into recently implemented policies for data protection such as in the EU General Data Protection Regulation (GDPR), and the California Consumer Privacy Act of 2018 (CCPA or AB-375) - both of which regulates the processing of collected personal or non-personal data of any data subject (the natural person to whom the data and the decision process relate) \cite{goodman2016european}. More specifically, any data controller shall inform data subjects before collecting their data, and is required to clearly explain the purpose of collecting data and how data will be processed, upon data subjects' requests (\quotes{\emph{right to explanation}} and \quotes{\emph{right to non-discrimination}}) \cite{goodman2016european}. 
\emph{However, a major side effect of providing transparency and fairness guarantees to the decision-facing clients is an unwanted risk to the privacy of other clients in a database.} To this end, there exists a substantial literature pointing out potential privacy threats in ML \cite{papernot2016towards}, including membership attacks \cite{shokri2016membership}, training data extraction (model inversion attack) \cite{fredrikson2014privacy, fredrikson2015model}, model extraction \cite{tramer2016stealing}, and so on. However, for ATRs, although it has been pointed out that transparency, proposed by legislature to protect people's rights, may hurt privacy \cite{dat16, ananny2018seeing}, \emph{it is yet to be made methodically clear how transparency can hurt privacy.} 

\begin{figure}[!t]
\centering
\includegraphics[width=3in]{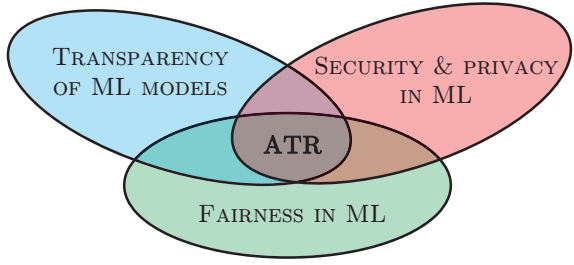}
\caption{A depiction of the realm of accountable ATRs}
\label{fig:realm}
\vspace*{-3mm}
\end{figure}

\textbf{Goal} - An \emph{accountable} ATR, especially for automated ML decision processes, should ideally include \emph{transparency} of the underlying algorithm, ability to inspect \emph{fairness} of the algorithmic decisions, and most importantly, preserve data subjects' \emph{privacy} (\quotes{\emph{right to privacy}} \cite{cate1994right}), as depicted in Fig. \ref{fig:realm}. \emph{Our goal in this paper is to work towards this goal and study the corresponding trade-offs between the triad elements.} 


%
%
%
In this paper, we investigate this problem and explicitly show that data subjects' private information can be inferred via various transparency schemes and fairness measures in announced ATRs. 

%
%

\textbf{Research Contributions} -  We make the following contributions in this research paper:
\begin{itemize}
\item We explicitly demonstrate inference attacks on data subjects' private information using a synthetic and a real dataset and show that such attacks can be performed on various transparency schemes \emph{without strong assumptions} of adversaries' knowledge. These instances expose the possible aspects of algorithmic transparency that could hurt data subjects' privacy and subsequently have negative socio-political implications. (See Section \ref{sec:problem} and Appendix \ref{sec:privacy_leak_method_2})
\item To protect data subjects' privacy in an ATR, we propose a privacy-aware mechanism perturbing the unveiled rationale of opaque decision-making algorithms to control the amount of disclosed information in ATRs, at the same time providing sufficient utility. Specifically, given a released privacy-preserving ATR, for \emph{honest-but-curious} adversaries which may know (i) the target individuals' public information used as inputs to the decision model, (ii) the target individuals' received decisions (model outputs), and (iii) side-information from auxiliary sources, the maximum confidence of inferring any sensitive information about any data subject is guaranteed not exceeding a predetermined privacy threshold (See Section \ref{sec:preliminaries} and \ref{sec:privacy_fidelity}). 
\item We study the trade-off between privacy and utility of ATRs. Specifically, we aim to understand the minimum required perturbation/distortion in order to provide a certain privacy guarantee, or the maximum privacy that can be provided/guaranteed subject to utility constraints, which can be formulated as an optimization problem for privacy-utility trade-off in ATRs. In addition, we analyze the impact of privacy perturbations on fairness measures. In this regard, our work provides useful quantitative trade-offs and influences between privacy, transparency, and fairness measures in ATRs (See Sections \ref{sec:privacy_fidelity} and \ref{sec:optimization}).
\item We deduce that our privacy-utility optimization problem equivalent to a \emph{generalized linear fractional programming problem} (LFP) \cite{zionts1968programming, boyd2004convex}. Such a problem can in general be solved as a sequence of linear programming feasibility problems, each with pseudo-polynomial time complexity with respect to the number of problem variables (the number of different \emph{decision regions}\footnote{The regions of input attributes partitioned by decision rules; see Section \ref{sec:problem} for examples.} in our problem), which, however, in the worst case, grows exponentially with the number of input attributes - hence lending the said optimization problem intractable for large record sizes. 
\emph{However, on a closer investigation}, we figure out that the region of interest in the solution space can be decomposed into disjoint \quotes{subspaces} leading to multiple independent sub-problems - each bearing important properties and amenable to propose \emph{closed-form} solutions. Subject to utility constraints, the optimal-privacy protection scheme can thus be solved from the optimization problem efficiently in \emph{linear time} (See Sections \ref{sec:optimal_solution} and \ref{sec:results}).
\end{itemize}

\section{Applicability and Ethicality of the Proposed Privacy Scheme}\label{sec:app}
In this section, we first describe a range of possible application domains for the proposed ATRs privacy protection scheme. We then discuss potential ethical concerns related to announcing a perturbed \quotes{transparency} report, specifically, the conflict between principles of \emph{transparency} and \emph{perturbation} for preserving data subjects' privacy.

\subsection{Applicability} 
The proposed privacy protection scheme, based on perturbing of decision mappings (Definition \ref{def:decision_mapping}), can be applied to numerous applications with characteristic that \emph{the decision regions of the application decision process are disjoint finite sets of the input attribute space, i.e., the application decision process can be represented by a finite number of decision mappings}\footnote{Our scheme may not be a good fit for some application domains where it may not be possible to represent the decision process using a finite number of decision mappings, e.g., applications of natural language processing, such as speech/music recognition, speech/text understanding, and text/intent classification, or applications of image and video processing, such as text recognition, item detection and alert, and image classification.}. Although, to our knowledge, to date there are no current instantiations of an \emph{algorithmic} transparency report, we believe that potential applicable domains for our scheme include the following (among others):
\begin{itemize}
    \item \emph{University admissions and job recruitment}: Fairness in university/college admissions has become a significant public concern, attracting more and more attention from the public as well as the media \cite{app:UA3}, even though the fact that the definition of \quotes{fairness} is still controversial, e.g., whether race should be used in admissions decisions to reflect racial diversity \cite{app:UA2, app:UA1}. To respond to the public's concerns, some governments \cite{app:UA4} and universities \cite{app:UA5} initiated work on \emph{admissions transparency}, providing statistical data from applications (inputs) and admissions (outputs). Currently, we are not aware of instantiations disclosing the admission decision process; however, a negligent report could leak applications' data or records, e.g., (range of) SAT scores, competition records and ranks, extracurricular activities, or volunteer work. Similar circumstances apply to applications and corresponding decisions in job recruitment and other related domains.
    \item \emph{Credit scores and the associated domains}: When it comes to evaluating and identifying everyone's financial creditworthiness based on credit scores, people have the following concerns: are their credit scores computed/treated fairly \cite{app:CR1} and whether they have the ability to identify and contest any (potentially) unfair credit decisions \cite{app:CR3}. Similar circumstances apply to credit card applications \cite{app:CR2} and a variety of loans, i.e., domains where credit scores may be taken into account. In this paper, we demonstrate potential privacy hazards that could be brought on by a bluntly disclosed ATR in credit card applications via various types of transparency schemes and fairness measures using a synthetic (Section \ref{sec:problem}) and a real dataset (Appendix \ref{sec:privacy_leak_method_2}). 
    \item \emph{Medical or pharmocogenetic models}: As noted in \cite{kamali2010pharmacogenetics}, a pharmocogenetic model has been built to predict proper dosages for patients based on their clinical histories, demographics, and genotypes. However, it has been shown in \cite{fredrikson2014privacy} that once accurate information about the model is leaked (or obtained through hacking), it can be utilized by an attacker to identify patients' genotypes, which could be exploited to further infer other private information, e.g., risk of getting a particular disease or someone's family ancestry. In the ATR setting, we focus on a related scenario where an adversary has no ability to access the pharmocogenetic model internals but can merely gain model information from an announced ATR. In such a case, our proposed privacy protection scheme can be applied to preserve patients' privacy and their genotype information.
    \item \emph{Open Government}: It has been reported in \cite{fink2018opening} that nowadays governments utilize algorithms to detect or to determine a variety of issues, such as illegal insider trading, eligibility for public health benefits, and tax evasion. In this regard, the Open Government organization \cite{app:OG} aims to bring transparency to the data and the algorithms used by governments, aiming for people and society to supervise governments' actions and decisions. However, opening a government blackbox can be very dangerous and can bring catastrophic results to society if the released information is not carefully treated, and hence it is crucial to have a provable privacy-preserving scheme for any planned-to-disclose information to protect people's privacy and secret information (tax data, health/medical records, banking information, business processes, and so on).
    \item \emph{Online advertising}: ML algorithms can be biased \cite{torralba2011unbiased, barocas2016big}, and it has been shown that bias also appears in online advertising, one of the ML applications that we probably experience daily. In \cite{shekhawat2019algorithmic} authors indicate a bias and a privacy issue associated with online advertising settings, particularly when users select the "Rather Not Say" category of gender. In addition, \cite{datta2015automated} also found that setting the gender to female results in receiving fewer instances of advertisements related to high paying jobs as compared to setting it to male. With concerns about ML bias and the purpose of the collected data, GDPR and CCPA stipulate rights to explanation and non-discrimination for data subjects; ad providers are required to respond to data subjects' requests regarding what data has been collected, how data is used, and if the applied ML models treat them fairly. Thus, all the disclosed information in an ATR may need to be further processed to protect data subjects' privacy.
\end{itemize}

\subsection{Ethicality}
When our proposed privacy protection scheme is applied to an ATR, the announced information regarding the opaque decision process may be more or less distorted, and the announced measured fairness/bias may also deviate from the true one. This may raise concerns about the \emph{ethicality} (manner of being ethical) of the process, i.e., whether the perturbed information could mislead the public into trusting or believing that a biased decision process is fair, and vice versa. 

Similarly to privacy preservation in data-mining (PPDM) and data-publishing (PPDP), a common theme is to find an optimal trade-off between utility and privacy, subject to a certain degree of privacy guarantee for data subjects. In the context of ATRs - although both transparency and privacy are major principles in data ethics \cite{stahl2018ethics, dataethics} - we believe that data subjects' privacy should have higher priority \cite{allen2016protecting}. Similarly to PPDM or PPDP, an auxiliary note could be appended with the announced information indicating that some listed information might be anonymized or perturbed for data subjects' privacy, which could help the public understand how to interpret the disclosed information appropriately. Moreover, in light of this, in this work, we propose a fidelity measure (Section \ref{subsec:fidelity}) for the announced decision mappings and characterize the influence of privacy perturbation on the measured fairness (Section \ref{subsec:fidelity_fairness}). This information can also be disclosed with the announced ATR in order to further assist the information recipients in understanding the range of true measures.

\section{Demonstrating Privacy Leakage via an ATR}\label{sec:problem}
As a necessary and important step, we first motivate our research by comprehensively demonstrating via an example consumer database of how a data subject's (i.e., consumer's) private information can be leaked via an announced algorithmic transparency report (ATR). \emph{In this work we only focus on reports that provide a rationale on the use of ML models to process individual records.} As section structure, we start by briefly reviewing transparency approaches on which privacy leakages can be induced, and follow it up with a specific example of privacy leakage on each transparency approach. 

\subsection{Algorithmic Transparency Report (ATR) in a Nutshell} \label{subsec:transparency}
ML models used to make decision on consumer individuals are often opaque to the latter, and act as blackboxes.  
A survey of popularity-gaining transparency schemes to explain ML blackboxes is provided in \cite{guidotti2018survey}. 
A common representative (from the survey) transparency approach collects both input data and labeled outputs (decision outcomes) as a training dataset, to train an ML \emph{surrogate model} (e.g., linear model, logistic regression, decision tree, decision rules) to mimic the behavior of the blackbox. Popular methods include Anchors \cite{ribeiro2018anchors} and PALM \cite{krishnan2017palm}. The output of such learned behavior must be interpret-able (understandable) by humans. Another common approach (e.g., \cite{breiman2001random, fisher2018model}) extracts certain important ``properties'' from blackbox models, such as contributions of input features, to model outputs. Specifically, these transparency schemes measure feature importance (based on the underlying $D$), using both amplitude and sign to represent importance/influence of input features, where larger amplitude represents greater influence, and the sign indicates positive or negative effect on the output. Popular methods include LIME \cite{ribeiro2016should}, FIRM \cite{zien2009feature}, QII \cite{dat16}, and Shapley Value \cite{kononenko2010efficient}, PDP \cite{friedman2001greedy}, ICE \cite{goldstein2015peeking}, and ALEPlot \cite{apley2016visualizing}.
In addition to transparency schemes, an ATR may also provide information regarding whether a decision algorithm or ML model is biased against a certain group or individual - in other words, an ATR may measure individual or group \emph{fairness} of ML-based decisions based on the different desired metrics discussed in existing literature \cite{corbett2017algorithmic,dwo11,feldman2015certifying,zemel2013learning,kamishima2012fairness,kamiran2013quantifying,biddle2006adverse}. 
We refer readers to Appendix \ref{sec:fairness} for detailed definitions of various individual and group fairness measures. 
\emph{In what follows, we investigate and demonstrate privacy leakage instances via various kinds~of transparency schemes and fairness measures, given \emph{honest-but-curious} adversaries.}

\subsection{Privacy Leakage via Interpretable Surrogate Models} \label{subsec:privacy_surrogate}

As noted, transparency schemes 
can interpret a blackbox's rules in a human-understandable manner, such as decision rules or decision trees. Here, we explain how such transparent information can hurt a data subject's privacy. Without loss of representativity, here we set up a synthetic scenario, in which we consider the existence of a perfect interpretable surrogate model\footnote{
We consider the most privacy-catastrophic case, a perfect interpretation, which has the most accurate information in an ATR.}, to illustrate the possibility of causing a catastrophic privacy leak.

Consider the following synthetic credit card application scenario (summarized in Table \ref{table:scenario}). A credit card application takes several input attributes from applicants, while the bank's decision process only depends on two input attributes: the applicants' annual income and their gender (which, depending on the country, may be illegal and in those cases should not be used in any decision process). Due to the suspicious differences in approval rates between male and female applicants, a third-party regulatory agency actively takes action. It collects all applicants' data and their received decisions, and trains an (assumed perfect) interpretable surrogate model, disclosing the decision rules used in the credit card application to all past applicants, as follows
\begin{equation*}
\label{eq:credit_decision_rules}
\begin{split}
& d({\text{\{\textit{Income}\}}} > 200{\text{k}}) = 1 \text{,} \\
& d({\text{\{\textit{Income}\}}}\in100{\text{k}}\simnot200{\text{k, Male}}) = 0.5 \text{,} \\
\end{split}
\end{equation*} 
where $d(\cdot)$ is \emph{decision rule} representing the probability of receiving a positive decision given the condition. An equivalent \emph{if-then} decision rule form is the following
\begin{equation*}
\label{eq:credit_ifelse_rules}
\begin{cases}
\text{ if \textit{Income $> 200$k}, then \textit{Positive Decision}; } \\
\text{ if \textit{$100$k $\le$ Income $\le$ $200$k} $\wedge$ \textit{Male}, then \textit{Random}; } \\
\text{ otherwise, then \textit{Negative Decision}. } \\
\end{cases}
\end{equation*} 
Note that other interpretable surrogate models such as a decision tree or logistic regression can also be equivalently expressed by decision rule $d(\cdot)$.

\begin{table}[tbp]
\centering
\caption{A Synthetic Credit Card Application Scenario}
\vspace*{-3mm}
\label{table:scenario}
\begin{tabular}{|>{\small}c||>{\small}c|>{\small}c||>{\small}c||>{\small}c|}
\hline
& & \multicolumn{3}{>{\small}c|}{Adversaries' Knowledge} \\                                                                                             \hline
& \multicolumn{2}{>{\small}c||}{Input Attributes} & ATR & Side-Info \\                                                                                            
\specialrule{.1em}{.05em}{.05em} 
\begin{tabular}[c]{@{}c@{}}Popu-\\lation\end{tabular} & \begin{tabular}[c]{@{}c@{}}Annual\\Income\end{tabular} & Gender & \begin{tabular}[c]{@{}c@{}}Decision\\Rule\end{tabular} & \begin{tabular}[c]{@{}c@{}}Census\\Statistics\end{tabular} \\ \hline
$139$   & $<100$k                & F      & $0$     &  93.1\%      \\ \hline                                              
$9$     & $100$k$\simnot200$k    & F      & $0$     & { 5.7\%}     \\ \hline
$2$     & $>200$k                & F      & $1$     & { 1.2\%}     \\ \hline                                              
$117$   & $<100$k                & M      & $0$     &  84.2\%      \\ \hline                                              
$18$    & $100$k$\simnot200$k    & M      & $0.5$   &  12.3\%      \\ \hline
$5$     & $>200$k                & M      & $1$     & { 3.5\%}     \\ \hline                                              
\end{tabular}
\end{table}

Next, we demonstrate how data subjects' sensitive information (annual income in this scenario) could be leaked. Revisit Table \ref{table:scenario} in which the key input attributes, population, and decision rule of the credit card application are listed. Population of applicants are aggregated according to \emph{decision regions}, i.e., the regions of input attributes partitioned by decision rules. Here the population proportion among decision regions refers to the U.S. census data, and adversaries assumed blind to population of applicants utilize the U.S. census data as side-information to estimate, for each decision region, the percentage of the total number of male/female applicants (listed in the \quotes{Census Statistics} attribute; for instance, the value 93.1\% in Table \ref{table:scenario} represents the following: given that the decision region is \{Annual Income < 100k; Gender=Female\}, 93.1\% of female applicants belong to this region). Adversaries know public information of targeted applicants and also know decision rules from an announced ATR.

When an ATR containing such decision rule is negligently announced, as it reveals strong depen-\newline dencies between annual income and decisions, any female using such a credit card in public instantly tells anyone who has ever seen the report that her annual income is above 200k, which not only results in a privacy hazard to her, but may also result in unexpected safety concerns. In such a case, an adversary does not even require auxiliary information to be able to infer someone's secret.

Male credit card owners are also at risk, although not as much. For a male credit card owner, the confidence of an adversary believing that his income is above 200k is only around 36\%, compared with 100\% in the case  of a female owner, while based on census statistics, the confidence of an adversary believing that his income is above 200k is merely 3.5\%. In other words, once such a negligent algorithmic transparency report is announced to the public, a high-income (\textgreater200k) male credit card owner's risk of exposing annual income information is increased 10 fold.

In summary, releasing precise information of interpretable surrogate models (that can be equivalently expressed by decision rules) can be harmful to data subjects' privacy, as such information gives adversaries a clear mapping
between input records and received decision. With assistance from public information and/or side-information, adversaries can abuse algorithmic transparency to undermine people's privacy.
The same privacy leakage concern applies when precise information of transparency scheme is released in the form of feature importance/interaction, which, however, in the interest of space, is explicitly demonstrated in Appendix \ref{sec:privacy_leak_method_2}, using a real dataset.

\subsection{Privacy Leakage via Fairness Measures} \label{subsec:privacy_fairness}
Recall that one of the main motivation for algorithmic transparency is to understand if a decision-making algorithm is fair and complies with regulations/law, e.g., the U.S. equal employment opportunity commission (EEOC) regulates the ratio of the hiring rates between women and men, which should not be lower than 80\% (80\%-rule). In an algorithmic transparency report, such fairness measures may be required upon data subjects' demands (e.g., GDPR, Article 22). 

To this end, consider again the credit card application in Table \ref{table:scenario}, in which the bank is under suspicion of discriminating against female applicants. Upon female applicants' demands, a regulation agency gets involved and discloses the following fairness measures for gender: (i) bias in statistical parity (SP) (Definition \ref{def:stat_parity}) for male and female applicants, (ii) bias in conditional statistical parity (CSP) (Definition \ref{def:condi_stat_parity}) for male and female applicants who have the same level of income. An ATR listing all the above fairness measures w.r.t. the credit card application is shown in Table \ref{table:fairness_measure} (see Remark \ref{rmk:Table_2} for details), which can be announced to the public in an electronic form, e.g., through a website (e.g., GDPR, Recital 58, information related to the public's concerns).

\begin{table}[tbp]
\centering
\caption{Fairness Measures for Table \ref{table:scenario} in an ATR}
\vspace*{-3mm}
\label{table:fairness_measure}
\begin{tabular}{>{\small}l} 
${{\CMcal{Y}}_1} = \{ \text{F} \}$, ${{\CMcal{Y}}_2} = \{ \text{M} \}$ \\
${{\CMcal{W}}_1} = \{ \text{Annual Income} \le \text{100k} \}$ \\
${{\CMcal{W}}_2} = \{ \text{100k} \le \text{Annual Income} \le \text{200k} \}$ \\
${{\CMcal{W}}_3} = \{ \text{Annual Income} \ge \text{200k} \}$ \\ \hline
Overall approval rate for female (${{\CMcal{Y}}_1}$) = 1.33\%; \\
Overall approval rate for male (${{\CMcal{Y}}_2}$) = 10\%; \\
Bias in SP for ${{\CMcal{Y}}_1}$ and ${{\CMcal{Y}}_2}$ = 0.0866; \\
Bias in CSP for $\{ {{\CMcal{Y}}_1},{{\CMcal{W}}_1} \}$ and $\{ {{\CMcal{Y}}_2},{{\CMcal{W}}_1} \}$ = 0;\\
Bias in CSP for $\{ {{\CMcal{Y}}_1},{{\CMcal{W}}_2} \}$ and $\{ {{\CMcal{Y}}_2},{{\CMcal{W}}_2} \}$ = 0.5;\\
Bias in CSP for $\{ {{\CMcal{Y}}_1},{{\CMcal{W}}_3} \}$ and $\{ {{\CMcal{Y}}_2},{{\CMcal{W}}_3} \}$ = 0.
\end{tabular}
\end{table}

Moreover, a data subject, which is a credit card applicant in our scenario, has the right to inquire about the decision principle w.r.t. his or her personal data. Mary, a low-income (\textless100k) female who would like to know why her applications are always denied, demands information regarding the decision processing for her record. The response indicates that the approval rate for a low-income female is 0. 
If we let $d_{i,j}$ be the decision rule for people in $\{ {{\CMcal{Y}}_i},{{\CMcal{W}}_j} \}$ in Table \ref{table:fairness_measure}, by utilizing the census statistics as shown in Table \ref{table:scenario}, and based on the definitions of SP and CSP for binary decisions in \eqref{eq:demo_parity} and \eqref{eq:condi_stat_parity_2}, respectively, the information provided in Table \ref{table:fairness_measure} tells us the following:

\begin{flalign}
&\text{Overall approval rate for female} ({{\CMcal{Y}}_1}) = 0.0133 \approx 0.931 d_{1,1} + 0.057 d_{1,2} + 0.012 d_{1,3} \label{eq:fh_ineq_1} \\
&\text{Overall approval rate for male} ({{\CMcal{Y}}_2}) = 0.1 \approx 0.842 d_{2,1} + 0.123 d_{2,2} + 0.035 d_{2,3} \label{eq:fh_ineq_2} \\
&\text{Bias in SP for } \{ {{\CMcal{Y}}_1},{{\CMcal{W}}_1} \} \text{ and } \{ {{\CMcal{Y}}_2},{{\CMcal{W}}_1} \} = 0 = \abs{d_{1,1} - d_{2,1}} \label{eq:fh_ineq_3} \\
&\text{Bias in SP for } \{ {{\CMcal{Y}}_1},{{\CMcal{W}}_2} \} \text{ and } \{ {{\CMcal{Y}}_2},{{\CMcal{W}}_2} \} = 0.5 = \abs{d_{1,2} - d_{2,2}} \label{eq:fh_ineq_4} \\
&\text{Bias in SP for } \{ {{\CMcal{Y}}_1},{{\CMcal{W}}_3} \} \text{ and } \{ {{\CMcal{Y}}_2},{{\CMcal{W}}_3} \} = 0 = \abs{d_{1,3} - d_{2,3}}. \label{eq:fh_ineq_5}
\end{flalign} 


Since Mary just got a reply indicating $d_{1,1} = 0$, from \eqref{eq:fh_ineq_3} and \eqref{eq:fh_ineq_5}, Mary then knows that $d_{1,1} = d_{2,1} = 0$, $d_{1,3} = d_{2,3}$, and from \eqref{eq:fh_ineq_4}, either $d_{1,2} = d_{2,2} + 0.5$ or $d_{1,2} = d_{2,2} - 0.5$. She can first assume $d_{1,2} = d_{2,2} + 0.5$, by plugging the values of $d_{1,1}$ into \eqref{eq:fh_ineq_1} and $d_{2,1}$ into \eqref{eq:fh_ineq_2}, and replacing $d_{2,2}$ and $d_{2,3}$ by $d_{1,2}-0.5$ and $d_{1,3}$ in \eqref{eq:fh_ineq_1} and \eqref{eq:fh_ineq_2}, respectively, she gets $0.057 d_{1,2} + 0.012 d_{1,3} = 0.0133$ from \eqref{eq:fh_ineq_1} and $0.123 d_{1,2} + 0.035 d_{1,3} = 0.1615$ from \eqref{eq:fh_ineq_2}. Since $d_{i,j}$ are probabilities, $\forall i,j$, $d_{1,2}$ and $d_{1,3}$ can not be grater than 1, and thus the obtained equation from \eqref{eq:fh_ineq_2} is infeasible, which implies the assumption is wrong. She then knows $d_{1,2} = d_{2,2} - 0.5$. Repeat the same steps and she will obtain $d_{1,2} = 0.0088$ and $d_{1,3} = 1.0692$. By understanding that any $d_{i,j}$ can not be greater than 1 and this is probably caused by the mismatch between the census statistics and the true distribution, she would thus update $d_{1,3} = 1$ and thus obtain $d_{1,2} = 0.0013 \approx 0$; these estimates are very close to the true the values. In addition, Mary can use the obtained $d_{1,2}$ and $d_{1,3}$ to further acquire $d_{2,2}$ and $d_{2,3}$. Therefore, by utilizing the decision processing rule for her record and the publicly announced fairness measures, she can obtain accurate decision rules for the credit card application. As in Section \ref{subsec:privacy_surrogate}, we know that a privacy disaster can happen when accurate decision rules are released or hacked. The adversary Mary now can utilize her hacked decision rules to infer other applicants' income.

From the above demonstrations, we have seen that a negligent ATR can result in a serious hazard to data subjects’ privacy. 
In the following sections, we formalize the privacy leakage problem, and propose the corresponding properties and solutions. \emph{We will revisit the examples demonstrated above again in Section \ref{sec:results}, with our proposed solutions applied.}

\begin{rmk}
\label{rmk:Table_2}
Here we demonstrate how the numbers in Table \ref{table:fairness_measure} are calculated based on Table \ref{table:scenario}. The definitions of SP and CSP for binary decisions can be found in \eqref{eq:demo_parity} and \eqref{eq:condi_stat_parity_2}, respectively.\\
\phantom{xxxx} Overall approval rate for female (${{\CMcal{Y}}_1}$) $= (2 \times 1)/(139+9+2)= 1.33\%$; \\
\phantom{xxxx} Overall approval rate for male (${{\CMcal{Y}}_2}$) $= (18 \times 0.5 + 5 \times 1)/(117+18+5) = 10\%$; \\
\phantom{xxxx} Bias in SP for ${{\CMcal{Y}}_1}$ and ${{\CMcal{Y}}_2} = \abs{1.33\% - 10\%} = 0.0866$; \\
\phantom{xxxx} Bias in CSP for $\{ {{\CMcal{Y}}_1},{{\CMcal{W}}_1} \}$ and $\{ {{\CMcal{Y}}_2},{{\CMcal{W}}_1} \} = \abs{0 - 0} = 0$;\\
\phantom{xxxx} Bias in CSP for $\{ {{\CMcal{Y}}_1},{{\CMcal{W}}_2} \}$ and $\{ {{\CMcal{Y}}_2},{{\CMcal{W}}_2} \} = \abs{0 - 0.5} = 0.5$;\\
\phantom{xxxx} Bias in CSP for $\{ {{\CMcal{Y}}_1},{{\CMcal{W}}_3} \}$ and $\{ {{\CMcal{Y}}_2},{{\CMcal{W}}_3} \} = \abs{1 - 1} = 0$.
\end{rmk}


\section{Problem Setup}\label{sec:preliminaries}
In the following sections, we formalize and analyze the privacy leakage problem in ATR.
To begin with, in this section, we provide essential notations listed in Table \ref{table:notation} and useful definitions for problem setup, followed by adversarial settings and definition of privacy violation in ATRs formally.

\subsection{Decision Mapping} \label{subsec:decision_mapping}
\begin{figure}[!t]
\centering
\includegraphics[width=3.5in]{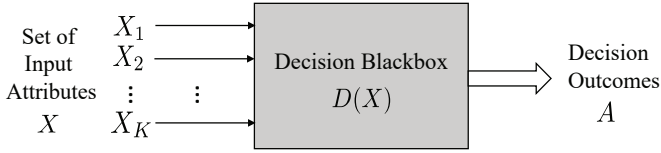}
\caption{A representative illustration of a decision blackbox}
\label{fig:blackbox}
\vspace*{-3mm}
\end{figure}

Fig. \ref{fig:blackbox} illustrates an opaque decision-making blackbox, which is essentially an unknown \textit{decision mapping} function defined as follows. 

\begin{defn}
\label{def:decision_mapping}
(\textit{Decision Mapping} \cite{dwo11}) 
Consider a decision process as illustrated in Fig. \ref{fig:blackbox}, where $X = \{ {X_k} \mid k = 1, \ldots ,K\}$ is a set of input attributes, $A$ the output attribute (decision outcomes), and ${\CMcal{A}}$ the range of $A$. 
Recall that $\Delta(S)$ is a set of probability distributions over $S$.
A decision mapping ${D_{\CMcal{A}}}: {{{\CMcal{R}}_{X}}} \to \Delta({\CMcal{A}})$ is a function mapping from  the range of input attributes to a set of probability distributions over the range of decision outcomes. Formally, 
\begin{equation}
\label{eq:decision_mapping}
\begin{split}
{D_{\CMcal{A}}}(X) = \{ {P_{A|X}}(A = a|X) \mid \forall a \in {\CMcal{A}} \} 
= \{ {{D_{a}}(X)} \mid \forall a \in {\CMcal{A}} \}.
\end{split}
\end{equation}
Particularly, for binary decisions ($0=$\quotess{\text{negative}} and $1=$\quotess{\text{positive}}), we let
\begin{equation}
\label{eq:decision_mapping_binary}
{D_{\CMcal{A}}}(X) = 
\begin{cases}
{D_{1}}(X) = {d(X)} & \text{, for $a=1$} \\
{D_{0}}(X) = {1-d(X)} & \text{, for $a=0$}, 
\end{cases} 
\end{equation}
where $d(X)$ is decision rule \cite{corbett2017algorithmic} representing probabilities of mapping from input space to the positive decision outcome.
\end{defn}
Clearly, decision mapping is more comprehensive, while decision rule is more concise and convenient for an ATR, e.g., decision rule in Table \ref{table:scenario}.

As noted in Section \ref{subsec:transparency}, an ATR opens an opaque decision blackbox via transparency schemes such as an interpretable surrogate model (a surrogate of ${D_{\CMcal{A}}}$) or feature importance/interaction (a function of ${D_{\CMcal{A}}}$). 
In addition, an ATR may also contain fairness measures (functions of ${D_{\CMcal{A}}}$, see Appendix \ref{sec:fairness}). 
Clearly, an ATR is in general a function of decision mapping ${D_{\CMcal{A}}}$ (when there is no confusion, we omit the subscript and simply write $D$ in the rest of the paper for conciseness); while released, the mapping from decision inputs to outputs are made public, and thus it is very crucial to ensure the reverse inference is not possible, or limited with low confidence.
To explicitly characterize the reverse inference, we first need to understand the capability of the adversaries.

is a legitimate participant in a communication protocol who will not deviate from the defined protocol but will attempt to learn all possible information from legitimately received messages.

\subsection{Adversarial Settings} \label{subsec:adversary}
For the privacy leakage problem brought on by releasing ATRs, we consider \emph{honest-but-curious} (or \emph{curious-but-not-malicious}) adversaries, i.e., 
adversaries who only perform legitimate actions and will not deviate from the defined protocol but would like to learn as much as possible (including others' secrets); in our ATR setting, this implies that an adversary will not hack into the system and steal information but only acquires as much as possible information that is made public or is widely-available. 
For example, the adversaries may know public information about his friends, e.g., gender, race, ZIP code, age, etc; the adversaries may also have knowledge about the census data \cite{uscensus} providing side-information (with \emph{week inference}) between public and private attributes, e.g., joint distributions between age, race, marriage status, household size, and income. 
Such kind of adversaries are ubiquitous, making privacy leakage via released ATRs omnipresent.

It's worth noting that 
the adversaries do not have access to \emph{all} the input features and \emph{all} the output responses (decision outcomes), and thus are not able to extract any information about the blackbox from the limited knowledge. Instead, the adversaries may just know the public information and the received decisions of the targeted several individuals. 
The reason that we particularly focus on honest-but-curious adversaries in this paper is that we would like to convey an important message that \emph{candidly releasing an ATR could result in privacy hazards even for weak adversaries who are not able to probe or hack into the system but possess some public information and/or widely-available side-information}. Moreover, more powerful adversaries who may have access to all input features and output responses can train a more powerful surrogate model (as it need not be an interpretable model) to mimic the original model, and thus can obtain more accurate information w.r.t. the decision blackbox, as compared to what is provided in an announced ATR. In such a case, the privacy hazard is not due to the ATR, as adversaries have already obtained something more powerful (resulting in stronger inference), and thus such an adversarial setting is not meaningful for ATR.


\begin{table}[t!]
\centering
\caption{Notation}
\vspace*{-3mm}
\label{table:notation}
\setlength\tabcolsep{10pt} 
\begin{tabular}{|>{\small}c|>{\small}l|}  \hline
$\CMcal{U}$ & Set of all public attributes \\  \hline
$\CMcal{S}$ & Set of all private attributes \\  \hline
${X_{k}}$ & Random variable (r.v.) of attribute $k$ \\  \hline
${X_{\CMcal{U}}}$ & $ = \{ {X_{k}} \mid \forall k \in \CMcal{U} \}$; collection of r.v.'s of all public attributes \\  \hline
${X_{\CMcal{S}}}$ & $ = \{ {X_{k}} \mid \forall k \in \CMcal{S} \}$; collection of r.v.'s of all private attributes \\  \hline
$X$ & $= ( {X_{\CMcal{U}}}, {X_{\CMcal{S}}} )$; collection of r.v.'s of all attributes \\  \hline
${{\CMcal{R}}_{X}}$ & Range of $X$; the universe of inputs; ${{\CMcal{R}}_{X}} = {{\CMcal{R}}_{X_{\CMcal{U}}}} \times {{\CMcal{R}}_{X_{\CMcal{S}}}}$ \\  \hline
${{\bf{x}}_{\CMcal{U}}}$ & An instance of ${X_{\CMcal{U}}}$ \\  \hline
${{\bf{x}}_{\CMcal{S}}}$ & An instance of ${X_{\CMcal{S}}}$ \\  \hline
${{\bf{x}}}$ & $= ( {{\bf{x}}_{\CMcal{U}}}, {{\bf{x}}_{\CMcal{S}}} )$, an instance of $X$ \\  \hline
${T_{{\bf{x}}_{\CMcal{U}}}}$ & $ = \{ {\bf{x}'} \in {{\CMcal{R}}_{X}} \mid {{\bf{x}'}_{\CMcal{U}}} = {\bf{x}_{\CMcal{U}}} \} = $ range of $( {{\bf{x}}_{\CMcal{U}}}, {X_{\CMcal{S}}} )$  \\  \hline
$A$ & The r.v. of decision outcome \\  \hline
${\CMcal{A}}$ & Range of $A$ \\  \hline
$P(\cdot)$ & Aleatory probability; \emph{chance} \\  \hline
& \\[-1em]
${\tilde{P}}(\cdot)$ & Epistemic probability; \emph{credence} or \emph{belief} \\  \hline
$D(X)$ & $ = \{ {P}(A = a|X) \mid \forall a \in {\CMcal{A}} \}$; decision mapping (Definition \ref{def:decision_mapping}) \\  \hline
& \\[-1em]
${\tilde{D}}(X)$ & $ = \{ {{\tilde{P}}}(A = a|X) \mid \forall a \in {\CMcal{A}} \}$; announced decision mapping \\  \hline
$d(X)$ & $ = {P}(A = 1|X)$; decision rules (Definition \ref{def:decision_mapping}) \\  \hline
& \\[-1em]
${\tilde{d}}(X)$ & $ = {{\tilde{P}}}(A = 1|X)$; announced decision rules  \\  \hline
${\CMcal{M}}$ & A privacy protection scheme for an ATR \\  \hline
\end{tabular}
\end{table}

In practice, since honest-but-curious adversaries can be ubiquitous, the background knowledge that adversaries may possess could be diverse and unknown to agencies in charge of ATRs. Therefore, it is important that agencies should consider \emph{the worse-case scenario}, i.e., the most information that an honest-but-curious adversary can possess (which is the worst-case weak adversary). Hence, the agencies should assume that an adversary could possess \emph{precise and full} knowledge of 
\begin{itemize}
    \item the range ${{\CMcal{R}}_{X}}$ and the joint distribution ${P_X}(\bf{x})$ of all inputs ${\bf{x}}$;
    \item all public records (a.k.a. \textit{quasi-identifier} (QID) \cite{sweeney1997weaving, sweeney2000simple, acquisti2009predicting}) ${{\bf{x}}_{\CMcal{U}}}$ of specific individuals;
    \item the received decisions $a$ of the targeted individuals;
    \item the internal privacy parameters (e.g., the predefined required privacy level) of the privacy protection scheme ${\CMcal{M}}$ used for an ATR, if any.
\end{itemize}
The following information is assumed in general unknown (or known with little confidence) by adversaries before seeing an ATR: (i) data subjects' private records ${{\bf{x}}_{\CMcal{S}}}$ and (ii) the decision mapping $D$ of the black-box. Given the above adversarial settings, we clearly define privacy violation in releasing ATRs in the following.

\begin{defn}
\label{def:privacy_violation}
The release of an ATR is privacy violating if any private or confidential information of any data subject to whom decision algorithms, disclosed in the ATR, have been applied can be (unintentionally) inferred by any honest-but-curious unauthorized individual or entity to whom the ATR is released, with confidence exceeding a tolerable threshold, due to the release of the ATR.
\end{defn}

\begin{rmk}
\label{rmk:privacy_violation}
Given Definition \ref{def:privacy_violation}, inferring attribute values due to high correlations between attributes, e.g., knowing people who have ovarian cancer are female, should not be mistaken as privacy breach (not private information; not via an ATR). Similarly, releasing ATRs to a doctor for the ML-assist diagnoses of his patients should not be considered as privacy violation (an authorized personal).
\end{rmk}

\subsection{Comparison with PPDM and PPDP}
The main differences between privacy preservation in ATRs and privacy preservation in data-mining (PPDM) and data-publishing (PPDP) are their \emph{adversarial settings}. 

More specifically, in the PPDM setting, a dataset is not published; instead, users or data analysts send queries (a set of pre-defined/allowed deterministic functions, e.g., average, count, median, max, and min) to the curator, and the curator generates the corresponding query outputs based on the dataset. In such a setting, if the pre-defined queries are carefully designed, an adversary (a malicious user), in general, \emph{is not able to determine the direct mappings between public and private attribute values of a record} nor \emph{any private information of any individual} from any single query output. However, since query functions are known in advance, an adversary \emph{can send multiple queries and compare the obtained results} to extract data subjects' private information from the outputs. In this regard, differential privacy (DP) \cite{dwork2006calibrating, chen2016oblivious} is usually adopted to preserve privacy in PPDM.
In summary, the main differences between the settings in PPDM and ATRs are (i) in PPDM, mappings between public and private attributes are in general not available, or may be known only partially, while these could be known \emph{statistically} in the ATRs setting; (ii) an adversary can send multiple (deterministic) queries and/or collude with other adversaries to extract data subjects' private information, while an ATR is a one-shot announcement, and the announced decision mappings from the inputs to the outputs could be probabilistic (i.e., random decisions).

In the PPDP setting, a dataset is published. Therefore, the mappings between public and private attributes are \emph{clearly known} to an adversary (much stronger than auxiliary or side-information). When the published dataset shows \emph{uniqueness of a public record} or \emph{unique relationship between certain public and private attributes}, an adversary can utilize such uniqueness to identify data subjects' private information. Therefore, several techniques ($k$-anonymity \cite{samarati1998generalizing, samarati1998protecting}, $l$-diversity \cite{machanavajjhala2006diversity, machanavajjhala2007diversity}, etc.) are proposed in the literature to obfuscate such uniqueness in order to preserve data subjects' privacy. 
In summary, the main differences between the settings in PPDP and ATRs are (i) unlike in PPDP where privacy is leaked due to strong inference between attributes, in ATR, as we have emphasized in Section \ref{subsec:adversary}, we only consider the case of weak correlation/inference between public and private attributes, i.e., an adversary is not able to identify any private information with high confidence before seeing an ATR, while the confidence could be dramatically enhanced after an ATR is released; (ii) in PPDP, depending on the application, there may or may not exist output attributes. When there exist output attributes, as the dataset is published, \emph{all} output attribute values are available to an adversary, which could provide strong inference between some sensitive attributes and the output attributes (for learning purposes), and thus we need to guarantee that any output attribute value is not directly associated with any individual; while in ATRs, we consider the case that a decision outcome could be directly associated with an individual (credit card applications, university admissions, etc.), but an adversary knows a few decision outcomes only.



\section{Privacy, Utility, and Measured Fairness}\label{sec:privacy_fidelity}
Given clear context of adversarial settings and definition of privacy violation for releasing ATRs, we next formulate privacy leakage caused by inference attacks, and propose a privacy-preserving mechanism for ATRs. To this end, in this section, we provide a privacy measure to mathematically characterize and formulate the degree of privacy leakage. Based on the proposed privacy measure, we formulate the requirements for a privacy-preserving mechanism for ATRs, and introduce a utility measure to characterize the influences caused by the proposed privacy-preserving mechanism; similarly, we address the influence of the proposed privacy mechanism on fairness measures.

\subsection{Privacy Measure and Privacy-Preserving Mechanism}
\label{subsec:privacy} 
Recall in Section \ref{sec:problem} we have seen privacy leakage disasters when decision rules were divulged. The fundamental problem is that transparency schemes as well as fairness measures are closely related to, or functions of, decision mapping $D$, and more importantly, if $D$ provides strong inference from public knowledge to sensitive records, once it is utilized in an ATR and obtained by an adversary, the adversary can utilize it to further acquire data subjects' secrets with high confidence.

In light of this, here we propose the following: a carefully processed $D$, denoted by ${\tilde{D}}$, should be adopted as a substitute for $D$ in an ATR for preserving data subjects' privacy. ${\tilde{D}}$ should satisfy certain privacy requirements and can be safely announced (if an ATR chooses to release an interpretable surrogate model) or utilized by transparency schemes and fairness measures provided in an ATR. 

In such a case, when an ATR is released, an adversary acquires information about $\tilde{D}$, and could further utilize it in an inference channel $\langle {{X}_{\CMcal{U}}},A \xrightarrow{\small{{P_X},{\tilde{D}}}} {{X}_{\CMcal{S}}} \rangle$ which maps from inference source ${{X}_{\CMcal{U}}}$ and $A$ to sensitive attribute values ${X_{\CMcal{S}}}$. (When the context is clear, we will omit $P_X$ and $\tilde{D}$ above the arrow for simplicity.) 
Based on the adversarial settings in Section \ref{subsec:adversary}, one reasonable privacy measure to characterize the above inference (caused by $\tilde{D}$) is the \emph{maximum confidence} of an adversary in inferring \emph{any} data subject's sensitive value ${X_{\CMcal{S}}}$, which is also known as the \emph{worst-case posterior vulnerability} \cite{espinoza2013min}. 
In other words, even if an adversary knows ${\tilde{D}}$ and further utilizes it to perform inference attacks, the maximal confidence that the adversary can have is carefully controlled in advance to prevent privacy violation. In this regard, privacy measures of the announced version of decision mapping ${\tilde{D}}$ used in an ATR should reflect the maximal degree of an adversary's confidence in inferring any data subject's secret via ${\tilde{D}}$. 
Consider the case in which ${\CMcal{S}}$ is a singleton set, we define maximum confidence formally in the following.

\begin{defn}
\label{def:max_conf}
(\textit{Maximum Confidence}) Given the adversarial settings in Section \ref{subsec:adversary} and an inference channel $\langle {{X}_{\CMcal{U}}},A \to {{X}_{\CMcal{S}}} \rangle$, the confidence of inferring a certain sensitive attribute value ${x_{\CMcal{S}}}$ from a certain inference source $( {{\bf{x}}_{\CMcal{U}}}, a )$, denoted by $\mathit{conf}({{\bf{x}}_{\CMcal{U}}},a \to {{x}_{\CMcal{S}}})$, is the posterior epistemic probability of ${{x}_{\CMcal{S}}}$ given ${{\bf{x}}_{\CMcal{U}}}$ and $a$ as follow
\begin{equation*}
\mathit{conf}({{\bf{x}}_{\CMcal{U}}},a \to {{x}_{\CMcal{S}}})
= {\tilde{P}_{{X_{\CMcal{S}}}|{X_{\CMcal{U}}},A}}({x_{\CMcal{S}}}|{\bf{x}_{\CMcal{U}}}, a).
\end{equation*}
The maximum confidence of inferring a specific sensitive attribute value ${x_{\CMcal{S}}}$ from any inference sources, denoted by $\mathit{Conf}({X_{\CMcal{U}}},A \to {{x}_{\CMcal{S}}})$, is defined as 
\begin{equation*}
\mathit{Conf}({X_{\CMcal{U}}},A \to {{x}_{\CMcal{S}}}) \triangleq {\max_{{{\bf{x}}_{\CMcal{U}}}, a}} \{\mathit{conf}({{\bf{x}}_{\CMcal{U}}},a \to {{x}_{\CMcal{S}}})\}.
\end{equation*}
Accordingly, the maximum confidence of inferring any sensitive attribute value from any inference channel is 
\begin{equation*}
\mathit{Conf}({X_{\CMcal{U}}},A \to {{X}_{\CMcal{S}}}) \triangleq {\max_{{{\bf{x}}_{\CMcal{U}}}, a, {{x}_{\CMcal{S}}}}} \{\mathit{conf}({{\bf{x}}_{\CMcal{U}}},a \to {{x}_{\CMcal{S}}})\}. \qedhere
\end{equation*}
\end{defn}

The privacy requirement, similar to confidence bounding \cite{wang2005integrating,wang2007handicapping}, $\beta$-likeness \cite{cao2012publishing}, and privacy enforcement in \cite{kuvcera2017synthesis}, restricts the maximum confidence on inferring any sensitive attribute by a confidence threshold, a pre-determined privacy parameter $\beta$. 


\begin{defn}
\label{def:beta_max_conf}
($\beta$-\textit{Maximum Confidence}) In an algorithmic transparency report, $\tilde{D}$ satisfies the privacy requirement $\beta$-\textit{Maximum Confidence} if $\mathit{Conf}({X_{\CMcal{U}}},A \to {X_{\CMcal{S}}}) \le \beta$.
\end{defn}

\begin{lem}
\label{lem:beta_max_conf}
The privacy requirement $\beta$-Maximum Confidence imposes the following constraints to the announced decision mapping $\tilde{D}$, $\forall {\bf{x}} \in {{\CMcal{R}}_{X}}$, $\forall a \in {{\CMcal{A}}}$,
\begin{equation}
\label{eq:beta_max_conf}
{ \frac{ {{{\tilde{D}}_{a}}({\bf{x}})}{{P_X}(\bf{x})} }{ \sum\limits_{{\bf{x}'} \in {T_{{\bf{x}}_{\CMcal{U}}}}} {{{\tilde{D}}_{a}}({\bf{x}'})}{{P_X}({\bf{x}'})} } } \le \beta.
\end{equation}
\end{lem}
\begin{proof}
Please refer to Appendix \ref{sec:appendix_00} for detailed proof.
\end{proof}

\begin{rmk}
\label{rmk:privacy}
Note that a privacy requirement which only prevents an adversary from \emph{correctly} inferring any sensitive attribute value of any data subject is insufficient. The reason is that an adversary can possess the knowledge of the privacy protection scheme and its internal privacy parameters. If the privacy requirement allows an adversary to incorrectly infer wrong sensitive values with arbitrary high confidence, since an adversary may know the privacy requirement, he/she may perceive that \emph{any sensitive attribute value which can be inferred with confidence higher than the threshold is an incorrect one}; this could become additional side-information for the adversary. An adversary can further utilize such extra side-information to narrow down the range of conjectures, which enhances the confidence of correctly guessing the right sensitive value. The enhanced confidence could result in exceeding the privacy threshold, and thus cause a privacy hazard. 
\end{rmk}

The advantage of using maximum confidence as a privacy measure is that it results in intuitive understanding of $\beta$. This could be important when a privacy scheme is used for an ATR, the regulation may require a plain explanation for the adopted privacy scheme as well as the corresponding settings and meanings of its parameters. Alternatively, one can use other privacy measures, e.g., \emph{minimum uncertainty} (Appendix \ref{sec:min_uncertainty}), which is essentially conveying the same concept as maximum confidence, but the privacy parameter $\gamma$ grows with the strength of privacy. 

A privacy protection scheme ${\CMcal{M}}$ takes the original/true decision mapping $D$ as the input and generates a privacy-preserving decision mapping $\tilde{D}$ safe for announcement with careful processing based on privacy requirements. Inevitably, the original $D$ would differ from the generated $\tilde{D}$, which is a distorted/perturbed but private version of $D$. 
%
In the next section, we introduce a utility measure to characterize the distortion. 


\subsection{Utility Measure: Fidelity}
\label{subsec:fidelity}
In this section, we introduce an appropriate utility measure for our problem. Given proposed $D \xrightarrow{{\CMcal{M}}} {\tilde{D}}$, an appropriate utility measure should characterize the distortion from ${\tilde{D}}$ to $D$, or quantify the quality of faithfulness of ${\tilde{D}}$ (compared with $D$), and hence, particularly, is named \emph{fidelity} measure hereafter. By imposing fidelity constraints to ${\CMcal{M}}$, the maximal distortion between ${\tilde{D}}$ and $D$ is guaranteed to be bounded accordingly. 
%


\begin{defn}
\label{def:delta_fidelity}
(\textit{$\delta$-Fidelity}) A privacy perturbation method ${\CMcal{M}}:\Delta({\CMcal{A}})$ $\to \Delta({\CMcal{A}})$ satisfies $\delta$-fidelity, $\delta \in [0,1]$, if $\forall {\bf{x}} \in {{\CMcal{R}}_{X}}$ and $\forall a \in {\CMcal{A}}$, we have
\begin{equation}
\label{eq:delta_fidelity}
\abs{{{{\tilde{D}_a}}({\bf{x}})} - {{D_a}({\bf{x}})}} \le 1-\delta. \qedhere
\end{equation}
\end{defn}

\begin{defn}
\label{def:alpha_fidelity}
(\textit{$\alpha$-Fidelity}) A privacy perturbation method ${\CMcal{M}}:\Delta({\CMcal{A}})$ $\to \Delta({\CMcal{A}})$ satisfies $\alpha$-fidelity, $\alpha \in [0,1]$, if $\forall {\bf{x}} \in {{\CMcal{R}}_{X}}$ and $\forall a \in {\CMcal{A}}$, we have
\begin{equation}
\label{eq:alpha_fidelity}
\alpha \le {{{\tilde{D}_a}}({\bf{x}})}/{{D_a}({\bf{x}})} \le 1/{\alpha}. \qedhere
\end{equation}
\end{defn}

In the most general form, definition of fidelity can be 
\begin{equation}
\label{eq:fidelity_general}
{{{{\tilde{D}_a}}({\bf{x}})}_{\text{min}}} \le {{\tilde{D}_a}}({\bf{x}}) \le {{{{\tilde{D}_a}}({\bf{x}})}_{\text{max}}}, 
\end{equation}
which describes the restriction (the allowed range) of distortion of ${\tilde{D}}$ in a very general manner. The corresponding equivalent representations for $\delta$- and $\alpha$-fidelity are
\begin{align}
{{D_a}({\bf{x}})}-(1-\delta) \le {{\tilde{D}_a}}({\bf{x}})& \le {{D_a}({\bf{x}})}+(1-\delta), \label{eq:fidelity_delta} \\
\alpha{{D_a}({\bf{x}})} \le  {{\tilde{D}_a}}({\bf{x}})& \le \frac{1}{\alpha} {{D_a}({\bf{x}})}, \label{eq:fidelity_alpha}
\end{align}
in which the upper and the lower bounds ${{{{\tilde{D}_a}}({\bf{x}})}_{\text{max}}}$ and ${{{{\tilde{D}_a}}({\bf{x}})}_{\text{min}}}$ are functions of $D$ and $\delta$, or $\alpha$. In practice, $\delta$ and $\alpha$ should not be far from $1$.

\subsection{Influence of Privacy on Measured Fairness}\label{subsec:fidelity_fairness}
As demonstrated in Section \ref{subsec:privacy_fairness}, since fairness measures are functions of decision mapping/rule, releasing precise fairness measures could bring privacy hazards. On account of this, the released fairness measures should be computed based on privacy-preserving ${\tilde{D}}$, which, however, would influence and distort the measured bias $\varepsilon$. 
%
In this section, we show that by knowing the fidelity constraints to ${\CMcal{M}}$, we are able to characterize and bound the distortion of the measured fairness/bias.

\begin{figure}[!t]
\centering
\includegraphics[width=4.5in]{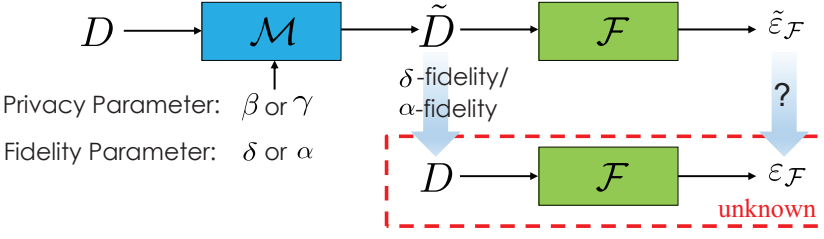}
\caption{A depiction of how fidelity of $\tilde{D}$ can aid in characterizing the difference between the measured bias ${\tilde{\varepsilon}}_{\CMcal{F}}$ and the true bias ${\varepsilon}_{\CMcal{F}}$, where ${\CMcal{F}}$ denotes a (general or specific) set of fairness definitions on which bias is computed. In Section \ref{sec:problem}, we have seen that releasing $D$ and ${\varepsilon}_{\CMcal{F}}$ can cause privacy leakage and thus should be prohibited, so that ${\varepsilon}_{\CMcal{F}}$ is unknown to the public. However, if the fidelity parameter used in ${\CMcal{M}}$ is known, we are able to characterize ${\varepsilon}_{\CMcal{F}}$ by ${\tilde{\varepsilon}}_{\CMcal{F}}$ based on Lemma \ref{lem:fidelity_2_fairness_tv}.
}
\label{fig:bias_measure}
\vspace*{-3mm}
\end{figure}

Fig. \ref{fig:bias_measure} is a representative illustration of the true bias ${\varepsilon}_{\CMcal{F}}$ and the measured bias ${\tilde{\varepsilon}}_{\CMcal{F}}$ influenced by ${\CMcal{M}}$, where ${\CMcal{F}}$ denotes a (general or specific) set of fairness definitions on which bias is computed.
Since the true decision mapping $D$ should not be released and utilized to compute fairness measures, the true bias computed based on $D$ is generally unknown. A natural question may arise: by knowing ${\tilde{\varepsilon}}_{\CMcal{F}}$, and the degree of fidelity of ${\tilde{D}}$ ($\delta$ or $\alpha$), what can we know about ${\varepsilon}_{\CMcal{F}}$? The following lemma answers the question: if the maximum distortion from ${\tilde{D}}$ to $D$ is known, the maximum distortion from ${\tilde{\varepsilon}}_{\CMcal{F}}$ to ${\varepsilon}_{\CMcal{F}}$ can be known, and thus the range of ${\varepsilon}_{\CMcal{F}}$ can be known.

\begin{lem}
\label{lem:fidelity_2_fairness_tv}
Let ${{\CMcal{F}}_{\textit{tv}}}$ denote the set of all total-variation-based fairness definitions, and ${{\CMcal{F}}_{\textit{rm}}}$ the set of all relative-metric-based fairness definitions (see Appendix \ref{sec:fairness}).
Given $D \xrightarrow{{\CMcal{M}}} {\tilde{D}}$, if $\CMcal{M}$ satisfies $\delta$-fidelity, we can guarantee the measured bias ${{\tilde{\varepsilon}}_{{\CMcal{F}}_{\textit{tv}}}}$ satisfies
\begin{equation}
\abs{ {{\tilde{\varepsilon}}_{{\CMcal{F}}_{\textit{tv}}}} - {{\varepsilon}_{{\CMcal{F}}_{\textit{tv}}}} } \le \min\{2(1-\delta),1\}.
\end{equation}
If $\CMcal{M}$ satisfies $\alpha$-fidelity, we can guarantee the measured bias ${{\tilde{\varepsilon}}_{{\CMcal{F}}_{\textit{rm}}}}$ satisfies
\begin{equation}
\abs{ {{\tilde{\varepsilon}}_{{\CMcal{F}}_{\textit{rm}}}} - {{\varepsilon}_{{\CMcal{F}}_{\textit{rm}}}} } \le \min\{-2\log\alpha,1\}.
\end{equation}
\end{lem}
\begin{proof}
By applying the \emph{reverse triangle inequality} \cite{thompson1978matrix}, the results trivially follow.
\end{proof}


\section{Privacy-Fidelity Trade-off}\label{sec:optimization}
Strong privacy perturbation could cause serious distortion on the announced information including decision rules and fairness measures,
%
and thus a privacy protection scheme should preserve privacy while guaranteeing a certain degree of fidelity to the announced information; this turns out a privacy-fidelity trade-off problem.
In this section, we mathematically formulate the trade-off problem, and revisit existing algorithms that can efficiently solve this problem.

\subsection{Optimization Formulation}
We describe the privacy-fidelity trade-off problem in the following: given fidelity constraints, we would like to find the greatest privacy (the smallest $\beta$) that we can achieve. The problem is mathematically formulated as follow. For conciseness, we omit the subscript of all probability measures and simply write, e.g., $P({\bf{x}})$ instead of ${P_{X}}({\bf{x}})$.
\begin{subequations}
\begin{alignat}{2}
\underline{\textbf{OPT}} &({{\CMcal{R}}_{X}} \ntimes {\CMcal{A}} ): \label{eq:OPT} \tag{OPT} \\
 \underset{{\tilde{D}}}{\text{min}}
       \quad & \beta \label{eq:objective}\\
 \text{s.t.}
       \quad & { \frac{ {{P}(\bf{x})}{{{\tilde{D}}_a}({\bf{x}})} }{ \sum\limits_{{\bf{x}'} \in {T_{{\bf{x}}_{\CMcal{U}}}}} {{P}({\bf{x}'})}{{{\tilde{D}}_a}({\bf{x}'})} } } \le \beta \,, \; && \forall {{\bf{x}}} \in {{\CMcal{R}}_{X}}, \forall a \in \CMcal{A} \label{eq:privacy_constraints}\\
       \quad & \quad\quad\text{ }\text{ }  {{\tilde{D}_a}}({\bf{x}}) \le {{{{\tilde{D}_a}}({\bf{x}})}_{\text{max}}} \,, \;  && \forall {{\bf{x}}} \in {{\CMcal{R}}_{X}}, \forall a \in \CMcal{A}  \label{eq:fairness_constraints_max}\\
       \quad & \quad\quad\text{ }\text{ }  {{\tilde{D}_a}}({\bf{x}}) \ge {{{{\tilde{D}_a}}({\bf{x}})}_{\text{min}}} \,, \;  && \forall {{\bf{x}}} \in {{\CMcal{R}}_{X}}, \forall a \in \CMcal{A}  \label{eq:fairness_constraints_min}\\
      \quad & \quad\quad\quad\quad\quad\quad {{{\tilde{D}}_a}({\bf{x}})} \ge 0 \,, \;  && \forall {{\bf{x}}} \in {{\CMcal{R}}_{X}}, \forall a \in \CMcal{A} \label{eq:nonnegativity_constraints} \\
      \quad & \quad\quad\quad\quad \sum_{a \in \CMcal{A}}{{{\tilde{D}}_a}({\bf{x}})} = 1 \,, \; && \forall {{\bf{x}}} \in {{\CMcal{R}}_{X}} \,. \label{eq:tol_prob_constraints}
\end{alignat}
\end{subequations}
The first constraint in \eqref{eq:privacy_constraints} is the privacy constraint $\beta$-Maximum Confidence defined in Definition \ref{def:beta_max_conf} and Lemma \ref{lem:beta_max_conf}, and the last two constraints in \eqref{eq:nonnegativity_constraints} and \eqref{eq:tol_prob_constraints} are probability distribution conditions. The second and the third constraints in \eqref{eq:fairness_constraints_max} and \eqref{eq:fairness_constraints_min} are fidelity constraints introduced in \eqref{eq:fidelity_general}. Its corresponding representations for $\delta$- or $\alpha$-fidelity can be found in \eqref{eq:fidelity_delta} and \eqref{eq:fidelity_alpha}, respectively. The objective in \eqref{eq:objective} is to find the minimal $\beta$ subject to the feasibility of $\tilde{D}$ based on the above-mentioned constraints. \emph{A careful observation of the optimization problem \eqref{eq:OPT} will reveal that it is an equivalent formulation of a generalized linear fractional programming (LFP) problem \cite{zionts1968programming}}.

\subsection{Drawbacks of Existing Methods to Solve Generalized LFP Problems}
It has been known that a generalized LFP is quasi-convex and not reducible to a linear programming (LP) problem; however, it can be solved as a sequence of LP feasibility problems \cite{boyd2004convex}, i.e., solving numerous sub-level LP problems iteratively according to bisection method. By efficient algorithms such as interior point method, the solution of a LP problem can be obtained in pseudo-polynomial time $O({\frac{n^3}{\log n}}L)$ \cite{anstreicher1999linear}, where $n$ is the number of variables, $L$ is the input size, i.e., the length of the binary coding of the input data to represent the problem, which is roughly proportional to the number of constraints. 
For our problem, based on \eqref{eq:privacy_constraints}--\eqref{eq:tol_prob_constraints}, it is clear that the number of constraints is proportional to $| {{\CMcal{R}}_{X}} \ntimes {\CMcal{A}} |$, which is the number of variables $n = |{{{\tilde{D}}_a}({\bf{x}})}|$, exponential in the number of input attributes $K$.
For example, suppose the cardinality for each input attribute is consistent, e.g., $|{X_k}|=l$, $\forall k = 1, \ldots, K$, we have $n = | {{\CMcal{R}}_{X}} \ntimes {\CMcal{A}} | = 2 \cdot l^K$ and roughly $4n = 8 \cdot l^K$ constraints.
Even for a relatively small example, e.g., a binary decision process with $K=10$ input attributes, each with $l=5$ possible values, we have $n \approx 2 \cdot 10^{7}$ variables and $\approx 8 \cdot 10^{7}$ constraints for each sub-level LP problem, with time complexity $O({\frac{n^4}{\log n}})$, i.e., not tractable on typical machines (e.g., as reported in \cite{lp_benchmark}, \quotess{spal\_004} with $10203$ rows (w.r.t. constrains) and $321696$ columns (w.r.t. variables) can encounter out of memory or timeout ($>25,000$ seconds) issue on a Linux-PC with a 4GHz i7-4790K CPU and 32GB RAM).
To solve a generalized LFP problem, we need to solve such a huge sub-level LP problem iteratively. 
In practice, ML algorithms may require nontrivial amounts of attributes to aid in decision-making; therefore, without an efficient solver, the privacy-fidelity trade-off problem could be intractable and the feasibility of the associated privacy protection scheme could be dramatically reduced. \emph{In this regard, it is crucial and essential to propose a more efficient method to solve the proposed privacy-fidelity trade-off optimization problem.}

\section{Linear-Time Optimal Privacy Solution}\label{sec:optimal_solution}
In this section, we analyze the optimization problem \eqref{eq:OPT}, reveal its important properties, and propose efficient methods to solve it.
We first investigate the \emph{decomposability} of the optimization problem, i.e, whether the problem can be decomposed into several smaller sub-problems for efficient solving. We find \eqref{eq:OPT} decomposable and can be solved using divide-and-conquer approach. In addition, we propose \emph{closed-form solutions} for each optimization sub-problem. The optimization problem can thus be solved very efficiently by solving multiple independent sub-problems in \emph{linear time}. Moreover, analysis insights into the optimal solutions are also provided in this section.

\subsection{Decomposability}
In the following, we show that the optimization problem can actually be decomposed into numerous small sub-problems and thus can be solved more efficiently. An optimization problem is \emph{separable} or \emph{trivially parallelizable} if the variables can be partitioned into disjoint subvectors and each constraint involves only variables from one of the subvectors \cite{boyd2007notes}. By observing (i) each constraint in \eqref{eq:fairness_constraints_max}, \eqref{eq:fairness_constraints_min}, and \eqref{eq:nonnegativity_constraints} involves only a single variable ${{{\tilde{D}}_{a}}({\bf{x}})}$, (ii) each constraint in \eqref{eq:tol_prob_constraints} involves a set of variables $\{ {{{\tilde{D}}_{a}}({\bf{x}})} \mid \forall a \in {\CMcal{A}} \}$, and (iii) each constraint in \eqref{eq:privacy_constraints} involves a set of variables $\{ {{{\tilde{D}}_{a}}({\bf{x}})} \mid \forall {\bf{x}} \in {{T_{{\bf{x}}_{\CMcal{U}}}}} \}$, we notice that any variable ${{{\tilde{D}}_{a}}({\bf{x}})}$ is a \textit{complicating variable} in ${T_{{\bf{x}}_{\CMcal{U}}}} \ntimes {\CMcal{A}}$ but is irrelevant to any other variables outside the QID group ${T_{{\bf{x}}_{\CMcal{U}}}}$. Hence, \eqref{eq:privacy_constraints}--\eqref{eq:tol_prob_constraints} are \textit{complicating constraints} within a tuple but \textit{separable constraints} among tuples. \eqref{eq:OPT} can thus be decomposed into multiple smaller sub-problems; each focuses on a particular QID group only. Let $h({{{\tilde{D}}_{a}}({\bf{x}})}, \beta) \ge 0$ be the affine function representing all linear inequality constraints \eqref{eq:privacy_constraints}--\eqref{eq:nonnegativity_constraints}. An optimization sub-problem can thus be expressed as follow.
\begin{subequations}
\begin{align}
\underline{\textbf{OPT}} & \underline{\textbf{-SUB}} ({T_{{\bf{x}}_{\CMcal{U}}}} \ntimes {\CMcal{A}}): \label{eq:OPT_sub} \tag{OPT-Sub} \\
    \underset{{\tilde{D}}}{\text{min}}
        \quad & \beta   \label{eq:sub_objective} \tag{OBJ-Sub} \\
    \text{s.t.} 
        \quad & h({{{\tilde{D}}_{a}}({\bf{x}})}, \beta) \ge 0 \,, \; \forall {{\bf{x}}} \in {T_{{\bf{x}}_{\CMcal{U}}}}, \forall a \in \CMcal{A} \tag{INEQ-Sub} \label{eq:sub_ineq_constraints}\\
        \quad & \quad \sum_{a \in \CMcal{A}}{{{\tilde{D}}_a}({\bf{x}})} = 1 \,, \;  \forall {{\bf{x}}} \in {T_{{\bf{x}}_{\CMcal{U}}}} \,.  \tag{EQ-Sub} \label{eq:sub_eq_constraints}
\end{align}
\end{subequations}

\eqref{eq:OPT} is then equivalent to the \emph{master problem} below.

\begin{subequations}
\begin{align}
\underline{\textbf{OPT}} & \underline{\textbf{-MASTER}} ({{\CMcal{R}}_{X}} \ntimes {\CMcal{A}}): \label{eq:OPT_MS} \tag{OPT-MS} \\
    \underset{{\tilde{D}}}{\text{min}}
        \quad & \beta   \label{eq:master_objective} \tag{OBJ-MS}\\
    \text{s.t.} 
        \quad & \text{(\ref{eq:sub_ineq_constraints}$({T_{{\bf{x}}_{\CMcal{U}}}} \ntimes {\CMcal{A}})$)} \,, \; \forall {T_{{\bf{x}}_{\CMcal{U}}}} \subseteq {{\CMcal{R}}_{X}} \label{eq:master_ineq_constraints} \tag{INEQ-MS}\\
        \quad & \quad \text{(\ref{eq:sub_eq_constraints}$({T_{{\bf{x}}_{\CMcal{U}}}} \ntimes {\CMcal{A}})$)} \,, \;  \forall {T_{{\bf{x}}_{\CMcal{U}}}} \subseteq {{\CMcal{R}}_{X}} \,.  \label{eq:master_eq_constraints} \tag{EQ-MS}
\end{align}
\end{subequations}

\begin{lem}
\label{lem:Separability_OPT}
Let $\beta^{*}_{T_{{\bf{x}}_{\CMcal{U}}}}$ denote the optimal value of a sub-problem \eqref{eq:OPT_sub}, $\beta^{*}$ the optimal value of \eqref{eq:OPT}. We have $\beta^{*} = \underset{{T_{{\bf{x}}_{\CMcal{U}}}} \subseteq {{\CMcal{R}}_{X}}}{\text{max}}{\beta^{*}_{T_{{\bf{x}}_{\CMcal{U}}}}}$.
\end{lem}
\begin{proof}
Since \eqref{eq:OPT} is a generalized LFP (in an equivalent formulation), according to \ref{eq:OPT_MS}, the result trivially follows.
\end{proof}
The Lemma above basically tells us that given the same fidelity constraints, the \emph{overall} highest privacy guarantee $\beta^{*}$ is the largest $\beta^{*}_{T_{{\bf{x}}_{\CMcal{U}}}}$ among all sub-problems, i.e., the \emph{weakest} optimal privacy guarantee among all QID groups.

\subsection{Solution Properties} \label{subsec:solution_prop}
According to the decomposability of the optimization problem, in the following, we only need to focus on solving an optimization sub-problem \eqref{eq:OPT_sub}. To characterize the privacy-fidelity trade-off, we are particularly interested in where the trade-off starts and ends. In this section, we propose lemmas addressing the above question. 

Before introducing the lemmas, we first define a useful quantity which will be further utilized to characterize the trade-off.
\begin{defn}
\label{def:max_conf_tuple}
\textit{(Maximum Posterior Confidence)} Given an optimization sub-problem \eqref{eq:OPT_sub} and 1-fidelity (100\% faithfulness) requirement, i.e., $\alpha=\delta=1$ and ${\tilde{D}} = D$, the highest confidence that an adversary can have on inferring any sensitive information from any decision outcome, denoted by $C^{*}$, is $C^{*} \triangleq \mathit{Conf}({X_{\CMcal{U}}} = {{\bf{x}}_{\CMcal{U}}},A \xrightarrow{\small{{P_X},{D}}} {X_{\CMcal{S}}}) = {\underset{{a, {{x}_{\CMcal{S}}}}}{\text{max}}} \{\mathit{conf}({{\bf{x}}_{\CMcal{U}}},a \to {{x}_{\CMcal{S}}})\}$.
\end{defn}

\begin{lem}
\label{lem:optimal_utility_condition}
An \eqref{eq:OPT_sub} has the 1-fidelity solution ${{{\tilde{D}}_a}({\bf{x}})} = {{{D}_{a}}({\bf{x}})}$, $\forall {{\bf{x}}} \in {T_{{\bf{x}}_{\CMcal{U}}}}$, $\forall a \in \CMcal{A}$, iff $\beta \ge C^{*}$.
\end{lem}
\begin{proof}
Please refer to Appendix \ref{sec:appendix_01} for detailed proof. We provide intuitive explanation as proof sketch here. Since the highest confidence that an adversary can have ($C^{*}$) is lower than the privacy requirement (the confidence threshold $\beta$), it is safe to release $D$ directly, i.e., $\tilde{D} = D$ with perfect fidelity. On the other hand, as long as $C^{*}$ is greater than $\beta$, releasing $D$ violates privacy requirement and cannot be a feasible solution.
\end{proof}
\vspace{-1.5mm}
\emph{Lemma Insight} - Lemma \ref{lem:optimal_utility_condition} tells us when $\beta \ge C^{*}$, there is no trade-off between privacy and fidelity: as long as $\beta$ is greater than $C^{*}$, increasing the strength of privacy (decreasing $\beta$) would not cause degradation in fidelity. In other words, alone the strength of privacy (from low to high), the trade-off between privacy and fidelity starts when $\beta$ is right below $C^{*}$. The next lemma will tell us the end of this trade-off region.

\begin{lem}
\label{lem:feasible_condition_no_fidelity}
For $\alpha=\delta=0$, i.e., fidelity constraints are trivialized or not presented, an \eqref{eq:OPT_sub} has feasible solutions if and only if (iff) $\beta \ge {\beta_{\text{min}}} \triangleq {\max_{{{\bf{x}} \in {T_{{\bf{x}}_{\CMcal{U}}}}}}} {{P}({\bf{x}}|{T_{{\bf{x}}_{\CMcal{U}}}})}$. In other words, there exists privacy limit, the strongest privacy that we can have, based on the adversarial settings in Section \ref{subsec:adversary}.
\end{lem}
\begin{proof}
Please refer to Appendix \ref{sec:appendix_02} for detailed proof. We provide intuitive behind this lemma as proof sketch here. The privacy limit ${\max_{{{\bf{x}} \in {T_{{\bf{x}}_{\CMcal{U}}}}}}} {{P}({\bf{x}}|{T_{{\bf{x}}_{\CMcal{U}}}})}$ is the greatest conditional probability over the tuple\footnote{Since $\forall {\bf{x}} \in {T_{{\bf{x}}_{\CMcal{U}}}}$, ${{\bf{x}}_{\CMcal{U}}}$ is the same, ${{P}({\bf{x}}|{T_{{\bf{x}}_{\CMcal{U}}}})}$ is also ${{P}({{x}_{\CMcal{S}}}|{T_{{\bf{x}}_{\CMcal{U}}}})}$, the conditional distribution over all sensitive records.}, which is actually the highest possible inference confidence of an adversary \emph{before releasing an ATR}. It is the \emph{baseline confidence}, which merely utilizes knowledge of public record ${{\bf{x}}_{\CMcal{U}}}$ and side-information $P({\bf{x}})$ in an inference channel $\langle {{\bf{x}}_{\CMcal{U}}} \xrightarrow{\small{P_X}} {{x}_{\CMcal{S}}} \rangle$. Since an ATR does not contribute to such an inference channel, an associated privacy protection scheme is not able to help further reduce this baseline confidence. While achieving such a privacy limit, an ATR basically reveals zero useful information to the public.
\end{proof}
\vspace{-1.5mm}
\emph{Lemma Insight} - Lemma \ref{lem:optimal_utility_condition} and \ref{lem:feasible_condition_no_fidelity} tell us the start and the end of the privacy-fidelity trade-off region along $\beta$. Next, we show that the end point can never happen before the starting point.

\begin{lem}
\label{lem:optimal_vs_feasible_condition_no_fidelity}
${C^{*}} \ge {\beta_{\text{min}}}$.
\end{lem}
\begin{proof}
Please refer to Appendix \ref{sec:appendix_03} for detailed proof. We first provide the intuition of the lemma as a proof sketch. The intuition here is very straightforward: the maximum posterior confidence can never be lower than the maximum prior confidence (\emph{prior vulnerability cannot exceed posterior vulnerability} in \cite{m2012measuring}). Equality holds when the revealed information is completely useless.
\end{proof}
\vspace{-1.5mm}
\emph{Lemma Insight} - According to Lemma \ref{lem:optimal_utility_condition}, when $\beta \in [{C^{*}}, 1]$, the true decision mapping $D$ can be safely released without perturbation (1-fidelity). Lemma \ref{lem:feasible_condition_no_fidelity} tells us when fidelity constraints are not imposed (0-fidelity), the feasible privacy region is $\beta \in [{\beta_{\text{min}}}, 1]$. Moreover, based on Lemma \ref{lem:optimal_vs_feasible_condition_no_fidelity}, the region $[{\beta_{\text{min}}}, {C^{*}}]$ is always non-empty. Clearly, this is the region where we trade off fidelity for privacy. Next, we propose our main theorem to characterize the trade-off in this region.

\subsection{Optimal Privacy and Solutions}
In the following, we propose our main theorem, which provides the optimal-privacy solutions to the optimization sub-problem \eqref{eq:OPT_sub} in a closed-form expression, in terms of fidelity. 
%
%
Given fidelity constraints, the proposed closed-form solution yields the optimal privacy value, and thus can be utilized to analytically characterize privacy-fidelity trade-off. 

\begin{thm}
\label{thm:feasible_condition_with_fidelity}
Consider an optimization sub-problem \eqref{eq:OPT_sub} for a QID group, in which we seek for the strongest privacy guarantee given fidelity constraints. For a decision outcome $a$, define
\begin{flalign*}
\begin{split}
&{{\bf{x}}^{a}} \triangleq {\argmax_{{ {{\bf{x}} \in {T_{{\bf{x}}_{\CMcal{U}}}}} }}} { {{P}(\bf{x})} {{{{\tilde{D}_a}}({\bf{x}})}_{\text{min}}} }, \\
& b({\bf{x}}) = {{P}({{\bf{x}}})} - {{\beta}} \textstyle\sum_{{\bf{x}}' \in {T_{{\bf{x}}_{\CMcal{U}}}}} {{P}({{\bf{x}}'})},\\
&{{{{\tilde{D}_a}}({\bf{x}})}_{\text{max}'}} \triangleq {\textstyle\frac{1}{{P}({{\bf{x}}})}} \min \{ {{P}({{\bf{x}}})}{{{{\tilde{D}_a}}({\bf{x}})}_{\text{max}}}, { {{P}({{\bf{x}}^{a}})} {{{{\tilde{D}_a}}({{\bf{x}}^{a}})}_{\text{min}}} } \}, \\
&{{{{\tilde{D}_a}}({\bf{x}})}_{\text{min}'}} \triangleq {\textstyle\frac{1}{{P}({{\bf{x}}})}} \max \{ {{P}({{\bf{x}}})}{{{{\tilde{D}_a}}({\bf{x}})}_{\text{min}}}, { {{P}({{\bf{x}}^{a}})} {{{{\tilde{D}_a}}({{\bf{x}}^{a}})}_{\text{min}}} } + b({\bf{x}})\}.
\end{split}
\end{flalign*}
For binary decisions, i.e., $a \in {\CMcal{A}} = \{0,1\}$, the optimal privacy ${\beta^{*}_{T_{{\bf{x}}_{\CMcal{U}}}}} = \max \{ {{\beta}_0}, {{\beta}_1}, {{\beta}_p} \}$, where
\begin{flalign*}
{{\beta}_0} &= \frac{ {{P}({{\bf{x}}^{0}})} {{{{\tilde{D}_0}}({{\bf{x}}^{0}})}_{\text{min}}} }{ { {{P}({{\bf{x}}^{0}})} {{{{\tilde{D}_0}}({{\bf{x}}^{0}})}_{\text{min}}} } + {\sum_{{{\bf{x}} \neq {{\bf{x}}^{0}}}, {{\bf{x}} \in {T_{{\bf{x}}_{\CMcal{U}}}}}}} {{{P}(\bf{x})} {{{{\tilde{D}_0}}({\bf{x}})}_{\text{max}'}}} } \text{,} \\
{{\beta}_1} &= \frac{ {{P}({{\bf{x}}^{1}})} {{{{\tilde{D}_1}}({{\bf{x}}^{1}})}_{\text{min}}} }{ { {{P}({{\bf{x}}^{1}})} {{{{\tilde{D}_1}}({{\bf{x}}^{1}})}_{\text{min}}} } + {\sum_{{{\bf{x}} \neq {{\bf{x}}^{1}}}, {{\bf{x}} \in {T_{{\bf{x}}_{\CMcal{U}}}}}}} {{{P}(\bf{x})} {{{{\tilde{D}_1}}({\bf{x}})}_{\text{max}'}}} } \text{,} \\
{{\beta}_p} &= \frac{ {{P}({{\bf{x}}^{1}})} {{{{\tilde{D}_1}}({{\bf{x}}^{1}})}_{\text{min}}} + {{P}({{\bf{x}}^{0}})} {{{{\tilde{D}_0}}({{\bf{x}}^{0}})}_{\text{min}}} }{ {\sum_{{\bf{x}} \in {T_{{\bf{x}}_{\CMcal{U}}}}}}{{P}(\bf{x})} } \text{,} 
\end{flalign*}
and the corresponding optimal privacy solutions are
\begin{fleqn}[80pt]
\begin{alignat*}{2}
\textit{When }{\beta^{*}_{T_{{\bf{x}}_{\CMcal{U}}}}} = {{\beta}_0}: \quad &{{\tilde{D}_0}}({\bf{x}}) = {{{{\tilde{D}_0}}({\bf{x}})}_{\text{max}'}}, \text{ } \forall {\bf{x}} \in {T_{{\bf{x}}_{\CMcal{U}}}} \\
\textit{When }{\beta^{*}_{T_{{\bf{x}}_{\CMcal{U}}}}} = {{\beta}_1}: \quad &{{\tilde{D}_1}}({\bf{x}}) = {{{{\tilde{D}_1}}({\bf{x}})}_{\text{max}'}}, \text{ } \forall {\bf{x}} \in {T_{{\bf{x}}_{\CMcal{U}}}} \\
\textit{When }{\beta^{*}_{T_{{\bf{x}}_{\CMcal{U}}}}} = {{\beta}_p}: \quad &{{\tilde{D}_a}}({{\bf{x}}^{a}}) = {{{{\tilde{D}_a}}({{\bf{x}}^{a}})}_{\text{min}}}, \text{ } \forall a \in {\CMcal{A}} \\
&\textstyle\sum_{{\bf{x}} \in {T_{{\bf{x}}_{\CMcal{U}}}}} {P({\bf{x}})}{{\tilde{D}_a}}({\bf{x}}) = \textstyle\frac{1}{{\beta}_{p}} {{P}({{\bf{x}}^{a}})} {{{{\tilde{D}_a}}({{\bf{x}}^{a}})}_{\text{min}}}, \text{ } \forall a \in {\CMcal{A}} \\
&{{{{\tilde{D}_a}}({\bf{x}})}_{\text{min}'}} \le {\tilde{D}_a}({\bf{x}}) \le {{{{\tilde{D}_a}}({\bf{x}})}_{\text{max}'}}, \text{ } \forall {\bf{x}} \in {T_{{\bf{x}}_{\CMcal{U}}}}, \forall a \in {\CMcal{A}}.
\end{alignat*}
\end{fleqn}
When ${\beta^{*}_{T_{{\bf{x}}_{\CMcal{U}}}}} = {{\beta}_p}$ and $\abs{{T_{{\bf{x}}_{\CMcal{U}}}}} > 3$, we have multiple solutions.
\end{thm}
\begin{proof}
We refer readers to Appendix \ref{sec:appendix_1} for the detailed proof.
\end{proof}

\emph{Practical Implications of Theorem} - Based on Theorem \ref{thm:feasible_condition_with_fidelity}, the optimal privacy guarantee ${\beta^{*}_{T_{{\bf{x}}_{\CMcal{U}}}}}$ for each QID group can be computed analytically (in closed form), and based on Lemma \ref{lem:Separability_OPT}, the overall strongest privacy guarantee is the largest ${\beta^{*}_{T_{{\bf{x}}_{\CMcal{U}}}}}$ among all QID groups. This is particularly useful and practical in releasing ATRs - given any value of tolerable distortion, we can now easily obtain the optimal privacy value without the need of solving an optimization problem, which can then be applied to aid in determining the desired trade-off between privacy and fidelity.

\emph{Linear Time Justification of Algorithm \ref{alg:opt_privacy}} - The net time to achieve a solution to \eqref{eq:OPT} is a function of the number of sub-problems - each of which is solved via Algorithm \ref{alg:opt_privacy} in \emph{linear-time} in the number of records $n$. Given an optimization sub-problem, the number of records ${\bf{x}} \in {T_{{\bf{x}}_{\CMcal{U}}}}$, i.e., $|{T_{{\bf{x}}_{\CMcal{U}}}}| \triangleq n$ is equal to the number of sensitive attribute values, as all records in a QID group ${T_{{\bf{x}}_{\CMcal{U}}}}$ have the same public record ${{\bf{x}}_{\CMcal{U}}}$.
To see that Algorithm 1 is in $O(n)$, it is first worth noting that, based on Theorem \ref{thm:feasible_condition_with_fidelity}, the time complexity of computing ${{\bf{x}}^{a}}$ and $b({\bf{x}})$ are in $O(n)$; consequently, the time complexity of computing ${{{{\tilde{D}_a}}({\bf{x}})}_{\text{max}'}}$ and ${{{{\tilde{D}_a}}({\bf{x}})}_{\text{min}'}}$ are in $O(1)$. Given these complexities, it is clear that \ref{alg:opt_privacy}, lines 1 to 4 in Algorithm 1 is in $O(n)$; all lines from line 5 to line 15, except line 13, are in $O(1)$; function $\textproc{Allocation}$ called in line 13 is in $O(n)$ - since lines 18 to 19, as well as line 20, are in $O(n)$, line 21 and 22 are in $O(1)$, and lines 23 to 27 is in $O(n)$). 
Therefore, the time complexity of Algorithm \ref{alg:opt_privacy} is in $O(n)$. \emph{By using multi-threaded coding structures solving \quotes{parallelizable} sub-problems via Algorithm \ref{alg:opt_privacy}}, \eqref{eq:OPT} can be solved in $O(n)$.

\begin{algorithm}[!t]
\caption{Optimal Privacy Protection Scheme}
\label{alg:opt_privacy}
\begin{flushleft}
\algorithmicrequire ${P({\bf{x}})}$, ${T_{{\bf{x}}_{\CMcal{U}}}}$, ${{{{\tilde{D}_a}}({\bf{x}})}_{\text{min}}}$, ${{{{\tilde{D}_a}}({\bf{x}})}_{\text{max}}}$ \newline
\algorithmicensure ${{\tilde{D}_a}}({\bf{x}})$, $\forall a$, $\forall {\bf{x}} \in {T_{{\bf{x}}_{\CMcal{U}}}}$
\end{flushleft}
\begin{algorithmic}[1]
\For{$a \in \{0,1\}$}
	\State find ${{\bf{x}}^{a}}$
	\ForAll{${\bf{x}} \in {T_{{\bf{x}}_{\CMcal{U}}}}$}
		\State compute ${{{{\tilde{D}_a}}({\bf{x}})}_{\text{max}'}}$
	\EndFor
\EndFor
\State compute $\beta_0$, $\beta_1$, $\beta_p$, and ${\beta^{*}_{T_{{\bf{x}}_{\CMcal{U}}}}} \gets \max \{ {{\beta}_0},{{\beta}_1},{{\beta}_p} \}$
\If{${\beta^{*}_{T_{{\bf{x}}_{\CMcal{U}}}}} = {{\beta}_0}$}
	\State ${{\tilde{D}_0}}({\bf{x}}) \gets {{{{\tilde{D}_0}}({\bf{x}})}_{\text{max}'}}$ 
	\State ${{\tilde{D}_1}}({\bf{x}}) \gets 1 - {{{{\tilde{D}_0}}({\bf{x}})}_{\text{max}'}}$
\ElsIf{${\beta^{*}_{T_{{\bf{x}}_{\CMcal{U}}}}} = {{\beta}_1}$}
	\State ${{\tilde{D}_1}}({\bf{x}}) \gets {{{{\tilde{D}_1}}({\bf{x}})}_{\text{max}'}}$
	\State ${{\tilde{D}_0}}({\bf{x}}) \gets 1 - {{{{\tilde{D}_1}}({\bf{x}})}_{\text{max}'}}$
\ElsIf{${\beta^{*}_{T_{{\bf{x}}_{\CMcal{U}}}}} = {{\beta}_p}$}
	\State ${{\tilde{D}_1}}({\bf{x}}) \gets$ \textproc{Allocation}()
	\State ${{\tilde{D}_0}}({\bf{x}}) \gets 1 - {{\tilde{D}_1}}({\bf{x}})$
\EndIf
\Return ${{\tilde{D}_a}}({\bf{x}})$, $\forall a$, $\forall {\bf{x}} \in {T_{{\bf{x}}_{\CMcal{U}}}}$
\State 

\Function{Allocation}{ }
	\ForAll{${\bf{x}} \in {T_{{\bf{x}}_{\CMcal{U}}}}$, ${{\bf{x}} \neq {\bf{x}}^{1}, {\bf{x}}^{0}}$}
		\State compute ${{{{\tilde{D}_1}}({\bf{x}})}_{\text{min}'}}$
	\EndFor
	\State $\vars{resid} \gets$ RHS of \eqref{eq:balenced_1_eqv} $-\textstyle\sum_{{\bf{x}} \neq {\bf{x}}^{1}, {\bf{x}}^{0}} {P({\bf{x}})}{{{{\tilde{D}_1}}({\bf{x}})}_{\text{min}'}}$
	\State ${{\tilde{D}_1}}({{\bf{x}}^{1}}) \gets {{{{\tilde{D}_1}}({{\bf{x}}^{1}})}_{\text{min}}}$
	\State ${{\tilde{D}_1}}({{\bf{x}}^{0}}) \gets 1 - {{{{\tilde{D}_0}}({{\bf{x}}^{0}})}_{\text{min}}}$
	\ForAll{${\bf{x}} \in {T_{{\bf{x}}_{\CMcal{U}}}}$, ${{\bf{x}} \neq {\bf{x}}^{1}, {\bf{x}}^{0}}$}
		\State $\vars{capacity} \gets {{{{\tilde{D}_1}}({\bf{x}})}_{\text{max}'}} - {{{{\tilde{D}_1}}({\bf{x}})}_{\text{min}'}}$
		\State $\vars{allocation} \gets \min\{ \textstyle\frac{\vars{resid}}{{P({\bf{x}})}} , \vars{capacity}\}$
		\State ${{\tilde{D}_1}}({\bf{x}}) \gets {{{{\tilde{D}_1}}({\bf{x}})}_{\text{min}'}} + \vars{allocation}$ 
		\State $\vars{resid} \gets \vars{resid} - {{P({\bf{x}})}}\cdot\vars{allocation}$
	\EndFor
	\Return ${{\tilde{D}_1}}({\bf{x}})$, $\forall {\bf{x}} \in {T_{{\bf{x}}_{\CMcal{U}}}}$
\EndFunction
\end{algorithmic}
\end{algorithm}

\subsection{Theorem Insights on Achieving Solution Optimality} \label{subsec:insights}
In this section, we provide some insights into the optimal-privacy solutions subject to fidelity constraints in Theorem \ref{thm:feasible_condition_with_fidelity}. 

An important observation is that the optimal-privacy candidate values (${{\beta}_0}$, ${{\beta}_1}$, and ${{\beta}_p}$) the inference confidence (left-hand-side of \eqref{eq:privacy_constraints}) are fully characterized by ${ {{P}(\bf{x})}{{{\tilde{D}}_{a}}({\bf{x}})} }$ pairs of product, which are the announced \emph{joint probabilities} ${{{\tilde{P}}_{X,A}}({\bf{x}},a)}$ representing the portion of population with input record ${\bf{x}}$ receiving decision $a$, bounded within ranges $[{{P}(\bf{x})}{{{{\tilde{D}_a}}({\bf{x}})}_{\text{min}}}, {{P}(\bf{x})}{{{{\tilde{D}_a}}({\bf{x}})}_{\text{max}}}]$ due to fidelity constraints. 
Solving \eqref{eq:OPT_sub} to obtain optimal-privacy solution is thus equivalent to \quotes{tune} those pairs of product within the allowed range, for all inputs ${\bf{x}} \in {T_{{\bf{x}}_{\CMcal{U}}}}$ and outputs $a \in {\CMcal{A}}$, to minimize the maximal possible inference confidence $\beta$ of an adversary. 
Particularly, from Theorem \ref{thm:feasible_condition_with_fidelity}, it turns out that for each decision outcome instance $a$, the term ${ {{P}({{\bf{x}}^{a}})} { {{{{\tilde{D}_a}}({{\bf{x}}^{a}})}_{\text{min}}} } } = {\max_{{ {{\bf{x}} \in {T_{{\bf{x}}_{\CMcal{U}}}}} }}} { {{P}(\bf{x})} {{{{\tilde{D}_a}}({\bf{x}})}_{\text{min}}} }$, the maximum of the lower bounds of the allowed ranges over ${{\bf{x}} \in {T_{{\bf{x}}_{\CMcal{U}}}}}$, plays a crucial role in solving \eqref{eq:OPT_sub}. Next, we show that the optimal-solution for ${\bf{x}} = {{\bf{x}}^{a}}$ can only be the minimum of its allowed range.

\begin{cor}
\label{cor:max_min}
${ {{{{\tilde{D}_a}}({{\bf{x}}^{a}})}_{\text{min}}} } = { {{{{\tilde{D}_a}}({{\bf{x}}^{a}})}_{\text{min}'}} } = { {{{{\tilde{D}_a}}({{\bf{x}}^{a}})}_{\text{max}'}}}$, $\forall a \in \{0,1\}$.
\end{cor}
\begin{proof}
By definitions of ${{\bf{x}}^{a}}$ and ${{{{\tilde{D}_a}}({\bf{x}})}_{\text{max}'}}$, the result ${ {{{{\tilde{D}_a}}({{\bf{x}}^{a}})}_{\text{min}}} } = { {{{{\tilde{D}_a}}({{\bf{x}}^{a}})}_{\text{max}'}} }$ trivially follows. Based on Lemma \ref{lem:feasible_condition_no_fidelity}, we have $b({\bf{x}}) = {{P}({{\bf{x}}})} - {{\beta}_p} \sum_{{\bf{x}}'} {{P}({{\bf{x}}'})} \leq 0$, and thus by plugging ${{\bf{x}}} = {{\bf{x}}^{a}}$ into ${{{{\tilde{D}_a}}({\bf{x}})}_{\text{min}'}}$, we obtain ${ {{{{\tilde{D}_a}}({{\bf{x}}^{a}})}_{\text{min}}} } = { {{{{\tilde{D}_a}}({{\bf{x}}^{a}})}_{\text{min}'}} }$.
\end{proof}



The effective lower limits ${{P}({{\bf{x}}})}{{{{\tilde{D}_a}}({\bf{x}})}_{\text{min}'}}$ and the effective upper limits ${{P}({{\bf{x}}})}{{{{\tilde{D}_a}}({\bf{x}})}_{\text{max}'}}$ represent the feasible region where the fidelity constraints and privacy constraints intersect. 
Based on Corollary \ref{cor:max_min}, the effective upper and lower limit of ${{\bf{x}}^{a}}$ are equal, which implies that when ${\bf{x}} = {{\bf{x}}^{a}}$, the only possible value for the optimal-privacy solution is ${ {{{{\tilde{D}_a}}({{\bf{x}}^{a}})}_{\text{min}}} }$. From Theorem \ref{thm:feasible_condition_with_fidelity}, we can see that this is true \emph{for all the three cases}. It is worth noting that the cases ${\beta^{*}_{T_{{\bf{x}}_{\CMcal{U}}}}} = {{\beta}_0}$ and ${\beta^{*}_{T_{{\bf{x}}_{\CMcal{U}}}}} = {{\beta}_1}$ are equivalent (by swapping $0$'s and $1$'s), so \emph{we only have two representative cases:} ${\beta^{*}_{T_{{\bf{x}}_{\CMcal{U}}}}} = {{\beta}_a}$, $a \in \{0,1\}$, and ${\beta^{*}_{T_{{\bf{x}}_{\CMcal{U}}}}} = {{\beta}_p}$.

\subsubsection{Representative Case 1:}
When ${\beta^{*}_{T_{{\bf{x}}_{\CMcal{U}}}}} = {{\beta}_a}$, $a \in \{0,1\}$, the allowed ranges of all ${ {{P}(\bf{x})}{{{\tilde{D}}_{a}}({\bf{x}})} }$ pairs are imposed by an \emph{additional upper limit} ${ {{P}({{\bf{x}}^{a}})} { {{{{\tilde{D}_a}}({{\bf{x}}^{a}})}_{\text{min}}} } }$ caused by privacy constraints. 
In these cases, effective upper limits are the minimum of the original upper limits ${{P}({{\bf{x}}})}{{{{\tilde{D}_a}}({\bf{x}})}_{\text{max}}}$ and the threshold, formally, ${{P}({{\bf{x}}})}{{{{\tilde{D}_a}}({\bf{x}})}_{\text{max}'}} = \min \{ {{P}({{\bf{x}}})}{{{{\tilde{D}_a}}({\bf{x}})}_{\text{max}}}, { {{P}({{\bf{x}}^{a}})} {{{{\tilde{D}_a}}({{\bf{x}}^{a}})}_{\text{min}}} } \}$. 
%
According to Theorem \ref{thm:feasible_condition_with_fidelity}, when ${\beta^{*}_{T_{{\bf{x}}_{\CMcal{U}}}}} = {{\beta}_a}$, the corresponding optimal-privacy solution is simply the effective upper limit ${{{\tilde{D}}_{a}}({\bf{x}})} = {{{{\tilde{D}_a}}({\bf{x}})}_{\text{max}'}}$, $\forall {\bf{x}} \in {T_{{\bf{x}}_{\CMcal{U}}}}$.

\begin{figure}[!t]
\centering
\includegraphics[width=3.4in]{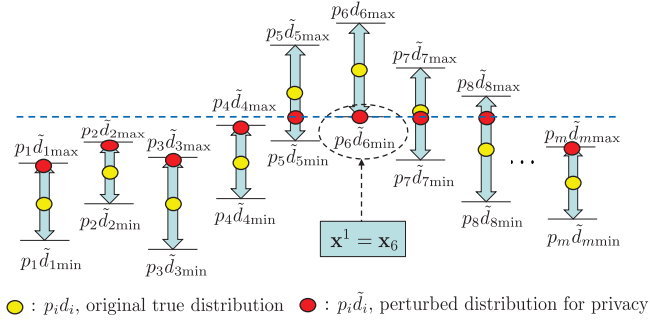}
\caption{An representative illustration for changes of joint probabilities caused by the optimal-privacy scheme.}
\label{fig:insights}
\end{figure}

An illustration that aids in understanding the intuition behind Theorem \ref{thm:feasible_condition_with_fidelity} is shown in Fig. \ref{fig:insights}, in which joint probabilities for $a=1$ and $\forall {\bf{x}} \in {T_{{\bf{x}}_{\CMcal{U}}}}$ are depicted for the case ${\beta^{*}_{T_{{\bf{x}}_{\CMcal{U}}}}} = {{\beta}_1}$.
For conciseness, let $m$ denote $\left| {T_{{\bf{x}}_{\CMcal{U}}}} \right|$\footnote{Since ${T_{{\bf{x}}_{\CMcal{U}}}}$ is the range of $( {{\bf{x}}_{\CMcal{U}}}, {X_{\CMcal{S}}} )$, $m$ represents the number of distinct sensitive records in the tuple.}, $p_i = P({{\bf{x}}_i})$, and ${\tilde{d}}_i = {\tilde{d}}({{\bf{x}}_i}) = {{{\tilde{D}_{1}}}({\bf{x}}_i)}$, $\forall i = 1, \ldots, m$. 
The yellow spots in Fig. \ref{fig:insights} denote the true joint probabilities ${p_i}{d_i}$, and the regions indicated by grey arrows interpret the allowed perturbation ranges $[{p_i}{{\tilde{d}}_{i\text{min}}}, {p_i}{{\tilde{d}}_{i\text{max}}}]$. 
In this example, the maximum of the lower limits is ${p_6}{{\tilde{d}}_{6\text{min}}}$, i.e., ${{\bf{x}}^{1}} = {{\bf{x}}_{6}}$. 
The value ${p_6}{{\tilde{d}}_{6\text{min}}}$ serves as a threshold (the blue dash line), imposing an upper limit on all perturbation ranges. 
The output of the optimal-privacy solution is denoted by the red spots, which take values from the effective upper bounds $\min \{ {{p_i}{{\tilde{d}}_{i\text{max}}}}, {{p_6}{{\tilde{d}}_{6\text{min}}}} \}$, $\forall i$. 
Here we get a clear insight into the optimal-privacy solutions for a QID sub-group ${T_{\{ {\bf{x}}_{\CMcal{U}},a \}}}$: the optimality is achieved by \emph{flattening the joint distribution ${ {{P}(\bf{x})}{{{\tilde{D}}_{a}}({\bf{x}})} }$ over all ${\bf{x}} \in {T_{{\bf{x}}_{\CMcal{U}}}}$ as much as possible} subject to the allowed perturbation range imposed by fidelity constraints. 
By flattening the joint distribution, the (Bayesian) posterior distribution over distinct sensitive values seen by an adversary becomes more uniform, and hence the maximal inference confidence is reduced.

Define $\overline{a}$ the complement of $a$, e.g., $\overline{a}=0$ if $a=1$. 
For binary decisions, the optimal-privacy scheme for a QID group ${T_{{\bf{x}}_{\CMcal{U}}}}$ is also privacy optimal for a sub-group ${T_{\{ {\bf{x}}_{\CMcal{U}},a \}}}$ when the joint distribution of ${T_{\{ {\bf{x}}_{\CMcal{U}}, \overline{a} \}}}$ is \emph{much} \quotes{flatter} (more private) than ${T_{\{ {\bf{x}}_{\CMcal{U}}, a \}}}$. 
In other words, the optimal-privacy solution flattens the least private distribution as much as possible; although this might influence the other (the much more private) one and cause it to be less private\footnote{Based on \eqref{eq:sub_eq_constraints}, any changes made to ${{{\tilde{D}}_{a}}({\bf{x}})}$ will also change ${{{\tilde{D}}_{\overline{a}}}({\bf{x}})}$.}, as long as its maximal inference confidence is less than ${{\beta}_a}$, the optimal privacy for the entire QID group 
is dominated by ${{\beta}_a}$, and thus the optimal privacy scheme for the sub-group ${T_{\{ {\bf{x}}_{\CMcal{U}},a \}}}$ is the optimal privacy scheme for the entire QID group.

\subsubsection{Representative Case 2:} 
When neither distribution is \emph{much} flatter than the other one, 
making one sub-group highly private causes the other one's privacy to degrade, i.e., none of the optimal schemes for any QID sub-group can be optimal for the entire QID group. 
In such a case, both sub-groups need to find a \quotes{balanced point} at which both sub-groups are equally private. 
Such a balanced privacy value for the maximal inference confidence for two sub-groups is denoted by ${{\beta}_p}$ in Theorem \ref{thm:feasible_condition_with_fidelity}, representing the optimal privacy for the QID group. 
As shown in Theorem \ref{thm:feasible_condition_with_fidelity}, in general, we have multiple solutions to achieve this balanced privacy value.
%
This is because, in such a case, the optimality of privacy is guaranteed if (i) ${{\tilde{D}_a}}({{\bf{x}}^{a}}) = {{{{\tilde{D}_a}}({{\bf{x}}^{a}})}_{\text{min}}}$, $\forall a$, and (ii) the following two equalities hold
\begin{align}
\textstyle\sum_{{\bf{x}}} {P({\bf{x}})}{{\tilde{D}_0}}({\bf{x}}) &= \textstyle\frac{1}{{\beta}_{p}} {{P}({{\bf{x}}^{0}})} {{{{\tilde{D}_0}}({{\bf{x}}^{0}})}_{\text{min}}}, \label{eq:balenced_0} \\
\textstyle\sum_{{\bf{x}}} {P({\bf{x}})}{{\tilde{D}_1}}({\bf{x}}) &= \textstyle\frac{1}{{\beta}_{p}} {{P}({{\bf{x}}^{1}})} {{{{\tilde{D}_1}}({{\bf{x}}^{1}})}_{\text{min}}}, \label{eq:balenced_1}
\end{align}
While in the following, we show that the above two equalities are equivalent, i.e., one implies the other. 

\begin{cor}
\label{cor:eq_eqv}
When ${\beta^{*}_{T_{{\bf{x}}_{\CMcal{U}}}}} = {\beta}_{p}$, \eqref{eq:balenced_0} implies \eqref{eq:balenced_1}, and vice versa.
\end{cor}
\begin{proof}
Recall ${\beta}_{p}$ from Theorem \ref{thm:feasible_condition_with_fidelity}, we have
\begin{equation}
\label{eq:beta_p}
{ {\textstyle\sum_{{\bf{x}}}} {{P}(\bf{x})} } = \textstyle\frac{1}{{{\beta}_p}} {{P}({{\bf{x}}^{1}})} {{{{\tilde{D}_1}}({{\bf{x}}^{1}})}_{\text{min}}} + \textstyle\frac{1}{{{\beta}_p}} {{P}({{\bf{x}}^{0}})} {{{{\tilde{D}_0}}({{\bf{x}}^{0}})}_{\text{min}}}.
\end{equation}
Since ${{\tilde{D}_0}}({\bf{x}}) + {{\tilde{D}_1}}({\bf{x}}) = 1$, subtract \eqref{eq:balenced_0} from \eqref{eq:beta_p}, we obtain \eqref{eq:balenced_1}. Similarly, subtract \eqref{eq:balenced_1} from \eqref{eq:beta_p}, we obtain \eqref{eq:balenced_0}.
\end{proof}

Therefore, to compute an optimal solution when ${\beta^{*}_{T_{{\bf{x}}_{\CMcal{U}}}}} = {\beta}_{p}$, we only need to solve \eqref{eq:balenced_1}. Since ${{\tilde{D}_0}}({\bf{x}}) + {{\tilde{D}_1}}({\bf{x}}) = 1$, and ${{\tilde{D}_a}}({{\bf{x}}^{a}}) = {{{{\tilde{D}_a}}({{\bf{x}}^{a}})}_{\text{min}}}$, $\forall a$, in general we only have $m-2$ variables (see Remark \ref{rmk:single_dominant}), and based on \eqref{eq:beta_p}, equality \eqref{eq:balenced_1} is equivalent to
\begin{gather}
\label{eq:balenced_1_eqv} 
\sum_{\substack{{\bf{x}} \neq {\bf{x}}^{0}, {\bf{x}}^{1}}} {P({\bf{x}})}{{\tilde{D}_1}}({\bf{x}}) = \big( \textstyle\frac{1-2{{\beta}_p}}{{{\beta}_p}} \big) {{P}({{\bf{x}}^{1}})} {{{{\tilde{D}_1}}({{\bf{x}}^{1}})}_{\text{min}}} - b({{\bf{x}}^{0}}). 
\end{gather}
When ${\beta^{*}_{T_{{\bf{x}}_{\CMcal{U}}}}} = {\beta}_{p}$, the right-hand-side (RHS) of \eqref{eq:balenced_1_eqv} is strictly bounded by $\big[\sum_{{\bf{x}} \neq {\bf{x}}^{0}, {\bf{x}}^{1}} {P({\bf{x}})}{{{{\tilde{D}_1}}({\bf{x}})}_{\text{min}'}},$ $\sum_{{\bf{x}} \neq {\bf{x}}^{0}, {\bf{x}}^{1}} {P({\bf{x}})}{{{{\tilde{D}_1}}({\bf{x}})}_{\text{max}'}}\big]$
, which implies there always exists a feasible solution for \eqref{eq:balenced_1}. When $m > 3$, since the number of variables to solve ($m-2$) is greater than the number of equation (one, which is \eqref{eq:balenced_1_eqv}), an optimal solution, in general, is not unique.

\begin{rmk}
\label{rmk:single_dominant}
For the special case ${{\bf{x}}^{0}} = {{\bf{x}}^{1}}$, we have $m-1$ variables. Such a case can happen when the population of a certain record dominates its corresponding QID group. When this is the case, the prior (distribution) knowledge provides very high (baseline) confidence on inferring this record. In particular, for such a case, we must have ${{\beta}_p} = {{\beta}_{\text{min}}}$. If ${{\beta}_p} > {{\beta}_a}$, $\forall a$, i.e., ${\beta^{*}_{T_{{\bf{x}}_{\CMcal{U}}}}} = {{\beta}_p}$, this becomes trivial: according to Lemma \ref{lem:feasible_condition_no_fidelity} and its following discussion, the announced ATR can only provide trivial information to achieve this lowest-possible baseline confidence.
\end{rmk}

\section{Numerical Examples}\label{sec:results}

\begin{table*}[t]
\centering
\caption{Detailed Inputs and Computations of the Provided Numerical Example}
\vspace*{-3mm}
\begin{minipage}{\textwidth} 
\label{table:computations}
\setlength\tabcolsep{1pt} 
\renewcommand{\arraystretch}{0.8}
\begin{tabular}{|>{\small}c|>{\tiny}c|>{\tiny}c|>{\tiny}c|>{\tiny}c|>{\tiny}c|>{\tiny}c|>{\tiny}c|>{\tiny}c|>{\tiny}c|>{\tiny}c|>{\tiny}c|>{\tiny}c|>{\tiny}c|}
\hline
 & \multicolumn{7}{>{\small}c|}{Inputs} & \multicolumn{6}{>{\small}c|}{Computations}\\                                                                                             \hline
${\bf{x}}$ & \miniscule{${{P}(\bf{x})}$} & \miniscule{${{{{D}_1}}({\bf{x}})}$} & \miniscule{${{{{D}_0}}({\bf{x}})}$} & \miniscule{${{{{\tilde{D}_1}}({\bf{x}})}_{\text{min}}}$} & \miniscule{${{{{\tilde{D}_1}}({\bf{x}})}_{\text{max}}}$} & \miniscule{${{{{\tilde{D}_0}}({\bf{x}})}_{\text{min}}}$} & \miniscule{${{{{\tilde{D}_0}}({\bf{x}})}_{\text{max}}}$} & \miniscule{${{P}(\bf{x})}{{{{\tilde{D}_1}}({\bf{x}})}_{\text{min}}}$} & \miniscule{${{P}(\bf{x})}{{{{\tilde{D}_1}}({\bf{x}})}_{\text{max}}}$} & \miniscule{${{P}(\bf{x})}{{{{\tilde{D}_1}}({\bf{x}})}_{\text{max}'}}$} & \miniscule{${{P}(\bf{x})}{{{{\tilde{D}_0}}({\bf{x}})}_{\text{min}}}$} & \miniscule{${{P}(\bf{x})}{{{{\tilde{D}_0}}({\bf{x}})}_{\text{max}}}$} & \miniscule{${{P}(\bf{x})}{{{{\tilde{D}_0}}({\bf{x}})}_{\text{max}'}}$} \\ 
\hline
${\bf{x}}_1$   & $0.3$   & $0$   & $1$    & $0$   & $0.1$ & $0.9$ & $1$   & $0$      & $0.03$   &
 $0.03$   & $0.27$   & $0.3$    & $0.27$   \\ \hline                                            
${\bf{x}}_2$   & $0.125$ & $0$   & $1$    & $0$   & $0.1$ & $0.9$ & $1$   & $0$      & $0.0125$ &
 $0.0125$ & $0.1125$ & $0.125$  & $0.125$  \\ \hline                                              
${\bf{x}}_3$   & $0.075$ & $1$   & $0$    & $0.9$ & $1$   & $0$   & $0.1$ & $0.0675$ & $0.075$  &
 $0.0675$ & $0$      & $0.0075$ & $0.0075$ \\ \hline                                              
${\bf{x}}_4$   & $0.225$ & $0$   & $1$    & $0$   & $0.1$ & $0.9$ & $1$   & $0$      & $0.0225$ &
 $0.0225$ & $0.2025$ & $0.225$  & $0.2025$  \\ \hline                                            
${\bf{x}}_5$   & $0.175$ & $0.5$ & $0.5$  & $0.4$ & $0.6$ & $0.4$ & $0.6$ & $0.07$   & $0.105$  &
 $0.09$   & $0.07$   & $0.105$  & $0.105$  \\ \hline                                              
${\bf{x}}_6$   & $0.1$   & $1$   & $0$    & $0.9$ & $1$   & $0$   & $0.1$ & $0.09$   & $0.1$    &
 $0.09$   & $0$      & $0.01$   & $0.01$   \\ \hline                                              
\end{tabular}
\end{minipage}
\end{table*}

In the previous section, we proposed lemmas characterizing important properties about privacy-fidelity trade-off, a theorem providing closed-form optimal solutions for the trade-off problem, and insights into the optimal solutions for both the representative cases.
In this section, we provide numerical examples to demonstrate \emph{privacy-fidelity trade-off regions}, and aid in understanding the properties of the trade-off regions and the insights into the optimal solutions for both the representative cases. 
Without loss of generality, we reuse the same examples demonstrated in Section \ref{sec:problem} showcasing how the proposed linear-time optimal-privacy scheme (Algorithm \ref{alg:opt_privacy}) can be applied in practice to solve the problem, as long as there is no privacy preference among sensitive attribute values.

Consider Table \ref{table:scenario} again but for a smaller size population \{12, 5, 3, 9, 7, 4\} (first column of the table) for ease of demonstration, and let ${\bf{x}}_i$ denote the record of the $i$-th row, $i = 1, \ldots, 6$. Suppose an announced ATR needs to satisfy a pre-determined fidelity constraint $\delta = 0.9$ ($90\%$-fidelity), and we would like to preserve data subjects' privacy as much as possible subject to the fidelity constraint.

First, consider the QID group of female  ${T_{{{\bf{x}}_{\CMcal{U}}}=\text{\{F\}}}}$, i.e., the tuple of records ${T_{\text{\{F\}}}}$ = \{${\bf{x}}_1$, ${\bf{x}}_2$, ${\bf{x}}_3$\}. 
Based on lines 1 to 4 in Algorithm \ref{alg:opt_privacy}, we first need to determine ${{\bf{x}}^{a}}$ and ${{{{\tilde{D}_a}}({\bf{x}})}_{\text{max}'}}$ $\forall a \in \{0,1\}$, $ \forall {\bf{x}} \in {T_{\text{\{F\}}}}$. Detailed computations are demonstrated in Remark \ref{rmk:numerical_example_computations}, and the computed results are presented in Table \ref{table:computations}; from which, we observe that ${{\bf{x}}^{1}} = {\bf{x}}_3$ and ${{\bf{x}}^{0}} = {\bf{x}}_1$ (see Remark \ref{rmk:numerical_example_computations} for details as well).

Proceeding to line 5, we compute $\beta_0$, $\beta_1$, and $\beta_p$ as follows:
\begin{flalign*}
{{\beta}_1} &= \frac{ {{P}({{\bf{x}}^{1}})} {{{{\tilde{D}_1}}({{\bf{x}}^{1}})}_{\text{min}}} }{ { {{P}({{\bf{x}}^{1}})} {{{{\tilde{D}_1}}({{\bf{x}}^{1}})}_{\text{min}}} } + {\sum_{{{\bf{x}} \neq {{\bf{x}}^{1}}}, {{\bf{x}} \in {T_{{\bf{x}}_{\CMcal{U}}}}}}} {{{P}(\bf{x})} {{{{\tilde{D}_1}}({\bf{x}})}_{\text{max}'}}} } = \frac{0.0675}{0.0675+0.03+0.0125} \approx 0.6136 \text{,} \\
\end{flalign*}
\begin{flalign*}
{{\beta}_0} &= \frac{ {{P}({{\bf{x}}^{0}})} {{{{\tilde{D}_0}}({{\bf{x}}^{0}})}_{\text{min}}} }{ { {{P}({{\bf{x}}^{0}})} {{{{\tilde{D}_0}}({{\bf{x}}^{0}})}_{\text{min}}} } + {\sum_{{{\bf{x}} \neq {{\bf{x}}^{0}}}, {{\bf{x}} \in {T_{{\bf{x}}_{\CMcal{U}}}}}}} {{{P}(\bf{x})} {{{{\tilde{D}_0}}({\bf{x}})}_{\text{max}'}}} } = \frac{0.27}{0.27+0.125+0.0075} \approx 0.6708 \text{,} \\
{{\beta}_p} &= \frac{ {{P}({{\bf{x}}^{1}})} {{{{\tilde{D}_1}}({{\bf{x}}^{1}})}_{\text{min}}} + {{P}({{\bf{x}}^{0}})} {{{{\tilde{D}_0}}({{\bf{x}}^{0}})}_{\text{min}}} }{ {\sum_{{\bf{x}} \in {T_{{\bf{x}}_{\CMcal{U}}}}}}{{P}(\bf{x})} } = \frac{0.0675+0.27}{0.5} = 0.675 \text{,}
\end{flalign*}
and obtain ${\beta^{*}_{T_{\text{\{F\}}}}} \triangleq \max \{ {{\beta}_0},{{\beta}_1},{{\beta}_p} \} = {{\beta}_p} = 0.675$. Proceeding to lines 12 and 13, in this case we need to call function \textproc{Allocation} in line 17. Based on lines 18 and 19, we first need to compute
\begin{flalign*}
\begin{split}
{{{{\tilde{D}_1}}({\bf{x}}_2)}_{\text{min}'}} = &{\textstyle\frac{1}{{P}({{\bf{x}}_2})}} \max \{ {{P}({{\bf{x}}_2})}{{{{\tilde{D}_1}}({\bf{x}}_2)}_{\text{min}}}, { {{P}({{\bf{x}}^{1}})} {{{{\tilde{D}_1}}({{\bf{x}}^{1}})}_{\text{min}}} } + b({\bf{x}}_2)\} \\
 = &{\frac{1}{0.125}} \max \{ 0, 0.0675 + 0.125 - (0.675)(0.5)\} = 0 \text{.}
\end{split}
\end{flalign*}
Proceeding to line 20, since ${{{{\tilde{D}_1}}({\bf{x}}_2)}_{\text{min}'}} = 0$, we have
\begin{equation*}
\begin{split}
\vars{resid} = \text{RHS of \eqref{eq:balenced_1_eqv}} &= \big( \textstyle\frac{1-2{{\beta}_p}}{{{\beta}_p}} \big) {{P}({{\bf{x}}^{1}})} {{{{\tilde{D}_1}}({{\bf{x}}^{1}})}_{\text{min}}} - b({{\bf{x}}^{0}}) \\
&= (\textstyle\frac{-0.35}{0.675})(0.075)(0.9) + (0.675)(0.5) - 0.3 = 0.0025 \text{.}
\end{split}
\end{equation*}
Based on lines 21 and 22, we obtain 
\begin{equation*}
\begin{split}
&{{\tilde{D}_1}}({{\bf{x}}_{3}}) = {{\tilde{D}_1}}({{\bf{x}}^{1}}) = {{{{\tilde{D}_1}}({{\bf{x}}^{1}})}_{\text{min}}} = {{{{\tilde{D}_1}}({{\bf{x}}_{3}})}_{\text{min}}} = 0.9 \text{,} \\
&{{\tilde{D}_1}}({{\bf{x}}_{1}}) = {{\tilde{D}_1}}({{\bf{x}}^{0}}) = 1 - {{{{\tilde{D}_0}}({{\bf{x}}^{0}})}_{\text{min}}} = 1 - {{{{\tilde{D}_0}}({{\bf{x}}_{1}})}_{\text{min}}} = 0.1 \text{.}
\end{split}
\end{equation*}
Moreover, proceeding to lines 23 to 27, we obtain 
\begin{equation*}
\begin{split}
&\vars{capacity} = {{{{\tilde{D}_1}}({\bf{x}}_2)}_{\text{max}'}} - {{{{\tilde{D}_1}}({\bf{x}}_2)}_{\text{min}'}} = \textstyle\frac{0.0125}{0.125} - 0 = 0.1 \text{,} \\
&\vars{allocation} = \min\{ \textstyle\frac{0.0025}{0.125}, 0.1\} = 0.02 \text{,} \\
&{{\tilde{D}_1}}({\bf{x}}_2) = {{{{\tilde{D}_1}}({\bf{x}}_2)}_{\text{min}'}} + \vars{allocation} = 0 + 0.02 = 0.02 \text{.}
\end{split}
\end{equation*}

We therefore obtain the optimal solution ${{\tilde{D}_1}}(\{{\bf{x}}_1, {\bf{x}}_2, {\bf{x}}_3\}) = [0.1, 0.02, 0.9]$ for the QID group of female, which yields maximum confidence of 67.5\% for an adversary inferring any sensitive information from any female data subject.

We then consider the QID group for male ${T_{{{\bf{x}}_{\CMcal{U}}}=\text{\{M\}}}}$, i.e., the tuple of records ${T_{\text{\{M\}}}}$ = \{${\bf{x}}_4$, ${\bf{x}}_5$, ${\bf{x}}_6$\}. Similarly, based on Table \ref{table:computations}, we obtain ${{\bf{x}}^{1}} = {\bf{x}}_6$, ${{\bf{x}}^{0}} = {\bf{x}}_4$, and
\begin{flalign*}
{{\beta}_1} &= \frac{0.09}{0.0225+0.09+0.09} \approx 0.4444 \text{,} \\
{{\beta}_0} &= \frac{0.2025}{0.2025+0.105+0.01} \approx 0.6378 \text{,} \\
{{\beta}_p} &= \frac{0.09+0.2025}{0.5} = 0.585 \text{,}
\end{flalign*}
and we get ${\beta^{*}_{T_{\text{\{M\}}}}} = \max \{ {{\beta}_0},{{\beta}_1},{{\beta}_p} \} = {{\beta}_0} \approx 0.6378$. Based on lines 6 to 8, we obtain the optimal solution for this group ${{\tilde{D}_1}}(\{{\bf{x}}_4, {\bf{x}}_5, {\bf{x}}_6\}) = 1 - {{{{\tilde{D}_0}}(\{{\bf{x}}_4, {\bf{x}}_5, {\bf{x}}_6\})}_{\text{max}'}} = 1 - [\textstyle\frac{0.2025}{0.225}, \textstyle\frac{0.105}{0.175}, \textstyle\frac{0.01}{0.1}] = [0.1, 0.4, 0.9]$ for QID group of male, which yields maximum confidence of 63.78\% for an adversary inferring any sensitive information from any male data subject. Based on Lemma \ref{lem:Separability_OPT}, the optimal-privacy $\beta^{*}$ for the entire dataset is $\max\{ {\beta^{*}_{T_{\text{\{F\}}}}}, {\beta^{*}_{T_{\text{\{M\}}}}} \} = \max\{0.675, 0.6378\} = 0.675$, which is the maximum confidence for an adversary inferring any sensitive information from any data subject from this dataset, based on the announced ATR.

The optimal solution for the QID group of female is a \quotes{balanced point} between inferring the annual income of ${\bf{x}}_1$ and ${\bf{x}}_3$ correctly, i.e., $\mathit{Conf}({\text{F}},A \to {\text{Annual Income}}) = \mathit{conf}({\text{F}},A=0 \to {<100\text{k}}) = \mathit{conf}({\text{F}},A=1 \to {>200\text{k}})$, where $A$ is defined in Table \ref{table:notation}, the random variable of decision outcome (0: negative; 1: positive), and
\begin{equation*}
\begin{split}
&\mathit{conf}({\text{F}},0 \to {<100\text{k}}) = { \frac{ {{P}({\bf{x}_1})}{{{\tilde{D}}_{0}}({\bf{x}_1})} }{ {{P}({\bf{x}_1})}{{{\tilde{D}}_{0}}({\bf{x}_1})} + {{P}({\bf{x}_2})}{{{\tilde{D}}_{0}}({\bf{x}_2})} + {{P}({\bf{x}_3})}{{{\tilde{D}}_{0}}({\bf{x}_3})} } } =  \textstyle\frac{0.3 \times 0.9}{0.3 \times 0.9 + 0.125 \times 0.98 + 0.075 \times 0.1} = 0.675 \text{,} \\
&\mathit{conf}({\text{F}},1 \to {>200\text{k}}) = { \frac{ {{P}({\bf{x}_3})}{{{\tilde{D}}_{1}}({\bf{x}_3})} }{ {{P}({\bf{x}_1})}{{{\tilde{D}}_{1}}({\bf{x}_1})} + {{P}({\bf{x}_2})}{{{\tilde{D}}_{1}}({\bf{x}_2})} + {{P}({\bf{x}_3})}{{{\tilde{D}}_{1}}({\bf{x}_3})} } } = \textstyle\frac{0.075 \times 0.9}{0.3 \times 0.1 + 0.125 \times 0.02 + 0.075 \times 0.9} = 0.675 \text{.}
\end{split}
\end{equation*}
Making either inference more private will cause the other one less private and hence degrades the overall privacy guarantee as discussed in Section \ref{subsec:insights}. In contrast, the optimal solution for the male group tries to minimize the confidence of correctly inferring the annual income of ${\bf{x}}_4$, i.e., $\mathit{Conf}({\text{M}},A \to {\text{Annual Income}}) = \mathit{conf}({\text{M}},A=0 \to {<100\text{k}})$, and
\begin{equation*}
\begin{split}
&\mathit{conf}({\text{M}},0 \to {<100\text{k}}) = { \frac{ {{P}({\bf{x}_4})}{{{\tilde{D}}_{0}}({\bf{x}_4})} }{ {{P}({\bf{x}_4})}{{{\tilde{D}}_{0}}({\bf{x}_4})} + {{P}({\bf{x}_5})}{{{\tilde{D}}_{0}}({\bf{x}_5})} + {{P}({\bf{x}_6})}{{{\tilde{D}}_{0}}({\bf{x}_6})} } } = \textstyle\frac{0.225 \times 0.9}{0.225 \times 0.9 + 0.175 \times 0.6 + 0.1 \times 0.1} \approx 0.6378 \text{.}
\end{split}
\end{equation*}
From the above equation, it is not hard to see that the optimal solution maximizes the denominator while minimizing the numerator in order to minimize the ratio for optimal privacy.

\begin{figure}
     \centering
     \begin{subfigure}[b]{0.45\textwidth}
         \centering
         \caption{P-F Tradeoff for the QID Group of Female}
         \includegraphics[width=\textwidth]{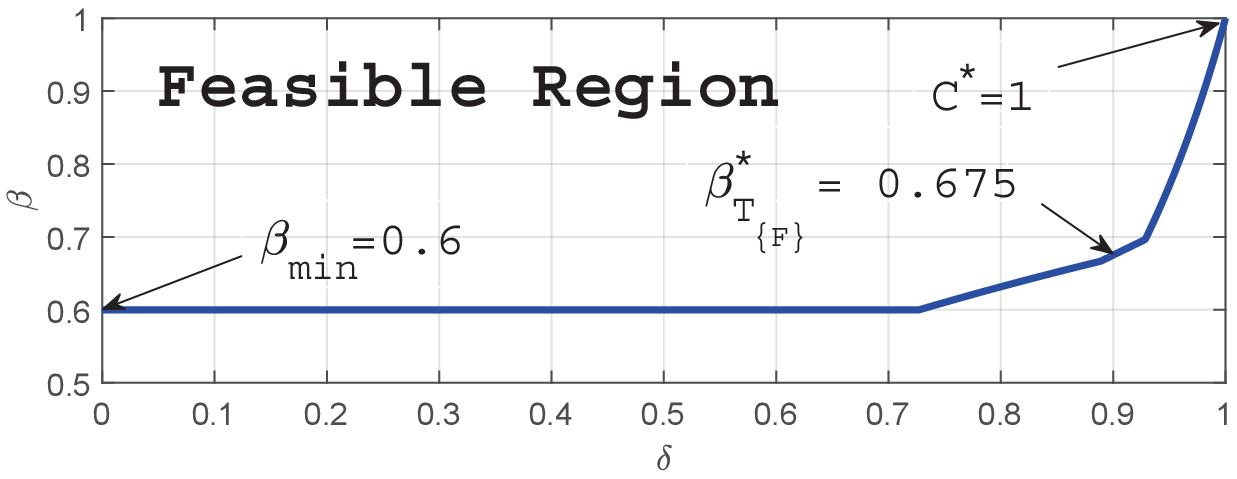}
         \vspace*{-3mm}
         \label{fig:privacy_fidelity_tradeoff_F}
     \end{subfigure}
     \begin{subfigure}[b]{0.45\textwidth}
         \centering
         \caption{P-F Tradeoff for the QID Group of Male}
         \includegraphics[width=\textwidth]{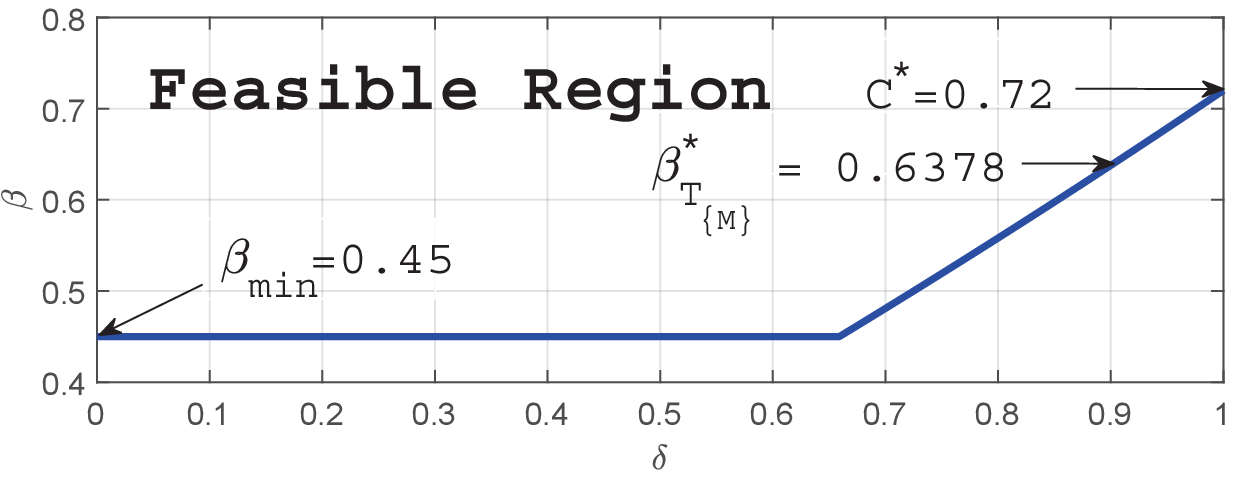}
         \vspace*{-3mm}
         \label{fig:privacy_fidelity_tradeoff_M}
     \end{subfigure}
        \vspace*{-4mm}
        \caption{Privacy-Fidelity (P-F) Tradeoffs for QID Groups of Female and Male}
        \label{fig:privacy_fidelity_tradeoffs}
\vspace{-1mm}
\end{figure}

From the above example, we demonstrated that subject to fidelity constraints, how an optimal-privacy ATR can be obtained efficiently using Algorithm \ref{alg:opt_privacy}. 
The maximum confidence of an adversary, which is $\mathit{conf}({\text{F}},1 \to {>200\text{k}})$, drops from 100\% to 67.5\% by setting a 10\%-distortion tolerance for the announced ATR. 

Figure \ref{fig:privacy_fidelity_tradeoffs} depicts the privacy-fidelity tradeoffs for both QID groups in this numerical example. Given any fidelity requirement $\delta$, the optimal privacy (i.e., the smallest possible $\beta$) that we can achieve is the boundary of the trade-off region (the blue curve).
%
%
The tradeoff region for $\beta$, as discussed in Section \ref{subsec:solution_prop}, should be within the range $[{\beta}_{min}, C^{*}]$, which can be easily computed based on Definition \ref{def:max_conf_tuple} and Lemma \ref{lem:feasible_condition_no_fidelity}: \\ \vspace{-1em} \\
For the QID group of female: $[{\beta}_{min}, C^{*}] =$\\
\phantom{xxxx}$[\textstyle\frac{P({\bf{x}_1})}{P({\bf{x}_1})+P({\bf{x}_2})+P({\bf{x}_3})}, \textstyle\frac{P({\bf{x}_1}){{D_1}({\bf{x}_1})}}{{P({\bf{x}_1}){{D_1}({\bf{x}_1})}}+{P({\bf{x}_2}){{D_1}({\bf{x}_2})}}+{P({\bf{x}_3}){{D_1}({\bf{x}_3})}}}] = [\textstyle\frac{0.3}{0.5}, \textstyle\frac{0.075 \times 1}{0.075 \times 1}] = [0.6, 1]$.\\ \vspace{-0.5em} \\
For the QID group of male: $[{\beta}_{min}, C^{*}] =$\\
\phantom{xxxx}$[\textstyle\frac{P({\bf{x}_4})}{P({\bf{x}_4})+P({\bf{x}_5})+P({\bf{x}_6})}, \textstyle\frac{P({\bf{x}_4}){{D_1}({\bf{x}_4})}}{{P({\bf{x}_4}){{D_1}({\bf{x}_4})}}+{P({\bf{x}_5}){{D_1}({\bf{x}_5})}}+{P({\bf{x}_6}){{D_1}({\bf{x}_6})}}}] = [\textstyle\frac{0.225}{0.5}, \textstyle\frac{0.225 \times 1}{0.225 \times 1 + 0.175 \times 0.5}] = [0.45, 0.72]$. \\ \vspace{-0.5em} \\
Both results show consistency with Figure \ref{fig:privacy_fidelity_tradeoffs} : in Figure \ref{fig:privacy_fidelity_tradeoff_F}, the range of $\beta$ is within [0.6, 1]; in Figure \ref{fig:privacy_fidelity_tradeoff_M}, the range of $\beta$ is within [0.45, 0.72]. Note that based on Lemma \ref{lem:feasible_condition_no_fidelity}, any privacy requirement with $\beta < \max\{0.6,0.45\} = 0.6$ is not feasible for this dataset, and based on Lemma \ref{lem:optimal_utility_condition}, any privacy requirement with $\beta > 0.72$ can have 1-fidelity solution for the QID group of male, i.e., no perturbation is needed. How much fidelity should be sacrificed in order to achieve a certain level of privacy can thus be known based on the tradeoff curves. 

\begin{rmk}
\label{rmk:numerical_example_computations}
Here we demonstrate how the values presented in Table \ref{table:computations} are computed. Note that, we only demonstrate the computation of values in the first row (i.e., for $\bf{x}_1$), as computations for values in all other rows (for all other $\bf{x}_i$'s) follow similar steps. In the following, we start from the left-most value and then move to the right.

For $\bf{x} = \bf{x}_1$, ${{\bf{x}}_{\CMcal{U}}} =$ \{F\} (Female), and since the total population is $12 + 5 + 3 + 9 + 7 + 4 = 40$, ${{P}(\bf{x})} = 12/40 = 0.3$. Based on Table \ref{table:scenario}, since the decision rule represents the probability of receiving a positive decision, ${{{{D}_1}}({\bf{x}})}$ is basically the decision rule in Table \ref{table:scenario}, and ${{{{D}_0}}({\bf{x}})}$ is simply $1-{{{{D}_1}}({\bf{x}})}$. The pre-defined fidelity parameter $\delta$ is $0.9$, i.e., the announced decision mapping ${{{{\tilde{D}_a}}({\bf{x}})}}$ can differ from the true decision mapping ${{{{D}_a}}({\bf{x}})}$ by at most $10\%$, $\forall a=0,1$. Therefore, $\abs{{{{\tilde{D}_1}}({\bf{x}})} - 0} \le 0.1$, and we get ${{{{\tilde{D}_1}}({\bf{x}})}_{\text{min}}} = 0$ and ${{{{\tilde{D}_1}}({\bf{x}})}_{\text{max}}} = 0.1$. Similarly, $\abs{{{{\tilde{D}_0}}({\bf{x}})} - 1} \le 0.1$, and we get ${{{{\tilde{D}_0}}({\bf{x}})}_{\text{min}}} = 0.9$ and ${{{{\tilde{D}_0}}({\bf{x}})}_{\text{max}}} = 1$. The values of the above terms are based on their definitions (refer to Theorem \ref{thm:feasible_condition_with_fidelity}) and the input parameters. Since now we have values for ${{P}(\bf{x})}$, ${{{{\tilde{D}_1}}({\bf{x}})}_{\text{min}}}$, ${{{{\tilde{D}_1}}({\bf{x}})}_{\text{max}}}$, ${{{{\tilde{D}_0}}({\bf{x}})}_{\text{min}}}$, and ${{{{\tilde{D}_0}}({\bf{x}})}_{\text{max}}}$, the values for the terms ${{P}(\bf{x})}{{{{\tilde{D}_1}}({\bf{x}})}_{\text{min}}}$, ${{P}(\bf{x})}{{{{\tilde{D}_1}}({\bf{x}})}_{\text{max}}}$, ${{P}(\bf{x})}{{{{\tilde{D}_0}}({\bf{x}})}_{\text{min}}}$, and ${{P}(\bf{x})}{{{{\tilde{D}_0}}({\bf{x}})}_{\text{max}}}$ are just simple multiplications.

Next, we show how the values for the terms ${{P}(\bf{x})}{{{{\tilde{D}_1}}({\bf{x}})}_{\text{max}'}}$ (third column in the \quotes{Computations} category) and ${{P}(\bf{x})}{{{{\tilde{D}_0}}({\bf{x}})}_{\text{max}'}}$ (the last column) are obtained. 
Based on Theorem \ref{thm:feasible_condition_with_fidelity}, ${{{{\tilde{D}_a}}({\bf{x}})}_{\text{max}'}} \triangleq {\textstyle\frac{1}{{P}({{\bf{x}}})}} \min \{ {{P}({{\bf{x}}})}{{{{\tilde{D}_a}}({\bf{x}})}_{\text{max}}}, { {{P}({{\bf{x}}^{a}})} {{{{\tilde{D}_a}}({{\bf{x}}^{a}})}_{\text{min}}} } \}$, where ${{\bf{x}}^{a}} \triangleq {\argmax_{{ {{\bf{x}} \in {T_{{\bf{x}}_{\CMcal{U}}}}} }}} { {{P}(\bf{x})} {{{{\tilde{D}_a}}({\bf{x}})}_{\text{min}}} }$, $\forall a=0,1$, 
we thus have ${{\bf{x}}^{0}} \triangleq {\argmax_{{ {{\bf{x}} \in {T_{\text{\{F\}}}}} }}} { {{P}(\bf{x})} {{{{\tilde{D}_0}}({\bf{x}})}_{\text{min}}} } = {\argmax_{{ {{\bf{x}} \in {\{{\bf{x}_1}, {\bf{x}_2}, {\bf{x}_3}\}}} }}} \{ 0.27, 0.1125, 0 \} = {\bf{x}_1}$ and ${{\bf{x}}^{1}} = {\argmax_{{ {{\bf{x}} \in {\{{\bf{x}_1}, {\bf{x}_2}, {\bf{x}_3}\}}} }}} \{ 0, 0, 0.0675 \} = {\bf{x}_3}$. 
Hence, for $\bf{x} = \bf{x}_1$, ${{{{\tilde{D}_1}}({\bf{x}})}_{\text{max}'}} = {\textstyle\frac{1}{0.3}} \min \{ 0.03, 0.0675 \} = 0.03/0.3$ and ${{{{\tilde{D}_0}}({\bf{x}})}_{\text{max}'}} = {\textstyle\frac{1}{0.3}} \min \{ 0.3, 0.27 \} = 0.27/0.3$. 
Therefore, we obtain ${{P}(\bf{x})}{{{{\tilde{D}_1}}({\bf{x}})}_{\text{max}'}} = 0.03$ and ${{P}(\bf{x})}{{{{\tilde{D}_0}}({\bf{x}})}_{\text{max}'}} = 0.27$.
\end{rmk}

\section{Related Work}\label{sec:related}
There is a huge literature on transparency \cite{guidotti2018survey,dovsilovic2018explainable} and fairness \cite{mehrabi2019survey} for ML. \cite{lepri2018fair} provides a detailed survey on techniques proposed for enhancing transparency and fairness for ML models. However, the perspectives of transparency and fairness in ML models may not be completely in sync with those in algorithmic transparency, e.g., the philosophy of fairness in ML is to train fair ML models or algorithms, while the philosophy of fairness in accountable algorithmic transparency is to verify or to demonstrate whether the examined ML algorithms comply with certain fairness requirements.

There are a number of studies on transparency and fairness and several addressing privacy in data transparency, e.g., \cite{stahl2018ethics, reidenberg2018achieving, young2019beyond}; however, \emph{there is little effort in considering the potential impact on privacy brought on by algorithmic transparency schemes and/or fairness measures.} 
\cite{datta2015automated} provides transparency in the interaction between Google Ads, users' Ad privacy settings, and user behaviors, showing the disparate impact that female gender setting has (vs. male gender setting) on results, e.g., with fewer instances of ads related to high paying jobs; while whether users' privacy could be leaked from Google Ads or the associated transparency report is unclear without further investigation.
\cite{ananny2018seeing} investigates the limitations of transparency and its impact on society and notes that transparency can threaten privacy, \emph{but it is yet to be made clear what possible aspects of transparency can hurt privacy, and by what privacy-preserving techniques could remedy the situation.} 
%
Here, we show that data subjects' privacy can be leaked via various kinds of transparency schemes and fairness measures in an announced ATR and propose a privacy protection scheme yielding privacy preserving information on an ATR.
Motivated by transparency and fairness, \cite{ekstrand2018privacy} raises questions regarding fair privacy for all participating users, as it is considered discriminatory when different users are protected by different levels of privacy; however, \emph{
on what notion of privacy should be fair, by what methodology to protect such privacy and to achieve it fairly are still unclear.} 
However, in contrast, numerical examples in Section \ref{sec:results} show that \emph{the optimal privacy for different QID groups, subject to the same fidelity constraints, is in general different} due to the disparity of prior distributions, prior vulnerabilities \cite{m2012measuring}, side-information, and associated decision mapping between groups.
\cite{sloan2018algorithm} studies the problem of providing transparency to consumers while preserving information privacy for them, and proposes \emph{informational norms} to constrain the collection, use, and distribution of transparent information in role-appropriate manners to fulfill the goal. \emph{However, the definition of role-appropriate manners is yet to be more specific, and it is also unclear how fairness measures, which compare decision rules between two individuals or groups (likely in different roles), should be announced under such a norm.} 
%
In contrast, our privacy protection scheme does not make any assumptions on informational norm and does not rely on any norm (potentially hard to accomplish) to protect users' privacy, and thus can be applied generally.
In addition, informational norm may still not be adequate to protect users' privacy: individuals belonging to the same role may still be able to infer private information of others, e.g., in Table \ref{table:scenario}, any female credit card owner can infer other female credit card owners' income range.

There exist a couple of works using differential privacy (DP) to remedy the privacy leakage/attack issue in algorithmic transparency or model explanations.
A recent work \cite{shokri2020privacy} demonstrates membership inference attacks \cite{shokri2016membership} on training datasets of ML models by utilizing information from the corresponding featured-based model explanations (i.e., feature importance/interaction transparency schemes). 
To address this issue, in \cite{patel2020model}, DP is applied to the gradient descent algorithm for generating feature-based model explanations.
\cite{dat16}, arguably the only previous work that addresses transparency, fairness, and privacy in an accountable ATR, proposes a feature-based measure, named \emph{quantitative input influence} (QII); based on which, the authors propose public and personalized transparency reports, as well as a fairness measure, named \emph{group disparity}, to measure potential disparate impacts on different groups of people. 
DP is adopted to the above measures in order to prevent potential privacy leaks caused by the provided QII and group disparity in an announced ATR. 
\emph{However, applying DP solely does not result in prevention of inference attacks, in particular, attribute inference attacks: once strong correlations between attribute values are known, sensitive attribute values can be inferred no matter whether a privacy victim belongs to a specific dataset or not, and thus DP cannot help in such a scenario} (see \cite{dwork2014algorithmic}, Section 2.3.2, the \emph{smoking-cause-cancer} example). 
In light of this, here, we propose a privacy-preserving scheme to prevent attribute inference attacks by limiting the attribute inference confidence from public/known attribute values, via an announced ATR with assistance of side-information, to any data subjects' private attribute values.


\section{Summary}\label{sec:conclusion}
In this work, we demonstrated how a honest-but-curious adversary can utilize widely-available information provided in an algorithmic transparency report, to obtain data subjects' private information. 
From this we glean which potential aspects of transparency and fairness measures can hurt privacy. We then propose a privacy scheme that perturbs the information to be announced, to remedy the privacy leaks. 
We systematically study the impact of such perturbation on fairness measures and the fidelity of the announced information, formulated as an optimization for optimal privacy subject to fidelity constraints. 
To efficiently solve the optimization problem, we reveal important properties and provide closed-form solutions, based on which we propose a privacy protection scheme.
Given fidelity requirements, the proposed scheme can efficiently produce optimal-privacy ATRs in linear time. In addition, we provide insight into our proposed optimal privacy scheme.
Our proposed methodology is suited to more general problems beyond algorithmic transparency, where the release of the model information is controlled and the input data cannot be modified - an example being in the setting of model inversion attacks in \cite{fredrikson2014privacy} where the model owner has no authority to modify the input data (patents' clinical history and genomic data) but has the control of the amount of information about the (dose-suggesting) model to be released. In such a scenario, our scheme can help privately release information of a model to pharmacists for better understanding of suggesting personalized dosage.

\bibliographystyle{ACM-Reference-Format}
\bibliography{refs}

  \newpage
\noindent\textbf{\Large{APPENDIX}}

\appendix
\section{Fairness Measures}\label{sec:fairness}
Another important motivation of providing algorithmic transparency is to understand if a decision-making algorithm is fair. GDPR Article 5 regulation indicates that personal data should be processed fairly and in a transparent manner. Many researchers are committed to providing proper measures for fairness and making ML algorithms fair \cite{corbett2017algorithmic,dwo11,feldman2015certifying,zemel2013learning,kamishima2012fairness,kamiran2013quantifying,biddle2006adverse}. In general, there are two main categories of fairness: (i) individual fairness, and (ii) group fairness. Popular definitions of group fairness includes statistical parity (SP), conditional statistical parity (CSP), and p\%-rule (PR) (see Appendix \ref{sec:fairness} for detailed definitions).

\subsection{Measures for Individual Fairness}
\begin{defn}
\label{def:individual_fairness}
($({\mathfrak{D}},{\mathcal{D}})$-\emph{Individual Fairness} \cite{dwo11}) Given a distance measure ${\mathcal{D}}: {{\CMcal{R}}_{X}} \times {{\CMcal{R}}_{X}} \to {{\mathbb{R}}^{+}} \triangleq [0,\infty)$ on individuals' records, a decision mapping $D:{{\CMcal{R}}_{X}} \to \Delta({\CMcal{A}})$ satisfies \emph{individual fairness} if it complies with the $({\mathfrak{D}},{\mathcal{D}})$-Lipschitz property for every two individuals' records ${{\bf{x}}_1},{{\bf{x}}_2} \in {{\CMcal{R}}_{X}}$, i.e.,
\begin{equation}
\label{eq:Lipschitz_condition}
{\mathfrak{D}} ( D({{\bf{x}}_1}),D({{\bf{x}}_2}) ) \le {\mathcal{D}}({{\bf{x}}_1},{{\bf{x}}_2}),
\end{equation}
where ${\mathfrak{D}}: \Delta({\CMcal{A}}) \times \Delta({\CMcal{A}}) \to {{\mathbb{R}}^{+}}$ is a distance measure on distributions over ${\CMcal{A}}$. Moreover, we define $D$ satisfying individual fairness up to bias $\varepsilon$ if for all ${{\bf{x}}_1},{{\bf{x}}_2} \in {{\CMcal{R}}_{X}}$, we have
\begin{equation}
\label{eq:Lipschitz_condition_relaxation}
{\mathfrak{D}} ( D({{\bf{x}}_1}),D({{\bf{x}}_2}) ) \le {\mathcal{D}}({{\bf{x}}_1},{{\bf{x}}_2}) + \varepsilon. \qedhere
\end{equation}
\end{defn}

Individual fairness ensures a decision mapping maps similar people similarly. When two individuals' records ${{\bf{x}}_1}$ and ${{\bf{x}}_2}$ are similar, i.e., ${\mathcal{D}}({{\bf{x}}_1},{{\bf{x}}_2}) \cong 0$, the Lipschitz condition in equation (\ref{eq:Lipschitz_condition}) ensures that both records map to similar distributions over ${\CMcal{A}}$. Candidates for distance measure ${\mathfrak{D}}$ include (but not limited to) \emph{statistical distance} and \emph{relative ${l_{\infty}}$ metric}. The relative ${l_{\infty}}$ metric ( a.k.a. \emph{relative infinity norm}) of two distributions ${Z_1}$ and ${Z_2}$, defined as follow
\begin{equation}
\label{eq:D_relative_inf}
{{\mathfrak{D}}_{\infty}}( {Z_1},{Z_2} ) = \sup_{a\in {\CMcal{A}}} \log \bigg( \max \bigg\{ \frac{{Z_1}(a)}{{Z_2}(a)} , \frac{{Z_2}(a)}{{Z_1}(a)} \bigg\} \bigg),
\end{equation}
is considered a potential better choice in the aspect that it does not require the distance measure ${\mathcal{D}}$ to be re-scaled within $[0,1]$ \footnote{The normalization could bring non-trivial burden, especially when the maximal distance can be arbitrarily large.}. However, it has the shortcoming that it is sensitive to small probability values. The statistical distance, or the \emph{total variation norm}, of two distributions ${Z_1}$ and ${Z_2}$, defined as follow
\begin{equation}
\label{eq:D_stats_dis}
{{\mathfrak{D}}_{\emph{tv}}}( {Z_1},{Z_2} ) = {\frac{1}{|{\CMcal{A}}|}} {\sum_{a\in {\CMcal{A}}} \Big| {Z_1}(a) - {Z_2}(a) \Big|},
\end{equation}
is a more stable measure in this aspect.

\subsection{Measures for Group Fairness}
Popular measures for group fairness include (but not limited to) statistical parity (SP) (a.k.a. demographic parity) \cite{dwo11,feldman2015certifying,zemel2013learning,kamishima2012fairness}, conditional statistical parity (CSP) \cite{kamiran2013quantifying,corbett2017algorithmic}, $p$-\% rule (PR) \cite{feldman2015certifying,biddle2006adverse}, accuracy parity (a.k.a. equalized odds) \cite{har16}, and true positive parity (a.k.a. equal opportunity) \cite{har16}. However, the last two measures require knowledge of labeled outputs and is thus particular used to train fair ML algorithms in supervised learning. For algorithmic transparency, we use the former three measures for group fairness. 

Define $g(X)$ a projection function from input attributes $X$ onto a group in protected attributes, $v(X)$ a score/valuation function from $X$ onto a set scores, and ${T_{{\CMcal{Y}}}} \triangleq \{ {\bf{x}} \in {{\CMcal{R}}_{X}} \mid {g({\bf{x}})} \in {{\CMcal{Y}}} \}$ the set/tuple in which records belong to a protected group ${\CMcal{Y}}$. We summarize definitions of measures for group fairness in the following:
\begin{defn}
\label{def:stat_parity}
(\emph{Statistical Parity (SP)}) A decision mapping $D:{{\CMcal{R}}_{X}} \to \Delta({\CMcal{A}})$ satisfies statistical parity for two groups ${{\CMcal{Y}}_1}$ and ${{\CMcal{Y}}_2}$ up to bias $\varepsilon$ if for every decision outcome $a \in {\CMcal{A}}$, we have the following property
\begin{equation}
\label{eq:stat_parity}
{{\mathfrak{D}}_{\emph{tv}}} \Big( {{\operatorname{\mathbb{E}}}\left[ {D_a}(X)|{T_{{{\CMcal{Y}}_1}}} \right]},{{\operatorname{\mathbb{E}}}\left[ {D_a}(X)|{T_{{{\CMcal{Y}}_2}}} \right]} \Big) \le \varepsilon. \qedhere
\end{equation}
\end{defn}

\begin{defn}
\label{def:condi_stat_parity}
(\emph{Conditional Statistical Parity (CSP)}) Given a score/valuation function $v(X)$ based on input attributes $X$, define ${T_{{\CMcal{Y}},{\CMcal{V}}}} \triangleq \{ {\bf{x}} \in {{\CMcal{R}}_{X}} \mid {g({\bf{x}})} \in {{\CMcal{Y}}}, {v({\bf{x}})} \in {{\CMcal{V}}} \}$ the set/tuple in which records belong to a protected group ${\CMcal{Y}}$ having scores in a set ${{\CMcal{V}}}$. A decision mapping $D:{{\CMcal{R}}_{X}} \to \Delta({\CMcal{A}})$ satisfies conditional statistical parity given the same score conditions ${{\CMcal{V}}}$ for two groups ${{\CMcal{Y}}_1}$ and ${{\CMcal{Y}}_2}$ up to bias $\varepsilon$ if for every decision outcome $a \in {\CMcal{A}}$, we have the following property
\begin{equation}
\label{eq:condi_stat_parity}
{{\mathfrak{D}}_{\emph{tv}}} \Big( {{\operatorname{\mathbb{E}}}\left[ {D_a}(X)|{T_{{{\CMcal{Y}}_1},{\CMcal{V}}}} \right]},{{\operatorname{\mathbb{E}}}\left[ {D_a}(X)|{T_{{{\CMcal{Y}}_2},{\CMcal{V}}}} \right]} \Big) \le \varepsilon. \qedhere
\end{equation}
\end{defn}

\begin{defn}
\label{def:p_perct_rule}
(\emph{$p$-\% Rule (PR)}) A decision mapping $D:{{\CMcal{R}}_{X}} \to \Delta({\CMcal{A}})$ satisfies $p$-\% rule for two groups ${{\CMcal{Y}}_1}$ and ${{\CMcal{Y}}_2}$ if for every decision outcome $a \in {\CMcal{A}}$, we have the following property
\begin{equation}
\label{eq:p_perct_rule}
\bigg| \log\bigg(\frac{ {\operatorname{\mathbb{E}}}\left[ {D_a}(X)|{T_{{\CMcal{Y}}_1}} \right] }{ {\operatorname{\mathbb{E}}}\left[ {D_a}(X)|{T_{{\CMcal{Y}}_2}} \right] }\bigg) \bigg| \le -\log{p}. \qedhere
\end{equation}
\end{defn}

In particular, for binary decisions, we say a decision rule $d$ satisfies SP, CSP, or PR for two groups ${{\CMcal{Y}}_1}$ and ${{\CMcal{Y}}_2}$ up to bias $\varepsilon$ (SP and CSP only) if 
\begin{align}
\text{SP: } & \Big| {\operatorname{\mathbb{E}}}\left[ d(X)|{T_{{\CMcal{Y}}_1}} \right] - {\operatorname{\mathbb{E}}}\left[ d(X)|{T_{{\CMcal{Y}}_2}} \right] \Big| \le \varepsilon \label{eq:demo_parity} \\
\text{CSP: } & \Big| {\operatorname{\mathbb{E}}}\left[ d(X)|{T_{{{\CMcal{Y}}_1},{\CMcal{V}}}} \right] - {\operatorname{\mathbb{E}}}\left[ d(X)|{T_{{{\CMcal{Y}}_2},{\CMcal{V}}}} \right] \Big| \le \varepsilon \label{eq:condi_stat_parity_2} \\
\text{PR: } & p \le \frac{ {\operatorname{\mathbb{E}}}\left[ d(X)|{T_{{\CMcal{Y}}_1}} \right] }{ {\operatorname{\mathbb{E}}}\left[ d(X)|{T_{{\CMcal{Y}}_2}} \right] } \le \frac{1}{p} \text{.} \label{eq:p_perct_rule_2} 
\end{align}

Note that all fairness definitions are based on the distance between the decision  of two groups\footnote{
More precisely, from \eqref{eq:stat_parity}, \eqref{eq:condi_stat_parity}, and \eqref{eq:p_perct_rule}, the decision distribution of a group is the expected decision mapping among the group, over all decision outcomes $a \in {\CMcal{A}}$.}, specifically, \emph{total variation} \eqref{eq:D_stats_dis} and relative metric \eqref{eq:D_relative_inf}. 
Let ${\CMcal{F}}$ denote the set of all fairness definitions. Based on the use of distance metrics, ${\CMcal{F}}$ can be classified as follows:
\begin{itemize}
\item \emph{Total-variation-based fairness definitions (${\CMcal{F}}_{\textit{tv}}$)}: Definitions include $({{\mathfrak{D}}_{\textit{tv}}},{\mathcal{D}})$-individual fairness, statistical parity, and conditional statistical parity.
\item \emph{Relative-metric-based fairness definitions (${\CMcal{F}}_{\textit{rm}}$)}: Definitions include $({{\mathfrak{D}}_{\infty}},{\mathcal{D}})$- individual fairness and p\%-rule. \qedhere
\end{itemize}

\section{Privacy Leakage via Feature Importance/Interaction}\label{sec:privacy_leak_method_2}
Feature (value) importance, or feature (value) interaction, measures the \emph{importance} (or \emph{influence}) of input attributes (or attribute values) to the decision outcomes. The importance of an input attribute (value) is measured based on \emph{the corresponding change of output due to change of that certain input}. By changing an input, if the change of output is significant, it implies the input is important (has significant influence) to the output. On the other hand, if the output changes very little, the input contributes very little to the output.

Different works may propose different measures, but their philosophies are almost the same (as stated above). For example, the measures for change of an input can be (i) removing the presence of an input attribute, or (ii) permuting attribute values on an input attribute. The measures of outputs are many, e.g., (i) accuracy of the (predicted) outputs \cite{breiman2001random, fisher2018model}, (ii) probability of receiving a certain outcome \cite{dat16}, (iii) statistics measures, such as partial dependence \cite{friedman2001greedy, greenwell2018simple}, H-statistic \cite{friedman2008predictive}, or variable interaction networks \cite{hooker2004discovering}, or (iv) a self-defined quantity or a score/gain function. The measures for the change of outputs can be (i) difference (i.e., subtraction), (ii) ratio, or (iii) averaged difference/contribution, e.g., the Shapley value \cite{kononenko2010efficient}, of the measured outputs. In this regard, it is impractical for us to demonstrate the privacy leakage issue for all present methods. However, since the philosophies of all these methods are similar, it is reasonable for us to demonstrate the privacy hacking procedures via a representative one. The principles of hacking procedures can be transferred and applied to other methods. 

We investigate potential privacy leakage via the \emph{quantitative input influence} (QII) proposed in the most pioneering work \cite{dat16} in accountable ATR. For QII, the measure for change of an input is permuting attribute values (called \emph{intervention} in the paper) on an input attribute. The measure of output can be user-specified, called \emph{quantity of interest}, denoted by $Q$. The measure for change of output is difference between (subtraction of) two measured outputs. Formally, the QII of an input attribute $k$ for a quantity of interest $Q$ is defined as
\begin{equation}
{\mathit{I}}^{Q}(k) = Q(X) - Q({X_{-k}}{U_{k}}),
\end{equation}
in which ${X_{-k}}{U_{k}}$, meaning that attribute $k$ is (removed from input $X$ and) replaced by a permuted version ${U_{k}}$, represents intervention on attribute $k$. In particular, for $Q(X) = P\{c(X)=1|X \in {T_{\CMcal{W}}}\}$, the fraction of records belonging to a set ${T_{\CMcal{W}}}$ (e.g., women) with positive classification, the QII of an input attribute $k$ is
\begin{equation}
{\mathit{I}}(k) = P\{c(X)=1|X \in T_{\CMcal{W}}\} - P\{c({X_{-k}}{U_{k}})=1|X \in T_{\CMcal{W}}\},
\end{equation}
where $c(\cdot)$ is a classifier (decision-maker). The QII of a set of input attributes ${\CMcal{K}}$ is defined similarly, using ${\CMcal{K}}$ instead of $k$.

\begin{table}[t!]
\centering
\caption{Attribute Information of the Credit Approval Dataset}
\vspace*{-3mm}
\label{table:dataset_attribute}
\setlength\tabcolsep{2.5pt} 
\renewcommand{\arraystretch}{1}

\begin{tabular}{|>{\small}l|>{\small}l|>{\small}l|>{\small}l|}  
\hline
A1:& b, a. &A9:& t, f. \\ 
A2:& continuous. &A10:& t, f. \\ 
A3:& continuous. &A11:& continuous. \\ 
A4:& u, y, l, t. &A12:& t, f. \\ 
A5:& g, p, gg. &A13:& g, p, s. \\ 
A6:& c, d, cc, i, j, k, m, r, q, w, x, e, aa, ff. &A14:& continuous. \\ 
A7:& v, h, bb, j, n, z, dd, ff, o. &A15:& continuous. \\ 
A8:& continuous. &A16:& +,- (class attribute) \\ \hline
\end{tabular}
\end{table}

In the following, we conduct an experiment to demonstrate the hacking of decision rules via provided QII's on an ATR for a real dataset, and utilize the hacked decision rules to further infer private records as what we did in Section \ref{subsec:privacy_surrogate}. We use the Australian credit approval dataset from UCI machine learning repository \cite{Dua2017} in our experiment\footnote{Since we are demonstrating stealing private information from a real dataset, the chosen dataset needs to contain critical information, and its size needs to be adequate: on the one hand, it should not be too large for ease of demonstration; on the other hand, it should not be too small for the accuracy of the trained classifier.}. The dataset has 690 instances, with 15 input attributes and 1 output attribute. All attribute information can be found in Table \ref{table:dataset_attribute}. In order to protect confidentiality of the data, all attribute names and values have been changed to meaningless symbols by the dataset provider. Based on the dataset, with adequate data cleaning and pre-processing, we train a classifier based on a fully-connected neural network with one input layer (36 inputs, after one-hot encoding for categorical attribute values), two hidden layers (147 and 85 neurons, respectively), and one output layer (binary outputs), with dropout rate 0.5. The averaged testing accuracy of the trained classifier is 89.5\%. 

The trained classifier is served as the knowledge of a trust-worthy 3rd-party regulation agency which feeds both inputs and outputs of the dataset to a ML model in order to learn the unknown decision-making rules of this Australian credit card company. Since QII is a data-mining based approach \cite{dat16}, the regulation agency provides information regarding input influences (QII) in an ATR upon users' demand. Since the access control is still an open question, we assume a user is able to request such information in a reasonable manner.

Based on the above experimental settings, we first construct a scenario to demonstrate the hacking. 

\vspace{3mm}
\noindent\textbf{Scenario}:
\begin{itemize}[leftmargin=*]
\item Let $\CMcal{U} =$ \{A4, A5, A6, A7\} be public attributes and all other attributes are private and unknown to adversaries (See Remark \ref{rmk:A4567}).
\item Alice has public record ${{\bf{x}}_{\CMcal{U}}} =$ \{y, p, k, v\}. She gets a positive decision (+) and receives a credit card.
\item Tom also has the same public record ${{\bf{x}}_{\CMcal{U}}} =$ \{y, p, k, v\}. He gets a negative decision (-).
\item An adversary is a friend of both, knowing their public records, knowing that Alice owns such a credit card but Tom doesn't. The adversary also has the knowledge of joint distribution of A4$\sim$A7, A9, and A11, e.g., demographic statistics of age, marriage status, race, and annual income.
\end{itemize}
A snapshot of the QID group ${T_{{{\bf{x}}_{\CMcal{U}}}=\text{\{y, p, k, v\}}}}$ is shown in Table \ref{table:ypkv}\footnote{We remove attributes A1$\sim$A3 in the interest of space}, in which public attributes are marked as grey, class attribute (decision outcome) is marked as light blue, and for those attributes that an adversary has associated side-information (joint distribution) are marked as bold.

\begin{table}[tbp]
\centering
\caption{A Snapshot of the QID Group ${T_{{{\bf{x}}_{\CMcal{U}}}=\text{\{y, p, k, v\}}}}$}
\vspace*{-3mm}
\centering
\begin{adjustwidth}{+1in}{-0in}
\begin{minipage}{0.5\textwidth} 
\label{table:ypkv}
\centering
\setlength\tabcolsep{2.5pt} 
\renewcommand{\arraystretch}{0.8}
\begin{tabular}{
>{\small\columncolor[HTML]{EFEFEF}}c 
>{\small\columncolor[HTML]{EFEFEF}}c 
>{\small\columncolor[HTML]{EFEFEF}}c 
>{\small\columncolor[HTML]{EFEFEF}}c 
>{\small}c>{\small}c>{\small}c>{\small}c>{\small}c>{\small}c>{\small}c>{\small}c
>{\small\columncolor[HTML]{ECF4FF}}c}
{\color[HTML]{000000} \textbf{A4}} & {\color[HTML]{000000} \textbf{A5}} & {\color[HTML]{000000} \textbf{A6}} & {\color[HTML]{000000} \textbf{A7}} & \textbf{A8} & {\color[HTML]{000000} \textbf{A9}} & \textbf{A10} & {\color[HTML]{000000} \textbf{A11}} & \textbf{A12} & \textbf{A13} & \textbf{A14} & \textbf{A15} & \textbf{A16} \\
{\color[HTML]{000000} \textbf{y}}  & {\color[HTML]{000000} \textbf{p}}  & {\color[HTML]{000000} \textbf{k}}  & {\color[HTML]{000000} \textbf{v}}  & 0.125       & {\color[HTML]{000000} \textbf{f}}  & f            & {\color[HTML]{000000} \textbf{0}}   & f            & g            & 160          & 0            & -            \\
{\color[HTML]{000000} \textbf{y}}  & {\color[HTML]{000000} \textbf{p}}  & {\color[HTML]{000000} \textbf{k}}  & {\color[HTML]{000000} \textbf{v}}  & 0.25        & {\color[HTML]{000000} \textbf{f}}  & f            & {\color[HTML]{000000} \textbf{0}}   & f            & g            & 224          & 0            & -            \\
{\color[HTML]{000000} \textbf{y}}  & {\color[HTML]{000000} \textbf{p}}  & {\color[HTML]{000000} \textbf{k}}  & {\color[HTML]{000000} \textbf{v}}  & 0.29        & {\color[HTML]{000000} \textbf{f}}  & f            & {\color[HTML]{000000} \textbf{0}}   & f            & s            & 200          & 0            & -            \\
{\color[HTML]{000000} \textbf{y}}  & {\color[HTML]{000000} \textbf{p}}  & {\color[HTML]{000000} \textbf{k}}  & {\color[HTML]{000000} \textbf{v}}  & 1.25        & {\color[HTML]{000000} \textbf{f}}  & f            & {\color[HTML]{000000} \textbf{0}}   & t            & g            & 280          & 0            & -            \\
{\color[HTML]{000000} \textbf{y}}  & {\color[HTML]{000000} \textbf{p}}  & {\color[HTML]{000000} \textbf{k}}  & {\color[HTML]{000000} \textbf{v}}  & 0.125       & {\color[HTML]{000000} \textbf{f}}  & f            & {\color[HTML]{000000} \textbf{0}}   & f            & g            & 140          & 4            & -            \\
{\color[HTML]{000000} \textbf{y}}  & {\color[HTML]{000000} \textbf{p}}  & {\color[HTML]{000000} \textbf{k}}  & {\color[HTML]{000000} \textbf{v}}  & 0.125       & {\color[HTML]{000000} \textbf{f}}  & f            & {\color[HTML]{000000} \textbf{0}}   & f            & g            & 200          & 70           & -            \\
{\color[HTML]{000000} \textbf{y}}  & {\color[HTML]{000000} \textbf{p}}  & {\color[HTML]{000000} \textbf{k}}  & {\color[HTML]{000000} \textbf{v}}  & 0.085       & {\color[HTML]{000000} \textbf{f}}  & f            & {\color[HTML]{000000} \textbf{0}}   & f            & g            & 216          & 2100         & -            \\
{\color[HTML]{000000} \textbf{y}}  & {\color[HTML]{000000} \textbf{p}}  & {\color[HTML]{000000} \textbf{k}}  & {\color[HTML]{000000} \textbf{v}}  & 0.415       & {\color[HTML]{000000} \textbf{f}}  & t            & {\color[HTML]{000000} \textbf{1}}   & t            & g            & 280          & 80           & -            \\
{\color[HTML]{000000} \textbf{y}}  & {\color[HTML]{000000} \textbf{p}}  & {\color[HTML]{000000} \textbf{k}}  & {\color[HTML]{000000} \textbf{v}}  & 2.5         & {\color[HTML]{000000} \textbf{t}}  & t            & {\color[HTML]{000000} \textbf{1}}   & f            & g            & 180          & 20           & -            \\
{\color[HTML]{000000} \textbf{y}}  & {\color[HTML]{000000} \textbf{p}}  & {\color[HTML]{000000} \textbf{k}}  & {\color[HTML]{000000} \textbf{v}}  & 0.25        & {\color[HTML]{000000} \textbf{t}}  & t            & {\color[HTML]{000000} \textbf{10}}  & t            & g            & 320          & 0            & +            \\
{\color[HTML]{000000} \textbf{y}}  & {\color[HTML]{000000} \textbf{p}}  & {\color[HTML]{000000} \textbf{k}}  & {\color[HTML]{000000} \textbf{v}}  & 0.25        & {\color[HTML]{000000} \textbf{t}}  & t            & {\color[HTML]{000000} \textbf{11}}  & f            & g            & 380          & 2732         & +           
\end{tabular}
\end{minipage}
\end{adjustwidth}
\end{table}

We next demonstrate privacy hacking procedures in the following. Let ${{\CMcal{W}}_0}=$\{A4=y, A5=p, A6=k, A7=v, A11$\in$[0,1]\}, and ${{\CMcal{W}}_1}=$\{A4=y, A5=p, A6=k, A7=v, A11$\in$[10,11]\}.

\vspace{3mm}
\noindent\textbf{Privacy Hacking}:
\begin{enumerate}[leftmargin=*,parsep=0pt]

\item  Since the input to QII can be a set of attributes, i.e., the joint influence of a set of input attributes. Let $\CMcal{S}$ be the collection of all private attributes as denoted in Table \ref{table:notation}, which is \{A1$\sim$A3, A8$\sim$A15\} in our scenario. The adversary then sends the following QII query to the regulation agency:
    \begin{itemize}
      \item{Input Attribute: $\CMcal{S}$}
      \item{Quantity of Interest: $Q(X) = P\{c(X)=1|X \in {T_{{\CMcal{W}}_1}}\}$}
    \end{itemize}

\item The adversary gets a response ${\mathit{I}}({\CMcal{S}}) = 0.66475333$, which indicates the degree of influence of all private input attributes $\CMcal{S}$ to the group ${{\CMcal{W}}_1}$.

\item The adversary sends the following QII query to the regulation agency:
    \begin{itemize}
      \item{Input Attribute: $\CMcal{S}$}
      \item{Quantity of Interest: $Q(X) = P\{c(X)=1|X \in {T_{{\CMcal{W}}_0}}\}$}
    \end{itemize}

\item The adversary gets a response ${\mathit{I}}({\CMcal{S}}) = -0.33524666$, which indicates the degree of influence of all private input attributes $\CMcal{S}$ to the group ${{\CMcal{W}}_0}$. Note that negative sign stands for negative impact as mentioned in Section \ref{subsec:transparency}.

\item From the above two query responses, the adversary has
\newcounter{enumi_saved}
\setcounter{enumi_saved}{\value{enumi}}
\end{enumerate}
\vspace{-2mm}
\begin{align*}
 0.66475333 &= P\{c(X)=1|X \in {T_{{\CMcal{W}}_1}}\} - P\{c({X_{-{\CMcal{S}}}}{U_{{\CMcal{S}}}})=1|X \in {T_{{\CMcal{W}}_1}}\} \\
-0.33524666 &= P\{c(X)=1|X \in {T_{{\CMcal{W}}_0}}\} - P\{c({X_{-{\CMcal{S}}}}{U_{{\CMcal{S}}}})=1|X \in {T_{{\CMcal{W}}_0}}\}
\end{align*}
\vspace{-5mm}
\begin{enumerate}[leftmargin=*,nosep]
\setcounter{enumi}{\value{enumi_saved}}

\item Since ${{\CMcal{W}}_1}$ and ${{\CMcal{W}}_0}$ have the same public record ${{\bf{x}}_{\CMcal{U}}} =$ \{y, p, k, v\}, for the same classifier, we must have
\setcounter{enumi_saved}{\value{enumi}}
\end{enumerate}
\vspace{-1mm}
\begin{align*}
P\{c({X_{-{\CMcal{S}}}}{U_{{\CMcal{S}}}})=1|X \in {T_{{\CMcal{W}}_1}}\} 
=P\{c({X_{-{\CMcal{S}}}}{U_{{\CMcal{S}}}})=1|X \in {T_{{\CMcal{W}}_0}}\}.
\end{align*}
\vspace{-5mm}
\begin{enumerate}[leftmargin=*,nosep]
\setcounter{enumi}{\value{enumi_saved}}

\item Utilize the above equality, the adversary obtains
\begin{align*}
P\{c(X)=1|X \in {T_{{\CMcal{W}}_1}}\} - P\{c(X)=1|X \in {T_{{\CMcal{W}}_0}}\} = 1.
\end{align*}
Since probabilities are always within $[0,1]$, the adversary thus obtains decision rules
\begin{align*}
&P\{c(X)=1|X \in {T_{{\CMcal{W}}_1}}\} = 1, \\
&P\{c(X)=1|X \in {T_{{\CMcal{W}}_0}}\} = 0.
\end{align*}
\end{enumerate}

It is worth mentioning that the attack may not be unique. As shown in the following, there could exist many ways to obtain decision rules, and thus it seems hopeless to cease the attack simply by access control.

\vspace{3mm}
\noindent\textbf{Privacy Hacking (Method 2)}:
\begin{enumerate}[leftmargin=*,parsep=0pt]

\item The adversary sends the following QII query to the regulation agency:
    \begin{itemize}
      \item{Input Attribute: A9}
      \item{Quantity of Interest: $Q(X) = P\{c(X)=1|X \in {T_{{\CMcal{W}}_1}}\}$}
    \end{itemize}
    
\item The adversary gets a response ${\mathit{I}}(\text{A9}) = 0.45142778$, which indicates the degree of influence of input attribute A9 to the group ${{\CMcal{W}}_1}$.

\item The adversary analyzes the response ${\mathit{I}}(\text{A9})$.Define $P_{t} = P\{c({X_{-\text{A9}}}{U_{\text{A9}}})=1|X \in {T_{{\CMcal{W}}_1}}, {U_{\text{A9}}}=\text{t}\}$ 
and $P_{f} = P\{c({X_{-\text{A9}}}{U_{\text{A9}}})=1|X \in {T_{{\CMcal{W}}_1}}, {U_{\text{A9}}}=\text{f}\}$. He gets
\newcounter{enumi_saved_2}
\setcounter{enumi_saved_2}{\value{enumi}}
\end{enumerate}
\vspace{-1.5mm}
\begin{alignat*}{2}
0.45142778 
=&P\{c(X)=1|X \in {T_{{\CMcal{W}}_1}}\} &&- P\{c({X_{-\text{A9}}}{U_{\text{A9}}})=1|X \in {T_{{\CMcal{W}}_1}}\} \\ 
=&P\{c(X)=1|X \in {T_{{\CMcal{W}}_1}}\} &&-P\{ {U_{\text{A9}}}=\text{t} \}{P_{t}} - P\{ {U_{\text{A9}}}=\text{f} \}{P_{f}} 
\end{alignat*}
\vspace{-5mm}
\begin{enumerate}[leftmargin=*,topsep=0pt]
\setcounter{enumi}{\value{enumi_saved_2}}

\item The adversary realizes the fact that, for the same classifier, we must have
\begin{align*}
  P\{c(X)=1|X \in {T_{{\CMcal{W}}_1}}\} 
=&P\{c(X)=1|X \in {T_{{\CMcal{W}}_1}}, {A9}=\text{t}\} \\
=&P\{c({X_{-A9}}{U_{A9}})=1|X \in {T_{{\CMcal{W}}_1}}, {U_{A9}}=\text{t}\}=P_t
\end{align*}

\item Since the adversary has joint distribution knowledge as mentioned in the scenario, he knows the marginal distribution ${P\{ {U_{A9}}=\text{f} \}} = 1-{P\{ {U_{A9}}=\text{t} \}} = 0.45142857$, he then gets
\begin{align*}
P\{c(X)=1|X \in {T_{{\CMcal{W}}_1}}\} - P_f = \frac{0.45142778}{0.45142857} \approx 1
\end{align*}

\item Since probabilities are always within $[0,1]$, the adversary knows $P_f \approx 0$, and 
\begin{align*}
P\{c(X)=1|X \in {T_{{\CMcal{W}}_1}}\} \approx 1.
\end{align*}
The adversary obtains very accurate information regarding decision rule for ${{\CMcal{W}}_1}$.
\end{enumerate}

Based on the hacked decision rules above, the adversary has 100\% confidence that Alice's record belongs to ${T_{{\CMcal{W}}_1}}$ and Tom's record belongs to ${T_{{\CMcal{W}}_0}}$. Based on Table \ref{table:ypkv}, he then knows that Alice's A11 attribute value is either 10 or 11, and Tom's is either 0 or 1. If the adversary has richer side-information, e.g., joint distribution including A8 and A14, then the adversary has 100\% confidence that Alice's A8 attribute value is 0.25, her A14 attribute value is in the range between 300 and 400, and Tom's A14 attribute value is in the range between 100 and 300.

It is worth mentioning that, based on our investigation, we do not find a general attack method that can be applied to all datasets and decision rules. However, this does not mean the attacks demonstrated above are cherry-picked. As we have shown, there could exist many feasible attack approaches. Adversaries can simply try multiple different attempts and/or collude their test results so that eventually acquire a successful attack result. Moreover, similar to the privacy incidents of AOL search data leak \cite{barbaro2006face} and de-anonymization of the Netflix Price dataset \cite{narayanan2008robust}, although there is no guarantee that the attacks can always succeed in all the cases, \emph{as long as the attack \textbf{can} succeed, there exists a privacy breach which can result in a catastrophic disaster}.

In fact, the authors of the pioneering work, i.e., \cite{dat16}, had already noticed the potential privacy issue in algorithmic transparency and added noise to make the measures differentially private.
Unfortunately, adding differentially private noise \cite{dwork2006calibrating} solely cannot mitigate the demonstrated privacy leakage issue. The fundamental reason is that differential privacy only guarantees a small amount of information leakage when an individual participates the survey or opts into a database. Differential privacy itself does not guarantee information leakage due to strong statistical inference between attributes; this has been noted in many previous works such as \cite{barthe2011information, alvim2011differential, du2012privacy}, and section 2.3.2 in \cite{dwork2014algorithmic}. The most classic example is the study of \quotes{smoking causes cancer}, in which no matter whether a person opts into the survey or not, once we know that he is a smoker, we know he has a certain high chance of getting lung cancer. What can be guaranteed in the proposed differentially private perturbation for an ATR is that an adversary can only gain very little information by comparing two ATRs of which the training data to train the classifiers are differ in only one data subject's record. When the size of dataset is very large, the required variance of DP noise is very small. This is why they claimed only very little noise needs to be added.

\begin{rmk}
\label{rmk:A4567}
Although all attribute names in the dataset are removed, we are still able to reasonably conjecture public and private attributes based on their influences to the decision outcome. Attributes with high influences are more likely to be private attributes such as income or credit score, and attributes with low influences are likely to be public ones. Observing that attribute A9, A11, and A15 are the most influential ones and others are less significant (from experiments). For ease of demonstration, we choose 4 adjacent categorical attributes from insignificant ones, A4 to A7, to serve as public attributes.
\end{rmk}

\section{Minimum Uncertainty}\label{sec:min_uncertainty}
\begin{defn}
\label{def:min_uncertainty}
(\textit{Minimum Uncertainty}) Given an inference channel $\langle {{X}_{\CMcal{U}}},A \to {{X}_{\CMcal{S}}} \rangle$, the uncertainty of inferring a certain sensitive attribute value ${x_{\CMcal{S}}}$ from a certain inference source $\{ {{\bf{x}}_{\CMcal{U}}}, a \}$ is defined as $\mathit{ucrt}({{\bf{x}}_{\CMcal{U}}},a \to {{x}_{\CMcal{S}}}) = -\log \big( \mathit{conf}({{\bf{x}}_{\CMcal{U}}},a \to {{x}_{\CMcal{S}}}) \big)$. The minimum uncertainty of inferring any sensitive value from any inference channel is 
\begin{equation*}
\begin{split}
\mathit{Ucrt}({X_{\CMcal{U}}},A \to {{X}_{\CMcal{S}}}) 
= & {\min_{{{\bf{x}}_{\CMcal{U}}}, a, {{x}_{\CMcal{S}}}}} \{ -\log \big( \mathit{conf}({{\bf{x}}_{\CMcal{U}}},a \to {{x}_{\CMcal{S}}}) \big) \} \\
= & -\log \big( {\max_{{{\bf{x}}_{\CMcal{U}}}, a, {{x}_{\CMcal{S}}}}} \{\mathit{conf}({{\bf{x}}_{\CMcal{U}}},a \to {{x}_{\CMcal{S}}})\} \big) \\
= & -\log \big( \mathit{Conf}({X_{\CMcal{U}}},A \to {{X}_{\CMcal{S}}}) \big). \qedhere
\end{split} 
\end{equation*}
\end{defn}
Similarly, the corresponding privacy requirement for minimal uncertainty is the following.
\begin{defn}
\label{def:gamma_min_uncertainty}
($\gamma$-\textit{Minimum Uncertainty}) In an algorithmic transparency report, $\tilde{D}$ satisfies $\gamma$-\textit{Minimum Uncertainty} if $\mathit{Ucrt}({X_{\CMcal{U}}},A \to {X_{\CMcal{S}}}) \ge \gamma$.
\end{defn}
The above privacy requirement is basically saying that an adversary's uncertainty on inferring any sensitive value from any inference channel cannot be too low and should be lower-bounded by a threshold $\gamma$; the larger the $\gamma$, the higher the minimum uncertainty, and thus the stronger the privacy. From definition (\ref{def:min_uncertainty}), it is clear that $\gamma$-Minimum Uncertainty implies ${e^{-\gamma}}$-Maximum Confidence, and $\beta$-Maximum Confidence implies $\unaryminus \log \beta$-Minimum Uncertainty.
\begin{lem}
\label{lem:gamma_min_uncertainty}
The privacy requirement $\gamma$-Minimum Uncertainty imposes the following constraints to the announced decision mapping $\tilde{D}$, $\forall {\bf{x}} \in {{\CMcal{R}}_{X}}$, $\forall a \in {{\CMcal{A}}}$,
\begin{equation}
\label{eq:gamma_min_uncertainty}
\log \big( { \sum_{{\bf{x}'} } {{{\tilde{D}}_{a}}({\bf{x}'})}{{P_X}({\bf{x}'})} } \big) - \log \big( { {{{\tilde{D}}_{a}}({\bf{x}})}{{P_X}(\bf{x})} } \big)  \ge \gamma.
\end{equation}
\end{lem}



\section{Proof of Lemma \ref{lem:beta_max_conf}}\label{sec:appendix_00}
\begin{proof}
Recall that $\mathit{conf}({{\bf{x}}_{\CMcal{U}}},a \to {{x}_{\CMcal{S}}})$, the confidence of inferring a sensitive attribute value ${{x}_{\CMcal{S}}}$, is a posterior epistemic probability which can be expressed as
\begin{equation}
\label{eq:conf_xs}
\begin{split}
\mathit{conf}({{\bf{x}}_{\CMcal{U}}},a \to {{x}_{\CMcal{S}}})
= {\tilde{P}_{{X_{\CMcal{S}}}|{X_{\CMcal{U}}},A}}({x_{\CMcal{S}}}|{\bf{x}_{\CMcal{U}}}, a) 
= \frac{{{\tilde{P}_{A|{X_{\CMcal{U}}},{X_{\CMcal{S}}}}}(a|{{\bf{x}}_{\CMcal{U}}}, {x_{\CMcal{S}}})}{{P_{{X_{\CMcal{U}}},{X_{\CMcal{S}}}}}({{\bf{x}}_{\CMcal{U}}}, {x_{\CMcal{S}}})}}{\sum\limits_{{x'_{\CMcal{S}}} \in {{\CMcal{R}}_{X_{\CMcal{S}}}}} {{\tilde{P}_{A|{X_{\CMcal{U}}},{X_{\CMcal{S}}}}}(a|{{\bf{x}}_{\CMcal{U}}}, {x'_{\CMcal{S}}}) {{P_{{X_{\CMcal{U}}},{X_{\CMcal{S}}}}}({{\bf{x}}_{\CMcal{U}}}, {x'_{\CMcal{S}}})}} }.
\end{split}
\end{equation}
Let ${\bf{x}} = ( {{\bf{x}}_{\CMcal{U}}}, {x_{\CMcal{S}}} )$ and define ${T_{{\bf{x}}_{\CMcal{U}}}} \triangleq \{ {\bf{x}'} \in {{\CMcal{R}}_{X}} \mid {{\bf{x}'}_{\CMcal{U}}} = {\bf{x}_{\CMcal{U}}} \}$ to denote the tuple in which records having the same QID ${\bf{x}_{\CMcal{U}}}$. We have a more comprehensive expression
\begin{equation}
\label{eq:conf_xs_2}
\begin{split}
\mathit{conf}({{\bf{x}}_{\CMcal{U}}},a \to {{x}_{\CMcal{S}}}) = \frac{ {{\tilde{P}_{A|X}}(a|{\bf{x}})}{{P_X}({\bf{x}})} }{ \sum\limits_{{\bf{x}'} \in {T_{{\bf{x}}_{\CMcal{U}}}}} {{\tilde{P}_{A|X}}(a|{\bf{x}'})}{{P_X}({\bf{x}'})} } 
= { \frac{ {{{\tilde{D}}_{a}}({\bf{x}})}{{P_X}(\bf{x})} }{ \sum\limits_{{\bf{x}'} \in {T_{{\bf{x}}_{\CMcal{U}}}}} {{{\tilde{D}}_{a}}({\bf{x}'})}{{P_X}({\bf{x}'})} } }.
\end{split}
\end{equation}
Therefore, based on Definitions \ref{def:max_conf} and \ref{def:beta_max_conf}, the privacy requirement $\beta$-Maximum Confidence imposes the following constraints for all ${\bf{x}} = ( {{\bf{x}}_{\CMcal{U}}}, {x_{\CMcal{S}}} ) \in {{\CMcal{R}}_{X}}$, $\forall a \in {{\CMcal{A}}}$.
\begin{equation}
\label{eq:privacy_requirements}
{ \frac{ {{{\tilde{D}}_{a}}({\bf{x}})}{{P_X}(\bf{x})} }{ \sum\limits_{{\bf{x}'} \in {T_{{\bf{x}}_{\CMcal{U}}}}} {{{\tilde{D}}_{a}}({\bf{x}'})}{{P_X}({\bf{x}'})} } } \le \beta.   \qedhere
\end{equation} 
\end{proof}


\section{Proof of Lemma \ref{lem:optimal_utility_condition}}\label{sec:appendix_01}
\begin{proof}
We first prove that if $\beta \ge C^{*}$, $\tilde{D} = D$ is a feasible solution. We then prove its converse: if $\tilde{D} = D$ is a feasible solution, we must have $\beta \ge C^{*}$. 

We first prove that if $\beta \ge C^{*}$, the 1-fidelity solution $\tilde{D} = D$ is a feasible solution, i.e., it satisfies all constraints. Obviously, the solution $\tilde{D} = D$ satisfies probability distribution conditions and fidelity constraints. Based on definition \ref{def:max_conf_tuple}, $\tilde{D} = D$ yields 
\begin{equation*}
{ \frac{ {{P}(\bf{x})}{{{\tilde{D}}_a}({\bf{x}})} }{ \sum\limits_{{\bf{x}'} \in {T_{{\bf{x}}_{\CMcal{U}}}}} {{P}({\bf{x}'})}{{{\tilde{D}}_a}({\bf{x}'})} } } = C^{*} \le \beta \,, \; \forall {{\bf{x}}} \in {{T_{{\bf{x}}_{\CMcal{U}}}}}, \forall a \in \CMcal{A}.
\end{equation*}
Therefore, it also satisfies privacy constraints, and hence when $\beta \ge C^{*}$, the 1-fidelity solution is a feasible solution.

Next, we prove the converse by proving its contrapositive, i.e., if $\beta < C^{*}$, $\tilde{D} = D$ is not a feasible solution. Apparently when $\tilde{D} = D$, the highest confidence that an adversary can have exceeds $\beta$, and hence it violates privacy requirements and cannot be a feasible solution. We therefore prove the converse.
\end{proof}

\section{Proof of Lemma \ref{lem:feasible_condition_no_fidelity}}\label{sec:appendix_02}
\begin{proof}
We first prove that if an \eqref{eq:OPT_sub} has feasible solutions, $\beta \ge {\beta_{\text{min}}}$. We then prove its converse: if $\beta \ge {\beta_{\text{min}}}$, an \eqref{eq:OPT_sub} must have feasible solutions. 

We first prove the conditional statement by proving its contrapositive, i.e., if $\beta < {\beta_{\text{min}}}$, there exists no feasible solution for an \eqref{eq:OPT_sub}. Since ${\tilde{D}}$ is non-negative, we can rewrite the privacy constraints as follow.
\begin{equation}
\label{eq:privacy_constraints_eqv}
{ {{P}(\bf{x})}{{{\tilde{D}}_{a}}({\bf{x}})} } - \beta { \sum_{\substack{ {{\bf{x}'} \in {T_{{\bf{x}}_{\CMcal{U}}}}} }} {{P}({\bf{x}'})}{{{\tilde{D}}_{a}}({\bf{x}'})} } \le 0 \,, 
\end{equation}
which has to be satisfied $\forall {{\bf{x}}} \in {T_{{\bf{x}}_{\CMcal{U}}}}$ and $\forall a \in {\CMcal{A}}$. Sum \eqref{eq:privacy_constraints_eqv} over all $a \in \CMcal{A}$, by \eqref{eq:sub_eq_constraints}, we have
\begin{equation}
\label{eq:privacy_constraints_eqv_sum}
{{P}(\bf{x})} - \beta { \sum_{\substack{ {{\bf{x}'} \in {T_{{\bf{x}}_{\CMcal{U}}}}} }} {{P}({\bf{x}'})} } \le 0 \,, \forall {{\bf{x}}} \in {T_{{\bf{x}}_{\CMcal{U}}}},
\end{equation}
which is equivalent to $\beta \ge {\max_{{{\bf{x}} \in {T_{{\bf{x}}_{\CMcal{U}}}}}}} {{P}({\bf{x}}|{T_{{\bf{x}}_{\CMcal{U}}}})}$. Therefore, if there exists \textit{any} ${{\bf{x}}} \in {T_{{\bf{x}}_{\CMcal{U}}}}$ such that $\beta < {{P}({\bf{x}}|{T_{{\bf{x}}_{\CMcal{U}}}})}$, then \eqref{eq:privacy_constraints_eqv} cannot be satisfied for \textit{all} ${{\bf{x}}} \in {T_{{\bf{x}}_{\CMcal{U}}}}$, and hence no feasible solution exists.

We then prove the converse. If $\beta \ge {\max_{{{\bf{x}} \in {T_{{\bf{x}}_{\CMcal{U}}}}}}} {{P}({\bf{x}}|{T_{{\bf{x}}_{\CMcal{U}}}})}$, there always exists a feasible solution ${{{\tilde{D}}_{a}}({\bf{x}'})} = 1/{|{\CMcal{A}}|}$, $\forall {{\bf{x}}} \in {T_{{\bf{x}}_{\CMcal{U}}}}$, $\forall a \in \CMcal{A}$. To see this, we only need to verify if it satisfies all constraints. It is very obvious that the solution satisfies probability distribution conditions. Since fidelity constraints are trivialized, we then only need to verify if the solution satisfies privacy constraints. Since ${{{\tilde{D}}_{a}}({\bf{x}'})}$ is a constant for all $a$ and ${\bf{x}}$, the left hand side of \eqref{eq:privacy_constraints_eqv} becomes ${{P}({\bf{x}}|{T_{{\bf{x}}_{\CMcal{U}}}})}$, and thus the privacy constraints are also satisfied. Hence ${{{\tilde{D}}_{a}}({\bf{x}'})} = 1/{|{\CMcal{A}}|}$ is a feasible solution and we proved the converse.
\end{proof}

\section{Proof of Lemma \ref{lem:optimal_vs_feasible_condition_no_fidelity}}\label{sec:appendix_03}
\begin{proof}
We prove this by contradiction. Assume that ${C^{*}} < {\beta_{\text{min}}}$. By their definitions in Lemma \ref{lem:optimal_utility_condition} and \ref{lem:feasible_condition_no_fidelity}, it follows that
\begin{equation}
\label{eq:controdition_eq1}
{\max_{\substack{  {{\bf{x}} \in {T_{{\bf{x}}_{\CMcal{U}}}}}, \\ {a \in \CMcal{A}} }}} {{ \frac{ {{P}(\bf{x})}{{{D}_a}({\bf{x}})} }{ \sum\limits_{{\bf{x}'} \in {T_{{\bf{x}}_{\CMcal{U}}}}} {{P}({\bf{x}'})}{{{D}_a}({\bf{x}'})} } }} < {\underset{{{\bf{x}} \in {T_{{\bf{x}}_{\CMcal{U}}}}}}{\max}} {{ \frac{ {{P}(\bf{x})} }{ \sum\limits_{{\bf{x}'} \in {T_{{\bf{x}}_{\CMcal{U}}}}} {{P}({\bf{x}'})} } }}.
\end{equation}
Let ${{\bf{x}}^{\dagger}} = {\argmax_{{ {{\bf{x}} \in {T_{{\bf{x}}_{\CMcal{U}}}}} }}} {{P}({\bf{x}}|{T_{{\bf{x}}_{\CMcal{U}}}})}$. The right hand side of \eqref{eq:controdition_eq1} is equivalent to ${{ { {{P}({{\bf{x}}^{\dagger}})} }/{ \sum_{{\bf{x}'} \in {T_{{\bf{x}}_{\CMcal{U}}}}} {{P}({\bf{x}'})} } }}$. If inequality \eqref{eq:controdition_eq1} holds, the following inequalities must hold
\begin{equation}
\label{eq:controdition_eq2}
{{ \frac{ {{P}({{\bf{x}}^{\dagger}})}{{{D}_a}({{\bf{x}}^{\dagger}})} }{ \sum\limits_{{\bf{x}'} \in {T_{{\bf{x}}_{\CMcal{U}}}}} {{P}({\bf{x}'})}{{{D}_a}({\bf{x}'})} } }} < {{ \frac{ {{P}({{\bf{x}}^{\dagger}})} }{ \sum\limits_{{\bf{x}'} \in {T_{{\bf{x}}_{\CMcal{U}}}}} {{P}({\bf{x}'})} } }}\,, \;  \forall a \in {\CMcal{A}},
\end{equation}
since the maximum of the left hand side of \eqref{eq:controdition_eq2} over all $a \in {\CMcal{A}}$ is not greater than the left hand side of \eqref{eq:controdition_eq1}. Therefore, if there exists any $a \in {\CMcal{A}}$ for which the corresponding inequality in \eqref{eq:controdition_eq2} does not hold, it implies our assumption ${C^{*}} < {\beta_{\text{min}}}$ is not true, and, if so, we are done with the proof.

If there exists no such an $a$ and \eqref{eq:controdition_eq2} holds, by eliminating ${{P}({{\bf{x}}^{\dagger}})}$ from both sides of \eqref{eq:controdition_eq2} and cross-multiplying (as all terms are non-negative), \eqref{eq:controdition_eq2} is equivalent to the following
\begin{equation}
\label{eq:controdition_eq3}
{{{D}_a}({{\bf{x}}^{\dagger}})} { \sum\limits_{{\bf{x}'} \in {T_{{\bf{x}}_{\CMcal{U}}}}} {{P}({\bf{x}'})} } < { \sum\limits_{{\bf{x}'} \in {T_{{\bf{x}}_{\CMcal{U}}}}} {{P}({\bf{x}'})}{{{D}_a}({\bf{x}'})} }\,, \;  \forall a \in {\CMcal{A}}.
\end{equation}
Sum \eqref{eq:controdition_eq3} over $a \in {\CMcal{A}}$ for both sides, based on \eqref{eq:sub_eq_constraints}, we obtain ${ \sum_{{\bf{x}'} \in {T_{{\bf{x}}_{\CMcal{U}}}}} {{P}({\bf{x}'})} } < { \sum_{{\bf{x}'} \in {T_{{\bf{x}}_{\CMcal{U}}}}} {{P}({\bf{x}'})} }$, which is obviously not true. Therefore, it implies the inequality \eqref{eq:controdition_eq3} (and \eqref{eq:controdition_eq2}, equivalently) cannot be true for all $a \in {\CMcal{A}}$, i.e., there must exist some $a$ for which the left hand side is not smaller than the right hand side of \eqref{eq:controdition_eq2}, so that both sides are equal when summed over all $a$. Therefore, the initial assumption is incorrect and the lemma is proved.
\end{proof}

\section{Proof of Theorem \ref{thm:feasible_condition_with_fidelity}}\label{sec:appendix_1}
For the convenience and conciseness of the proof, as long as there is no confusion, we abuse some notations in this section and the following Appendix sections. All notations in the following Appendix sections only follow their definitions in this section.

Recall that an optimization subproblem in \eqref{eq:OPT_sub} is formulated over a quasi-identifier (QID) group ${T_{{\bf{x}}_{\CMcal{U}}}}$ in which all public records are equal to ${{\bf{x}}_{\CMcal{U}}}$. Let $m = | {T_{{\bf{x}}_{\CMcal{U}}}} |$ be the cardinality of the QID group, or equivalently, the number of rows of this tuple. Let ${{\bf{x}}_{k}}$ be the unique record of row $k$ in the tuple, $k = 1, \ldots, m$, and define $p_k \triangleq P({{\bf{x}}_{k}})$, $x_k \triangleq {{{\tilde{D}_1}}({{\bf{x}}_{k}})}$, and $y_k \triangleq {{{\tilde{D}_0}}({{\bf{x}}_{k}})} = 1 - x_k$. The privacy constraints can thus be re-written as
\begin{flalign*}
{ \frac{ {p_k}{x_k} }{ \sum_{i=1}^{m} {p_i}{x_i} } } &\le \beta, \quad \forall k = 1, \ldots, m, \\
{ \frac{ {p_k}{y_k} }{ \sum_{i=1}^{m} {p_i}{y_i} } } &\le \beta, \quad \forall k = 1, \ldots, m,
\end{flalign*}
which can be combined as
\begin{equation}
\label{eq:privacy_constraints_simplified}
{p_k - \beta \sum_{i=1}^{m} {p_i} } \le { {p_k}{x_k} - \beta \sum_{i=1}^{m} {p_i}{x_i} } \le 0, 
\end{equation}
$\forall k = 1, \ldots, m$. Moreover, let ${\bf{x}} = [x_1, x_2, \cdots, x_m]^{T}$, where $T$ represents the transpose operator. Define ${\bf{A}}$ as
\begin{equation}
\label{eq:A}
{\bf{A}} = 
 \begin{pmatrix}
  (1-\beta)p_1 & {-\beta}p_2 & \cdots & {-\beta}p_m \\
  {-\beta}p_1 & (1-\beta)p_2 & \cdots & {-\beta}p_m \\
  \vdots  & \vdots  & \ddots & \vdots  \\
  {-\beta}p_1 & {-\beta}p_2 & \cdots & (1-\beta)p_m 
 \end{pmatrix},
\end{equation}
and let ${\bf{b}} = [b_1, b_2, \cdots, b_m]^{T}$, in which $b_k = {p_k - \beta \sum_{i=1}^{m} {p_i} }$. We can further simplify the privacy constraints as 
\begin{equation}
\label{eq:privacy_constraints_matrix_form}
{\bf{b}} \preceq {\bf{A}}{\bf{x}} \preceq {\bf{0}},
\end{equation}
where ${\bf{0}}$ is an $m \times 1$ zero vector.

\begin{rmk}
\label{rmk:b_k}
Note that $b_k = {p_k - \beta \sum_{i=1}^{m} {p_i} } \le 0$ due to Lemma \ref{lem:feasible_condition_no_fidelity}, or \eqref{eq:privacy_constraints_eqv_sum}, equivalently.
\end{rmk}

Similarly, the fidelity constraints can be re-written as 
\begin{flalign*}
{x_k}_{\text{min}} \le {x_k} \le {x_k}_{\text{max}}&, \quad \forall k = 1, \ldots, m, \\
{y_k}_{\text{min}} \le {y_k} \le {y_k}_{\text{max}}&, \quad \forall k = 1, \ldots, m. 
\end{flalign*}
However, since for binary decision, $y_k = 1 - x_k$, the above two constraints are basically equivalent (to see this, simply let ${{y_k}_{\text{min}}} = 1- {{x_k}_{\text{max}}}$ and ${{y_k}_{\text{max}}} = 1- {{x_k}_{\text{min}}}$), so we obtain the following fidelity constraints
\begin{equation}
\label{eq:fidelity_constraints_simplified}
{x_k}_{\text{min}} \le {x_k} \le {x_k}_{\text{max}}, \quad \forall k = 1, \ldots, m.
\end{equation}
Note that the $2m$ privacy constraints in \eqref{eq:privacy_constraints_simplified} (or their equivalent vectorized form in \eqref{eq:privacy_constraints_matrix_form}) form(s) a parallelotope in the $m$-dimensional space, and the $2m$ fidelity constraints in \eqref{eq:fidelity_constraints_simplified} form a hypercube in the $m$-dimensional space. Let $\CMcal{P}$ denote the parallelotope and $\CMcal{H}$ denote the
hypercube. Moreover, define $\CMcal{I} \triangleq \CMcal{P} \bigcap \CMcal{H}$ be the intersection of $\CMcal{P}$ and $\CMcal{H}$. $\CMcal{I} = \varnothing$ if and only if $\CMcal{P}$ and $\CMcal{H}$ are disjoint, where $\varnothing$ denotes the empty set. We have the following fact.

\begin{fact}
\label{fact:P&H}
An optimization subproblem has feasible solutions \textit{if and only if} $\CMcal{I} \neq \varnothing$, i.e., $\CMcal{P}$ and $\CMcal{H}$ intersect/collide with each other.
\end{fact}

To prove Theorem \ref{thm:feasible_condition_with_fidelity}, based on the above fact, it is hence equivalent to show that $\CMcal{P}$ and $\CMcal{H}$ collide with each other \textit{if and only if} $\beta \ge {\beta^{\text{*}}_{T_{{\bf{x}}_{\CMcal{U}}}}} \triangleq \max \{ {{\beta}_0}, {{\beta}_1}, {{\beta}_p} \}$. Let $\pi \triangleq {\argmax_k} { {p_k}{{x_k}_{\text{min}}} }$ and $\theta \triangleq {\argmax_k} { {p_k}{{y_k}_{\text{min}}} }$, we can re-write ${{\beta}_0}$, ${{\beta}_1}$, and ${{\beta}_p}$ in the following
\begin{flalign}
{{\beta}_0} &= \frac{ {p_{\theta}}{{y_{\theta}}_{\text{min}}} }{ { {p_{\theta}}{{y_{\theta}}_{\text{min}}} } + {\sum_{\substack{i=1 \\ i \neq {\theta}}}^{m}} {{p_i} {{y_i}_{\text{max}'}}} } \text{,} \label{eq:Beta_3} \\
{{\beta}_1} &= \frac{ {p_{\pi}}{{x_{\pi}}_{\text{min}}} }{ { {p_{\pi}}{{x_{\pi}}_{\text{min}}} } + {\sum_{\substack{i=1 \\ i \neq {\pi}}}^{m}} {{p_i} {{x_i}_{\text{max}'}}} } \text{,} \label{eq:Beta_1} \\
{{\beta}_p} &= \frac{ { {p_{\pi}}{{x_{\pi}}_{\text{min}}} } + { {p_{\theta}}{{y_{\theta}}_{\text{min}}} } }{ \sum_{i=1}^{m} {p_i} } \text{,} \label{eq:Beta_2}
\end{flalign}
where 
\begin{flalign}
{{x_i}_{\text{max}'}} &\triangleq \min \{ {{x_i}_{\text{max}}}, { \frac{p_{\pi}}{p_i} {{x_{\pi}}_{\text{min}}} } \}, \label{eq:x_i_max} \\
{{y_i}_{\text{max}'}} &\triangleq \min \{ {{y_i}_{\text{max}}}, { \frac{p_{\theta}}{p_i} {{y_{\theta}}_{\text{min}}} } \}.  \label{eq:y_i_max}
\end{flalign}

Consider the following two optimization problems for $x_j$, where $j$ is an arbitrary index, $1 \le j \le m$:
\begin{align*}
    \text{minimize}
        \quad & x_j  \label{eq:OPT_1} \tag{OPT-1} \\
    \text{s.t.} 
        \quad & {\bf{b}} \preceq {\bf{A}}{\bf{x}} \preceq {\bf{0}}, \\
        \quad & {x_k}_{\text{min}} \le {x_k} \le {x_k}_{\text{max}}, \text{for } k=1, \ldots, m, k \neq j.
\end{align*}
\begin{align*}
    \text{maximize}
        \quad & x_j  \label{eq:OPT_2} \tag{OPT-2} \\
    \text{s.t.} 
        \quad & {\bf{b}} \preceq {\bf{A}}{\bf{x}} \preceq {\bf{0}}, \\
        \quad & {x_k}_{\text{min}} \le {x_k} \le {x_k}_{\text{max}}, \text{for } k=1, \ldots, m, k \neq j.
\end{align*}
The above two problems have exactly the same constraints. The first line constraint forms the parallelotope $\CMcal{P}$, and let ${\CMcal{H}}'_j$ denote the hypercube formed by the second line constraints, i.e., ${x_k}_{\text{min}} \le {x_k} \le {x_k}_{\text{max}}$, for $k=1, \ldots, m, k \neq j$. In addition, define ${\CMcal{I}}'_j \triangleq \CMcal{P} \bigcap {\CMcal{H}}'_j$ be the intersection of $\CMcal{P}$ and ${\CMcal{H}}'_j$, interpreting the geometric space formed by the constraints of the above two optimization problems. Moreover, if ${\CMcal{I}}'_j \neq \varnothing$, (i.e., there exist feasible solutions for \eqref{eq:OPT_1} and \eqref{eq:OPT_2}), we let ${x_j^\dagger}$ and ${x_j^\ddagger}$ denote the optimal objective values of \eqref{eq:OPT_1} and \eqref{eq:OPT_2}, respectively. We have the following lemma.

\begin{lem}
\label{lem:Eqv_statement}
If ${\CMcal{I}}'_k \neq \varnothing$ for all $k = 1, \cdots, m$, $\CMcal{P}$ and $\CMcal{H}$ are disjoint ($\CMcal{I} = \varnothing$) \textit{if and only if} either ${x_j^\dagger} > {x_j}_{\text{max}}$ or ${x_j^\ddagger} < {x_j}_{\text{min}}$. In other words, $\CMcal{P}$ and $\CMcal{H}$ collide with each other \textit{if and only if} ${\CMcal{I}}'_k \neq \varnothing$, ${x_k^\dagger} \le {x_k}_{\text{max}}$, and ${x_k^\ddagger} \ge {x_k}_{\text{min}}$, $\forall k = 1, \cdots, m$.
\end{lem}
\begin{proof}
Apparently, since $\CMcal{H} = {{\CMcal{H}}'_j} \bigcap {{\CMcal{H}}'_k}$ for any $k \neq j$, we have $\CMcal{I} \subseteq {\CMcal{I}}'_j$ true for any $j$, which implies if there exists any $j$ such that ${\CMcal{I}}'_j = \varnothing$, $\CMcal{I} = \varnothing$, and $\CMcal{P}$ and $\CMcal{H}$ must be disjoint. Since $\CMcal{I} \subseteq {\CMcal{I}}'_j$ for every $j$, if ${\bf{x}} \notin {\CMcal{I}}'_j$ for any $j$, then ${\bf{x}} \notin \CMcal{I}$. Moreover, for any point ${\bf{x}} \in {\CMcal{I}}'_j$, ${x_j^\dagger} \le x_j \le {x_j^\ddagger}$.

If ${\CMcal{I}}'_k \neq \varnothing$ for all $k = 1, \cdots, m$, and either ${x_j^\dagger} > {x_j}_{\text{max}}$ or ${x_j^\ddagger} < {x_j}_{\text{min}}$, since for any ${\bf{x}} \in {\CMcal{I}}'_j$, ${x_j^\dagger} \le x_j \le {x_j^\ddagger}$, which implies either $x_j < {x_j}_{\text{min}}$ or $x_j > {x_j}_{\text{max}}$, and thus either $\CMcal{I} = \varnothing$, or $\CMcal{I} \nsubseteq {\CMcal{I}}'_j$ (which violates the truth). Therefore, $\CMcal{P}$ and $\CMcal{H}$ are disjoint.

We next prove the converse. If ${\CMcal{I}}'_k \neq \varnothing$, ${x_k^\dagger} \le {x_k}_{\text{max}}$, and ${x_k^\ddagger} \ge {x_k}_{\text{min}}$ are true for all $k = 1, \cdots, m$, since for any ${\bf{x}} \in {\CMcal{I}}'_k$, $\forall k = 1, \cdots, m$, ${x_k^\dagger} \le x_k \le {x_k^\ddagger}$, we have ${x_k}_{\text{min}} \le x_k \le {x_k}_{\text{max}}$, $\forall k$, which implies ${\bf{x}} \in \CMcal{I}$, so that $\CMcal{I} \neq \varnothing$, $\CMcal{P}$ and $\CMcal{H}$ collide with each other. We thus prove the converse and the proof is done.
\end{proof}

Based on Fact \ref{fact:P&H} and Lemma \ref{lem:Eqv_statement}, following statements are equivalent.
\begin{alignat*}{2}
     & \text{(S1) An optimization sub-problem has feasible solutions.} \\
\iff & \text{(S2) $\CMcal{P}$ and $\CMcal{H}$ intersect/collide with each other.}  \\
\iff & \text{(S3) \eqref{eq:OPT_1} and \eqref{eq:OPT_2} have feasible solutions \emph{for \textbf{all}} $j$.}  \\
\iff & \text{(S4) ${\CMcal{I}}'_j \neq \varnothing$, ${x_j^\dagger} \le {x_j}_{\text{max}}$ and ${x_j^\ddagger} \ge {x_j}_{\text{min}}$, $\forall j = 1, \cdots, m$.}
\end{alignat*}

\vspace{1em}
Our next goal is to show that (S1)$\simnot$(S4) are true \textit{if and only if} $\beta \ge \max \{ {{\beta}_0}, {{\beta}_1}, {{\beta}_p} \}$. To show this, we need the following lemma.

\begin{lem}
\label{lem:x_dagger}
Consider the optimization problem \eqref{eq:OPT_1} for some (arbitrary) $j$. If (S1)$\simnot$(S4) are true, we have $\beta \ge \max \{ {{\beta}_0}, {{\beta}_1}, {{\beta_p}_j} \}$, where
\begin{alignat}{2}
{\beta_0} &= \frac{ {p_{\theta}}{{y_{\theta}}_{\text{min}}} }{ { {p_{\theta}}{{y_{\theta}}_{\text{min}}} } + {\sum_{\substack{k=1 \\ k \neq {\theta}}}^{m}} {{p_k} {{y_k}_{\text{max}'}}} } \text{,} \label{eq:beta_3} \\
{\beta_1} &= \frac{ {p_{\pi}}{{x_{\pi}}_{\text{min}}} }{ { {p_{\pi}}{{x_{\pi}}_{\text{min}}} } + {\sum_{\substack{k=1 \\ k \neq {\pi}}}^{m}} {{p_k} {{x_k}_{\text{max}'}}} } \text{,} \label{eq:beta_1} \\
{{\beta_p}_j} &= \frac{ { {p_{\pi}}{{x_{\pi}}_{\text{min}}} } + { {p_{j}}{{y_{j}}_{\text{min}}} } }{ \sum_{k=1}^{m} {p_k} } \text{.} \label{eq:beta_2_j} 
\end{alignat}

\vspace{1em}
For each of the above cases, i.e., $\beta = {{\beta}_0} \text{, } {{\beta}_1} \text{, or } {{\beta_p}_j}$, the corresponding optimal objective value ${x_j^{\dagger}}$ and its corresponding optimal solutions are

\begin{alignat*}{2}
\beta = {{\beta}_0} 
\iff &{x_j^{\dagger}} = \frac{1}{p_j} \Big\{ {p_j} - \frac{\beta}{1-\beta} \sum_{\substack{k=1 \\ k \neq j}}^{m} {p_k}{{y_k}_{\text{max}'}} \Big\} \triangleq {x_j^{\dagger0}} \\
\iff &{y_j} = {y_{\theta}} = {{y_{\theta}}_{\text{min}}} \\
&{y_k} = {{y_k}_{\text{max}'}}, {\forall k = 1, \cdots, m}, \text{ } {k \neq {\theta}} \text{,} \\
\beta = {{\beta}_1} 
\iff &{x_j^{\dagger}} = \frac{1}{p_j} \Big\{ \frac{1-\beta}{\beta}{p_{\pi}}{x_{\pi}}_{\text{min}} - \sum_{\substack{k=1 \\ k \neq j, {\pi}}}^{m} {p_k}{{x_k}_{\text{max}'}} \Big\} \triangleq {x_j^{\dagger1}} \\
\iff &{x_{\pi}} = {{x_{\pi}}_{\text{min}}} \\
&{x_k} = {{x_k}_{\text{max}'}}, {\forall k = 1, \cdots, m}, \text{ } {k \neq {j, \pi}} \text{,} \\
\beta = {{\beta_p}_j} 
\iff &{x_j^{\dagger}} = \frac{1}{p_j} \Big\{ {p_{\pi}}{x_{\pi}}_{\text{min}} + {p_j - \beta \sum_{k=1}^{m} {p_k} } \Big\} \triangleq {x_j^{\dagger p}} \\
\iff &{x_{\pi}} = {{x_{\pi}}_{\text{min}}}  \\
&{\sum_{\substack{k=1 \\ k \neq {j, \pi}}}^{m}}{{p_k}{x_k}} = \frac{1-2\beta}{\beta} {p_{\pi}}{{x_{\pi}}_{\text{min}}} - p_j + \beta \sum_{k=1}^{m} {p_k} \text{.}
\end{alignat*}

\comment{
the optimal objective value of the optimization problem \eqref{eq:OPT_1} is ${x_j^\dagger} = \max \{ {x_j^{\dagger0}}, {x_j^{\dagger1}}, {x_j^{\dagger p}} \}$, where
\begin{flalign}
{x_j^{\dagger0}} &= \frac{1}{p_j} \Big\{ {p_j} - \frac{\beta}{1-\beta} \sum_{\substack{k=1 \\ k \neq j}}^{m} {p_k}{{y_k}_{\text{max}'}} \Big\}, \label{eq:x_j_d3} \\
{x_j^{\dagger1}} &= \frac{1}{p_j} \Big\{ \frac{1-\beta}{\beta}{p_{\pi}}{x_{\pi}}_{\text{min}} - \sum_{\substack{k=1 \\ k \neq j, {\pi}}}^{m} {p_k}{{x_k}_{\text{max}'}} \Big\}, \label{eq:x_j_d1} \\
{x_j^{\dagger p}} &= \frac{1}{p_j} \Big\{ {p_{\pi}}{x_{\pi}}_{\text{min}} + {p_j - \beta \sum_{k=1}^{m} {p_k} } \Big\}, \label{eq:x_j_d2}
\end{flalign}
and the corresponding optimal solutions are:
\begin{alignat*}{2}
{x_j^{\dagger}} = {x_j^{\dagger0}} \iff {y_{\theta}} &= {{y_{\theta}}_{\text{min}}} \\
{y_k} &= {{y_k}_{\text{max}'}}, {\forall k = 1, \cdots, m}, \text{ } {k \neq {j, \theta}}  \\
{x_j^{\dagger}} = {x_j^{\dagger1}} \iff {x_{\pi}} &= {{x_{\pi}}_{\text{min}}} \\
{x_k} &= {{x_k}_{\text{max}'}}, {\forall k = 1, \cdots, m}, \text{ } {k \neq {j, \pi}}  \\
{x_j^{\dagger}} = {x_j^{\dagger p}} \iff {x_{\pi}} &= {{x_{\pi}}_{\text{min}}}  \\
{\sum_{\substack{k=1 \\ k \neq {j, \pi}}}^{m}}{{p_k}{x_k}} &= \frac{1-2\beta}{\beta} {p_{\pi}}{{x_{\pi}}_{\text{min}}} - p_j + \beta \sum_{k=1}^{m} {p_k}.
\end{alignat*}

If (S1)$\simnot$(S4) are true, ${x_j^\dagger} \le {x_j}_{\text{max}}$, which implies ${x_j^{{\dagger}h}} \le {x_j}_{\text{max}}$, $\forall h = 0, 1, p$, and we have
\begin{flalign}
{x_j^{\dagger}} = {x_j^{\dagger0}} \iff {{\beta}} &\ge \frac{ {p_{\theta}}{{y_{\theta}}_{\text{min}}} }{ { {p_{\theta}}{{y_{\theta}}_{\text{min}}} } + {\sum_{\substack{k=1 \\ k \neq {\theta}}}^{m}} {{p_k} {{y_k}_{\text{max}'}}} } = {\beta_0} \text{,} \label{eq:beta_3} \\
{x_j^{\dagger}} = {x_j^{\dagger1}} \iff {{\beta}} &\ge \frac{ {p_{\pi}}{{x_{\pi}}_{\text{min}}} }{ { {p_{\pi}}{{x_{\pi}}_{\text{min}}} } + {\sum_{\substack{k=1 \\ k \neq {\pi}}}^{m}} {{p_k} {{x_k}_{\text{max}'}}} } = {\beta_1} \text{,} \label{eq:beta_1} \\
{x_j^{\dagger}} = {x_j^{\dagger p}} \iff {{\beta}} &\ge \frac{ { {p_{\pi}}{{x_{\pi}}_{\text{min}}} } + { {p_{j}}{{y_{j}}_{\text{min}}} } }{ \sum_{k=1}^{m} {p_k} } \triangleq {{\beta_p}_j} \text{.} \label{eq:beta_2_j} 
\end{flalign}
}

\end{lem}
\begin{proof}
Please refer to Appendix \ref{sec:appendix_2} for the proofs.
\end{proof}

When (S1)$\simnot$(S4) are true, ${\CMcal{I}}'_j \neq \varnothing$ and ${x_j^\dagger} \le {x_j}_{\text{max}}$ need to be met \emph{for all} $j = 1, \ldots, m$. Based on Lemma \ref{lem:x_dagger}, it implies that $\beta \ge {{\beta}_0}$, $\beta \ge {{\beta}_1}$, and 
\begin{equation*}
\beta \ge {{\beta_p}_j} = \frac{ { {p_{\pi}}{{x_{\pi}}_{\text{min}}} } + { {p_{j}}{{y_{j}}_{\text{min}}} } }{ \sum_{k=1}^{m} {p_k} }, \quad \forall j = 1, \ldots, m,
\end{equation*}
which is equivalent to
\begin{flalign*}
\beta \ge \underset{j}{\max} \frac{ { {p_{\pi}}{{x_{\pi}}_{\text{min}}} } + { {p_{j}}{{y_{j}}_{\text{min}}} } }{ \sum_{k=1}^{m} {p_k} } 
       = \frac{ { {p_{\pi}}{{x_{\pi}}_{\text{min}}} } + { {p_{\theta}}{{y_{\theta}}_{\text{min}}} } }{ \sum_{k=1}^{m} {p_k} } = {{\beta}_p}.
\end{flalign*}
We then obtain $\beta \ge \max \{ {{\beta}_0}, {{\beta}_1}, {{\beta}_p} \} = {\beta^{\text{*}}_{T_{{\bf{x}}_{\CMcal{U}}}}}$. Since when ${\beta^{\text{*}}_{T_{{\bf{x}}_{\CMcal{U}}}}} = {\beta_0}$, based on Lemma \ref{lem:x_dagger}, we have ${y_j} = {y_{\theta}} = {{y_{\theta}}_{\text{min}}}$. Based on \eqref{eq:beta_1} and \eqref{eq:beta_2_j}, we have
\begin{flalign*}
{\beta^{\text{*}}_{T_{{\bf{x}}_{\CMcal{U}}}}} = {\beta_1} \iff {x_j} &= {{x_j}_{\text{max}'}} \text{,} \\
{\beta^{\text{*}}_{T_{{\bf{x}}_{\CMcal{U}}}}} = {\beta_p} \iff {y_j} &= {y_{\theta}} = {{y_{\theta}}_{\text{min}}} \text{.}
\end{flalign*}
Combining the above results with Lemma \ref{lem:x_dagger}, we thus have
\begin{alignat*}{2}
{\beta^{\text{*}}_{T_{{\bf{x}}_{\CMcal{U}}}}} = {\beta_0}  \iff {y_{\theta}} &= {{y_{\theta}}_{\text{min}}} \\
{y_k} &= {{y_k}_{\text{max}'}}, {\forall k = 1, \cdots, m}, \text{ } {k \neq {\theta}}  \\
{\beta^{\text{*}}_{T_{{\bf{x}}_{\CMcal{U}}}}} = {\beta_1} \iff {x_{\pi}} &= {{x_{\pi}}_{\text{min}}} \\
{x_k} &= {{x_k}_{\text{max}'}}, {\forall k = 1, \cdots, m}, \text{ } {k \neq {\pi}}  \\
{\beta^{\text{*}}_{T_{{\bf{x}}_{\CMcal{U}}}}} = {\beta_p} \iff {x_{\pi}} &= {{x_{\pi}}_{\text{min}}}  \\
{y_{\theta}} &= {{y_{\theta}}_{\text{min}}} \\
{\sum_{\substack{k=1 \\ k \neq {\theta, \pi}}}^{m}}{{p_k}{x_k}} &= \frac{1-2\beta}{\beta} {p_{\pi}}{{x_{\pi}}_{\text{min}}} - p_j + \beta \sum_{k=1}^{m} {p_k}.
\end{alignat*}

Similarly, if (S1)$\simnot$(S4) are true, ${\CMcal{I}}'_j \neq \varnothing$ and ${x_j^\ddagger} \ge {x_j}_{\text{min}}$ need to be met for all $j = 1, \ldots, m$. By letting $y_k = 1 - x_k$, ${{y_k}_{\text{min}}} = 1- {{x_k}_{\text{max}}}$, and ${{y_k}_{\text{max}}} = 1- {{x_k}_{\text{min}}}$, the optimization problem \eqref{eq:OPT_2} is essentially equivalent to the following optimization problem:
\begin{align*}
    \text{minimize}
        \quad & y_j  \label{eq:OPT_3} \tag{OPT-3} \\
    \text{s.t.} 
        \quad & {\bf{b}} \preceq {\bf{A}}{\bf{y}} \preceq {\bf{0}}, \\
        \quad & {y_k}_{\text{min}} \le {y_k} \le {y_k}_{\text{max}}, \text{for } k=1, \ldots, m, k \neq j.
\end{align*}
Let ${y_j^\dagger}$ be the optimal objective value of the above optimization problem. Clearly, ${y_j^\dagger} = 1 - {x_j^\ddagger}$. Therefore, for all $j = 1, \ldots, m$, ${x_j^\ddagger} \ge {x_j}_{\text{min}}$ is equivalent to ${y_j^\dagger} \le {y_j}_{\text{max}}$. By applying results from ${x_j^\dagger}$ in Lemma \ref{lem:x_dagger}, we will obtain exactly the same conditions for $\beta$, i.e., $\beta \ge {\beta^{\text{*}}_{T_{{\bf{x}}_{\CMcal{U}}}}}$. Therefore, if (S1)$\simnot$(S4) are true, we have $\beta \ge \max \{ {{\beta}_0}, {{\beta}_1}, {{\beta}_p} \}$.


We next prove the converse. If $\beta \ge \max \{ {{\beta}_0}, {{\beta}_1}, {{\beta}_p} \}$, which, based on \eqref{eq:beta_3}, \eqref{eq:beta_1}, and \eqref{eq:beta_2_j}, implies ${x_j^\dagger} \le {{x_j}_{\text{max}}}$ and ${y_j^\dagger} \le {{y_j}_{\text{max}}}$, which is equivalent to ${x_j^\ddagger} \ge {x_j}_{\text{min}}$, $\forall j = 1, \cdots, m$. Therefore, based on Lemma \ref{lem:x_dagger}, ${x_j}$ is feasible, which implies ${\CMcal{I}}'_j \neq \varnothing$, $\forall j = 1, \cdots, m$, and thus (S4) is true. Since (S1)$\simnot$(S4) are equivalent, an optimization sub-problem has feasible solutions \emph{if and only if} $\beta \ge {\beta^{\text{*}}_{T_{{\bf{x}}_{\CMcal{U}}}}} = \max \{ {{\beta}_0}, {{\beta}_1}, {{\beta}_p} \}$. We thus finish the proof.

\section{Proof of Lemma \ref{lem:x_dagger}}\label{sec:appendix_2}
Here we demonstrate the proof of Lemma \ref{lem:x_dagger}, which shows the optimal objective value of the optimization problem \eqref{eq:OPT_1}.

If ${\CMcal{I}}'_j \neq \varnothing$, there exists (at least one or some) ${\bf{x}} \in {\CMcal{I}}'_j$, and for all ${\bf{x}}$, ${x_j^\dagger} \le x_j \le {x_j^\ddagger}$. Since ${\CMcal{I}}'_j = \CMcal{P} \bigcap {\CMcal{H}}'_j$, any ${\bf{x}} \in {\CMcal{I}}'_j$ also belongs to $\CMcal{P}$ and ${\CMcal{H}}'$. Since $\CMcal{P}$ is a $m$-dimensional parallelotope, and ${\bf 0} \in \CMcal{P}$ is a vertex of $\CMcal{P}$, any point ${\bf x} \in \CMcal{P}$ can be uniquely represented by a linear combination of $m$ linear independent edge vectors emitted from ${\bf 0}$, denoted by ${{\bf{L}}_k}$, $k = 1, \ldots, m$, and ${\bf x} = \sum_{k=1}^{m} \alpha_k {{\bf{L}}_k}$, $0 \le \alpha_k \le 1$, $\forall k=1, \ldots, m$. Let ${\bf{L}}$ be the collection of these $m$ vectors; specifically, ${\bf{L}} \triangleq [{\bf{L}}_1 {\bf{L}}_2 \cdots {\bf{L}}_m]$, where ${\bf{L}}_k$ is an $m \times 1$ column vector and ${\bf{L}}$ is an $m \times m$ matrix. ${\bf{L}}$ can be obtained by
\begin{equation}
{\bf{L}} = {\bf{A}}^{-1}{\bf{B}},
\end{equation}
where ${\bf{A}}$ is defined in \eqref{eq:A} and ${\bf{B}} = \text{dg}({\bf{b}})$ where $\text{dg}({\bf{b}})$ denotes a diagonal matrix with elements of ${\bf{b}} = (b_1, b_2, \cdots, b_m)$ along the diagonal. To find ${\bf{A}}^{-1}$, note that since ${\bf{A}}$ can be represented by 
\begin{equation}
{\bf{A}} = \text{dg}({\bf{p}}) + (-\beta){{\bf{1}}_m}{{\bf{p}}^{T}},
\end{equation}
where ${\bf{p}} = [p_1, p_2, \cdots, p_m]^{T}$ and ${{\bf{1}}_m}$ is an all-one vector with $m$ elements, we can thus apply the following matrix inversion formula \cite{henderson1981deriving}
\begin{equation}
\label{eq:matrix_inverse}
({\bf{Z}}+c{\bf{u}}{{\bf{v}}^{T}})^{-1} = {\bf{Z}}^{-1} - \frac{1}{1+c{{\bf{v}}^{T}}{{\bf{Z}}^{-1}}{\bf{u}}} {{\bf{Z}}^{-1}}{\bf{u}}{{\bf{v}}^{T}}{{\bf{Z}}^{-1}}
\end{equation}
to compute ${\bf{A}}^{-1}$ and obtain ${\bf{L}}$ as follow
\begin{equation}
\label{eq:L}
{\bf{L}} = \small{ \frac{1}{1-m\beta} } 
   \setlength{\arraycolsep}{2pt}
   \renewcommand{\arraystretch}{0.8}
 \begin{pmatrix}
  \frac{b_1}{p_1}[ \scriptstyle{1-(m-1)\beta} ] & \frac{b_2}{p_1}\beta & \cdots & \frac{b_m}{p_1}\beta \\
  \frac{b_1}{p_2}\beta & \frac{b_2}{p_2}[ \scriptstyle{1-(m-1)\beta} ] & \cdots & \frac{b_m}{p_2}\beta \\
  \vdots  & \vdots  & \ddots & \vdots  \\
  \frac{b_1}{p_m}\beta & \frac{b_2}{p_m}\beta & \cdots & \frac{b_m}{p_m}[ \scriptstyle{1-(m-1)\beta} ] 
 \end{pmatrix} \normalsize \text{.}
\end{equation}
Define $\frac{{\bf{b}}}{{\bf{p}}} \triangleq (\frac{b_1}{p_1}, \frac{b_2}{p_2}, \cdots, \frac{b_m}{p_m})$ as the element-wise division operation of two vectors. It is not hard to see that ${\bf{L}}$ can be represented as the following equivalent form
\begin{equation}
\label{eq:L_eqv}
{\bf{L}} = \text{dg}\Big(\frac{{\bf{b}}}{{\bf{p}}}\Big) + {\frac{\beta}{1-m\beta}}{\frac{{\bf{1}}}{{\bf{p}}}}{{\bf{b}}^{T}},
\end{equation}
which implies that its inverse can also be found by applying the matrix inversion formula in \eqref{eq:matrix_inverse}. We will utilize this property in the later of the proof.

Recall that if ${\bf x} \in {\CMcal{I}}'_j$, ${\bf x} \in {\CMcal{P}}$ as well. Therefore, any ${\bf x} \in {\CMcal{I}}'_j$ can be uniquely represented by
\begin{equation}
\label{eq:x_representation}
{\bf x} = \sum_{k=1}^{m} \alpha_k {{\bf{L}}_k}  \text{,}
\end{equation} 
in which $0 \le \alpha_k \le 1$, $\forall k=1, \ldots, m$. Recall that we are solving the optimization problem \eqref{eq:OPT_1} for some $j$, $1 \le j \le m$. We first take out the $j$-th row from \eqref{eq:x_representation},
\begin{equation}
\label{eq:x_j}
x_j = \sum_{k=1}^{m} \alpha_k {L_{k, j}}  \text{,}
\end{equation} 
and for the rest $m-1$ equalities, we move the term $\alpha_j {{\bf{L}}_j}$ from the right-hand-side (RHS) to the left-hand-side (LHS). We then obtain
\begin{equation}
\label{eq:trick}
\begin{bmatrix}
x_1 \\
x_2 \\
\vdots \\
x_{j-1} \\
x_{j+1} \\
\vdots \\
x_{m}
\end{bmatrix}
- {{\alpha}_j}
\begin{bmatrix}
L_{1, j} \\
L_{2, j} \\
\vdots \\
L_{j-1, j} \\
L_{j+1, j} \\
\vdots \\
L_{m, j}
\end{bmatrix} = 
\begin{bmatrix}
L_{1, 1} & L_{1, 2} & \cdots & L_{1, j-1} & L_{1, j+1} & \cdots & L_{1, m} \\
L_{2, 1} & L_{2, 2} & \cdots & L_{2, j-1} & L_{2, j+1} & \cdots & L_{2, m} \\
\vdots & \vdots & \ddots & \vdots & \vdots & \ddots & \vdots \\
L_{j-1, 1} & L_{j-1, 2} & \cdots & L_{j-1, j-1} & L_{j-1, j+1} & \cdots & L_{j-1, m} \\
L_{j+1, 1} & L_{j+1, 2} & \cdots & L_{j+1, j-1} & L_{j+1, j+1} & \cdots & L_{j+1, m} \\
\vdots & \vdots & \ddots & \vdots & \vdots & \ddots & \vdots \\
L_{m, 1} & L_{m, 2} & \cdots & L_{m, j-1} & L_{m, j+1} & \cdots & L_{m, m} \\
\end{bmatrix}
\begin{bmatrix}
{\alpha}_1 \\
{\alpha}_2 \\
\vdots \\
{\alpha}_{j-1} \\
{\alpha}_{j+1} \\
\vdots \\
{\alpha}_{m}
\end{bmatrix} ,
\end{equation} 
and let ${{\bf{x}}'} - {{\alpha}_j} {{\bf{L}}^{'}_{j}} = {{\bf{L}}_{\text{sub}}} {{\bm{\alpha}}'}$ be the corresponding vector form of \eqref{eq:trick}, in which, based on \eqref{eq:L}, ${L_{k,k}} = {\frac{1-(m-1)\beta}{1-m\beta}}{\frac{b_k}{p_k}}$, $\forall k = 1, \cdots, m$, and ${L_{k,i}} = {\frac{\beta}{1-m\beta}}{\frac{b_i}{p_k}}$, $\forall k,i = 1, \cdots, m$, $k \neq i$. Note that ${{\bf{L}}_{\text{sub}}}$ is an $(m-1) \times (m-1)$ square sub-matrix of ${\bf{L}}$ by removing the $j$-th row and the $j$-th column. Therefore, it has the similar form as shown in \eqref{eq:L_eqv} by removing the $j$-th row/element of $\bf{b}$ and $\bf{p}$, and thus its inverse ${{\bf{L}}^{-1}_{\text{sub}}}$ can also be found by \eqref{eq:matrix_inverse}. By applying ${{\bf{L}}^{-1}_{\text{sub}}}$ to both sides of \eqref{eq:trick}, we have
\begin{equation}
\label{eq:alpha_no_j}
{{\bm{\alpha}}'} = {{\bf{L}}^{-1}_{\text{sub}}}{{\bf{x}}'} - {{\alpha}_j}{{\bf{L}}^{-1}_{\text{sub}}} {{\bf{L}}^{'}_{j}}  \text{.}
\end{equation} 
Let ${\bf{u}} \triangleq {{\bf{L}}^{-1}_{\text{sub}}}{{\bf{x}}'}$ and ${\bf{v}} \triangleq {{\alpha}_j}{{\bf{L}}^{-1}_{\text{sub}}} {{\bf{L}}^{'}_{j}}$, we obtain
\begin{equation}
\label{eq:u_v_alpha}
\begin{split}
u_k &= {\frac{1}{1-\beta}} {\frac{1}{b_k}} {\Big[ (1-\beta){p_k}{x_k} - \beta {\sum_{\substack{i=1 \\ i \neq {j}}}^{m}}{{p_i}{x_i}} \Big]} \text{,} \\
v_k &= {\frac{\beta}{1-\beta}} {\frac{1}{b_k}} {{{\alpha}_j}{b_j}} \text{,} \\
{\alpha}_k &= u_k - v_k 
= {\frac{\beta}{1-\beta}} {\frac{1}{b_k}} { \Big[ {\frac{1-\beta}{\beta}} {p_k}{x_k} - {\sum_{\substack{i=1 \\ i \neq {j}}}^{m}}{{p_i}{x_i}} - {{\alpha}_j}{b_j} \Big] } \text{,}
\end{split}
\end{equation} 
for all $k = 1, \cdots, m$, $k \neq j$. Similarly, if we define ${{\alpha}_k}' \triangleq 1 - {\alpha}_k$, we have 
\begin{equation}
\label{eq:alpha_k_apostrophe}
{{\alpha}_k}'
= {\frac{\beta}{1-\beta}} {\frac{1}{b_k}} { \Big[ {\frac{1-\beta}{\beta}} {p_k}{y_k} - {\sum_{\substack{i=1 \\ i \neq {j}}}^{m}}{{p_i}{y_i}} - {{{\alpha}_j}'}{b_j} \Big] } \text{.}
\end{equation} 
Substituting the ${\alpha}_k$ in \eqref{eq:u_v_alpha} into \eqref{eq:x_j}, we obtain
\begin{equation}
\label{eq:x_j_2}
x_j = \sum_{k=1}^{m} \alpha_k {L_{k, j}}
= {\frac{1}{1-\beta}} {\frac{1}{p_j}} {\Big[ \beta {\sum_{\substack{k=1 \\ k \neq {j}}}^{m}}{{p_k}{x_k}} + {{\alpha}_j}{b_j} \Big]} \text{.}
\end{equation} 
Based on \eqref{eq:x_j_2}, we are looking for the values of ${\alpha}_j$ and $x_k$'s ($k \neq j$) yielding the minimum of $x_j$.

Since $0 \le \alpha_k \le 1$, $\forall k=1, \ldots, m$, which implies ${u_k} \ge {v_k}$, $\forall k=1, \ldots, m$, $k \neq j$, and
\begin{equation}
\label{eq:E_C}
\frac{u_k}{v_k} = {\frac{1}{\beta{\alpha_j}{b_j}}}{\Big[ (1-\beta){p_k}{x_k} - \beta {\sum_{\substack{i=1 \\ i \neq {j}}}^{m}}{{p_i}{x_i}} \Big]} \ge 1.
\end{equation} 
We then have 
\begin{equation}
\label{eq:alpha_j_UB}
{{\alpha}_j} \le {\frac{1}{\beta{b_j}}}{\Big[ (1-\beta){p_k}{x_k} - \beta {\sum_{\substack{i=1 \\ i \neq {j}}}^{m}}{{p_i}{x_i}} \Big]} \triangleq R_j^k,
\end{equation} 
for all $k=1, \ldots, m$, $k \neq j$. Let $R^{*}_j \triangleq \underset{k}{\min} R_j^k$. Combining with the fact that ${\alpha_j} \le 1$, we obtain 
\begin{equation}
\label{eq:alpha_j_UB_short}
{{\alpha}_j} \le \min(R^{*}_j,1).
\end{equation}

\vspace{3mm}
\noindent\textbf{Case 1}: 
\newline
We first consider the case $R^{*}_j \le 1$. Please refer to Case 2 for $R^{*}_j \ge 1$.

When $R^{*}_j \le 1$, based on \eqref{eq:alpha_j_UB_short}, we have ${{\alpha}_j} = R^{*}_j$.
Therefore, based on \eqref{eq:alpha_j_UB}, there must exist an $k$ (denoted by $\kappa$) such that
\begin{equation}
\label{eq:alpha_j_UB_2}
{{\alpha}_j} = {\frac{1}{\beta{b_j}}}{\Big[ (1-\beta){p_{\kappa}}{x_{\kappa}} - \beta {\sum_{\substack{i=1 \\ i \neq {j}}}^{m}}{{p_i}{x_i}} \Big]} = R^{*}_j \le 1.
\end{equation}
Because $b_j \le 0$ (see Remark \ref{rmk:b_k}), from the LHS of \eqref{eq:alpha_j_UB_2}, we have
\begin{equation}
\label{eq:critical_term}
{ \beta {\sum_{\substack{i=1 \\ i \neq {j}}}^{m}}{{p_i}{x_i}} + {{\alpha}_j}{b_j} } = (1-\beta){\Big[ {\frac{1}{\beta}}{p_{\kappa}}{x_{\kappa}} - {\sum_{\substack{i=1 \\ i \neq {j}}}^{m}}{{p_i}{x_i}} \Big]}.
\end{equation}
Substitute the LHS of \eqref{eq:critical_term} into \eqref{eq:x_j_2}. We obtain
\begin{equation}
\label{eq:x_j_ineq}
\begin{split}
{p_j}{x_j} = \frac{1}{\beta} {p_{\kappa}}{x_{\kappa}} - {\sum_{\substack{k=1 \\ k \neq {j}}}^{m}}{{p_i}{x_i}} 
= \frac{1-\beta}{\beta} {p_{\kappa}}{x_{\kappa}} - {\sum_{\substack{k=1 \\ k \neq {j, \kappa}}}^{m}}{{p_i}{x_i}},
\end{split}
\end{equation}
or equivalently,
\begin{equation}
\label{eq:beta_ineq}
\begin{split}
\beta = \frac{{p_{\kappa}}{x_{\kappa}}}{{\sum_{\substack{k=1}}^{m}}{{p_k}{x_k}}} = \frac{{p_{\kappa}}{x_{\kappa}}}{{{p_{\kappa}}{x_{\kappa}}} + {\sum_{\substack{k=1 \\ k \neq {\kappa}}}^{m}}{{p_k}{x_k}}}.
\end{split}
\end{equation}
Therefore, $x_j$ is minimized if the RHS of \eqref{eq:x_j_ineq} is minimized. In addition, in Case 1, based on \eqref{eq:beta_ineq}, $\beta$ achieves its minimum when (i) minimizing ${p_{\kappa}}{x_{\kappa}}$ and (ii) maximizing ${\sum_{\substack{k=1 \\ k \neq {\kappa}}}^{m}}{{p_k}{x_k}}$.

We next find the minimum of the RHS of \eqref{eq:x_j_ineq}. Since from \eqref{eq:alpha_j_UB_2}, we have
\begin{equation}
\label{eq:sum_pi_xi}
{\sum_{\substack{i=1 \\ i \neq {j}}}^{m}}{{p_i}{x_i}} =
{\sum_{\substack{k=1 \\ k \neq {j}}}^{m}}{{p_k}{x_k}} = \frac{1-\beta}{\beta} {p_{\kappa}}{x_{\kappa}} - {\alpha_j}{b_j},
\end{equation}
and from \eqref{eq:x_j_ineq}, we have
\begin{equation}
\label{eq:sum_pi_xi_2}
\beta {\sum_{\substack{k=1 \\ k \neq {j}}}^{m}}{{p_k}{x_k}} = {p_{\kappa}}{x_{\kappa}} - \beta {p_j}{x_j}.
\end{equation}
Substituting the LHS of \eqref{eq:sum_pi_xi} into \eqref{eq:u_v_alpha}, we obtain
\begin{equation}
\label{eq:alpha_k_noj}
{\alpha_k} = {\frac{1}{b_k}}({p_k}{x_k} - {p_{\kappa}}{x_{\kappa}}), \forall k = 1, \cdots, m, k \neq j,
\end{equation}
and substituting the LHS of \eqref{eq:sum_pi_xi} into \eqref{eq:sum_pi_xi_2}, we obtain
\begin{equation}
\label{eq:alpha_j}
{\alpha_j} = {\frac{1}{b_j}}({p_j}{x_j} - {p_{\kappa}}{x_{\kappa}}).
\end{equation}
Combining \eqref{eq:alpha_k_noj} and \eqref{eq:alpha_j}, we have
\begin{equation}
\label{eq:alpha_k}
{\alpha_k} = {\frac{1}{b_k}}({p_k}{x_k} - {p_{\kappa}}{x_{\kappa}}), \forall k = 1, \cdots, m.
\end{equation}
Since ${\alpha_k} \ge 0$ and $b_k \le 0$, we have ${p_{\kappa}}{x_{\kappa}} \ge {p_k}{x_k}$, $\forall k$, i.e., ${p_{\kappa}}{x_{\kappa}} = \underset{k}{\max} {p_k}{x_k}$. 
Moreover, from \eqref{eq:x_j_ineq}, since $\beta$ is non-negative
and $x_k \ge 0$ for all $k$, in order to minimize $x_j$, we need to (i) minimize ${p_{\kappa}}{x_{\kappa}}$ and (ii) maximize ${\sum_{\substack{k=1 \\ k \neq {j, \kappa}}}^{m}}{{p_k}{x_k}}$. Note that both (i) and (ii) minimize $\beta$ as well, which implies that the optimal solutions that minimize $x_j$ also minimize $\beta$.

To minimize ${p_{\kappa}}{x_{\kappa}}$, since ${p_{\kappa}}{x_{\kappa}} = \underset{k}{\max} {p_k}{x_k}$, and ${p_k}{x_k} \ge {p_k}{{x_k}_{\text{min}}}$, $\forall k$ (including $\kappa$), the minimal ${p_{\kappa}}{x_{\kappa}}$, i.e., ${p_{\kappa}}{{x_{\kappa}}_{\text{min}}}$, is therefore the largest effective lower limit ${p_k}{{x_k}_{\text{min}}}$ over all $k$, i.e., ${p_{\kappa}}{{x_{\kappa}}_{\text{min}}} = \underset{k}{\max} {p_k}{{x_k}_{\text{min}}} = {p_{\pi}}{{x_{\pi}}_{\text{min}}}$ by definition, and thus we get $\kappa = \pi$ and ${x_{\pi}} = {{x_{\pi}}_{\text{min}}}$.

To maximize ${\sum_{\substack{k=1 \\ k \neq {j, \kappa}}}^{m}}{{p_k}{x_k}}$, we need to find the maximum of each ${x_k}$. Since $0 \le {\alpha_k} \le 1$, by substituting ${p_{\kappa}}{x_{\kappa}} = {p_{\pi}}{{x_{\pi}}_{\text{min}}}$ into \eqref{eq:alpha_k}, we have ${p_{\pi}}{{x_{\pi}}_{\text{min}}} + {b_k} \le {p_k}{x_k} \le {p_{\pi}}{{x_{\pi}}_{\text{min}}}$, $\forall k = 1, \cdots, m$ (including $j$). 
By definitions, ${{x_k}_{\text{max}'}} \triangleq \min \{ {{x_k}_{\text{max}}}, { \frac{p_{\pi}}{p_k} {{x_{\pi}}_{\text{min}}} } \}$ and ${{x_k}_{\text{min}'}} \triangleq \max \{ {{x_k}_{\text{min}}}, { \frac{p_{\pi}}{p_k} {{x_{\pi}}_{\text{min}}} } + {b_k} \}$. 
Combining with the constraints ${{x_k}_{\text{min}}} \le {x_k} \le {{x_k}_{\text{max}}}$, we obtain ${{x_k}_{\text{min}'}} \le {x_k} \le {{x_k}_{\text{max}'}}$, $\forall k = 1, \cdots, m$, and thus ${\sum_{\substack{k=1 \\ k \neq {j, \kappa}}}^{m}}{{p_k}{x_k}}$ is maximized when ${x_k} = {{x_k}_{\text{max}'}}$, $\forall k$, $k \neq j, \pi$. 

By substituting the $x_k$'s we obtained above into \eqref{eq:x_j_ineq}, based on which, the optimal objective value ${x_j^{\dagger}}$, the minimum of $x_j$, is thus
\begin{equation}
\label{eq:x_j_opt1}
{x_j^{\dagger}} = \frac{1}{p_j} \Big\{ \frac{1-\beta}{\beta}{p_{\pi}}{x_{\pi}}_{\text{min}} - \sum_{\substack{k=1 \\ k \neq j, {\pi}}}^{m} {p_i}{{x_i}_{\text{max}'}} \Big\} \triangleq {x_j^{\dagger1}}.
\end{equation}
If (S1)$\simnot$(S4) are true, ${x_j^\dagger} \le {x_j}_{\text{max}}$. In addition, based on \eqref{eq:alpha_j}, we have ${p_j}{x_j} = {p_{\pi}}{{x_{\pi}}_{\text{min}}} + {\alpha_j}{b_j} \le {p_{\pi}}{{x_{\pi}}_{\text{min}}}$, and thus from \eqref{eq:x_i_max}, we get ${x_j}_{\text{max}} = {x_j}_{\text{max}'}$. 
Therefore, based on \eqref{eq:x_j_opt1}, the minimum $\beta$ such that \eqref{eq:OPT_1} has feasible solution, for Case 1, is
\begin{equation}
\label{eq:beta_case1}
\begin{split}
\beta 
&= \frac{ {p_{\pi}}{{x_{\pi}}_{\text{min}}} }{ { {p_{\pi}}{{x_{\pi}}_{\text{min}}} } + {p_j}{x_j^{\dagger}} + {\sum_{\substack{k=1 \\ k \neq {j, \pi}}}^{m}} {{p_k} {{x_k}_{\text{max}'}}} } \\
&\ge \frac{ {p_{\pi}}{{x_{\pi}}_{\text{min}}} }{ { {p_{\pi}}{{x_{\pi}}_{\text{min}}} } + {\sum_{\substack{k=1 \\ k \neq {\pi}}}^{m}} {{p_k} {{x_k}_{\text{max}'}}} } \triangleq {\beta_1}.
\end{split}
\end{equation}

We next prove the converse. If \eqref{eq:beta_case1} and \eqref{eq:x_j_opt1} holds, i.e., ${x_{\kappa}} = {x_{\pi}} = {{x_{\pi}}_{\text{min}}}$, and ${x_k} = {{x_k}_{\text{max}'}}$, $\forall k = 1, \cdots, m$, $k \neq j, \pi$,
since ${x_{\pi}}_{\text{min}} = {x_{\pi}}_{\text{max}'}$, we have ${x_k} = {{x_k}_{\text{max}'}}$, $\forall k = 1, \cdots, m$, $k \neq j$, and
\begin{equation}
\label{eq:p_j_x_j_opt1}
\begin{split}
{p_j}{x_j^{\dagger}} &=  \frac{1-\beta}{\beta}{p_{\pi}}{x_{\pi}}_{\text{min}} - \sum_{\substack{k=1 \\ k \neq j, {\pi}}}^{m} {p_k}{{x_k}_{\text{max}'}} \\ &= \frac{1}{\beta}{p_{\pi}}{x_{\pi}}_{\text{min}} - \sum_{\substack{k=1 \\ k \neq j {}}}^{m} {p_k}{{x_k}_{\text{max}'}}.
\end{split}
\end{equation}
Given the above $x_k$'s and \eqref{eq:p_j_x_j_opt1}, since ${x_{\pi}}_{\text{min}} = {x_{\pi}}_{\text{max}'}$ and $b_j \le 0$, and by definition in \eqref{eq:x_i_max}, ${p_{\pi}}{{x_{\pi}}_{\text{min}}} \ge {p_k}{{x_k}_{\text{max}'}}$, $\forall k$, we have
\begin{equation}
\label{eq:R_j_star}
\begin{split}
R^{*}_j 
& \triangleq \underset{k}{\min} R_j^k \\
&= \underset{k}{\min} {\frac{1}{\beta{b_j}}}{\Big[ (1-\beta){p_k}{x_{k}}_{\text{max}'} - \beta {\sum_{\substack{i=1 \\ i \neq {j}}}^{m}}{{p_i}{x_{i}}_{\text{max}'}} \Big]} \\
&= {\frac{1}{\beta{b_j}}} \underset{k}{\max} {\Big[ (1-\beta){p_k}{x_{k}}_{\text{max}'} - \beta {\sum_{\substack{i=1 \\ i \neq {j}}}^{m}}{{p_i}{x_{i}}_{\text{max}'}} \Big]} \\
&= {\frac{1}{\beta{b_j}}}{\Big[ (1-\beta){p_{\pi}}{{x_{\pi}}_{\text{min}}} - \beta {\sum_{\substack{i=1 \\ i \neq {j}}}^{m}}{{p_i}{x_{i}}_{\text{max}'}} \Big]} \\
&= {\frac{1}{b_j}}({p_j}{x_j^{\dagger}} - {p_{\pi}}{x_{\pi}}_{\text{min}}).
\end{split}
\end{equation}
Besides, by substituting ${x_k} = {{x_k}_{\text{max}'}}$, $\forall k = 1, \cdots, m$, $k \neq j$, into \eqref{eq:x_j_2}, we obtain
\begin{equation}
\label{eq:p_j_x_j_2}
{p_j}{x_j^{\dagger}}
= {\frac{1}{1-\beta}} {\Big[ \beta {\sum_{\substack{k=1 \\ k \neq {j}}}^{m}}{{p_k}{{x_{k}}_{\text{max}'}}} + {{\alpha}_j}{b_j} \Big]} \text{.}
\end{equation} 
By substituting the RHS of \eqref{eq:p_j_x_j_opt1} into \eqref{eq:p_j_x_j_2}, we get
\begin{equation}
\label{eq:alpha_j_2}
{\alpha_j} = {\frac{1}{b_j}}({p_j}{x_j^{\dagger}} - {p_{\pi}}{{x_{\pi}}_{\text{min}}}).
\end{equation}
Since ${\alpha_j} \le 1$, and according to \eqref{eq:R_j_star}, we obtain $R^{*}_j = {\alpha_j} \le 1$, and thus finish the proof of the converse.

\vspace{1em}
\textbf{We summarize Case 1 in the following:}
\begin{alignat*}{2}
     & R^{*}_j \le 1 \\
\iff & \frac{1-\beta}{\beta} {p_{\pi}}{{x_{\pi}}_{\text{min}}} - {b_j} \ge {\sum_{\substack{k=1 \\ k \neq {j}}}^{m}}{{p_i}{{x_i}_{\text{max}'}}}  \\
\iff & {x_{\pi}} = {{x_{\pi}}_{\text{min}}}\text{ and } {x_k} = {{x_k}_{\text{max}'}}, \forall k = 1, \cdots, m, k \neq j, \pi \\
\iff & {x_j^{\dagger}} = \frac{1}{p_j} \Big\{ \frac{1-\beta}{\beta}{p_{\pi}}{x_{\pi}}_{\text{min}} - \sum_{\substack{k=1 \\ k \neq j, {\pi}}}^{m} {p_k}{{x_k}_{\text{max}'}} \Big\} \triangleq {x_j^{\dagger1}} \\
\iff &\beta \ge \frac{ {p_{\pi}}{{x_{\pi}}_{\text{min}}} }{ { {p_{\pi}}{{x_{\pi}}_{\text{min}}} } + {\sum_{\substack{k=1 \\ k \neq {\pi}}}^{m}} {{p_k} {{x_k}_{\text{max}'}}} } \triangleq {\beta_1}.
\end{alignat*}

\vspace{3mm}
\noindent\textbf{Case 2}: 
\newline
Now consider the case $R^{*}_j \ge 1$. In this case, according to \eqref{eq:alpha_j_UB_short}, we have ${{\alpha}_j} = \min(R^{*}_j,1) = 1$, and \eqref{eq:x_j_2} thus becomes
\begin{equation}
\label{eq:x_j_case2}
{p_j}{x_j} 
= {\frac{1}{1-\beta}} {\Big[ \beta {\sum_{\substack{k=1 \\ k \neq {j}}}^{m}}{{p_k}{x_k}} + {b_j} \Big]} \text{.}
\end{equation} 
Moreover, based on \eqref{eq:alpha_j_UB}, there must exist an $k$ (denoted by $\kappa$) such that
\begin{equation}
\label{eq:alpha_j_UB_case2}
{{\alpha}_j} = 1 \le {\frac{1}{\beta{b_j}}}{\Big[ (1-\beta){p_{\kappa}}{x_{\kappa}} - \beta {\sum_{\substack{i=1 \\ i \neq {j}}}^{m}}{{p_i}{x_i}} \Big]} = R^{*}_j,
\end{equation}
which, since ${x_i} \le {{x_i}_{\text{max}'}}$, $\forall i$, yields
\begin{equation}
\label{eq:sum_pi_xi_LB_case2}
{\sum_{\substack{i=1 \\ i \neq {j}}}^{m}}{{p_i}{{x_i}_{\text{max}'}}} \ge
{\sum_{\substack{i=1 \\ i \neq {j}}}^{m}}{{p_i}{x_i}} \ge \frac{1-\beta}{\beta} {p_{\kappa}}{x_{\kappa}} - {b_j}.
\end{equation}
Since ${\sum_{\substack{i=1 \\ i \neq {j}}}^{m}}{{p_i}{x_i}} \ge {\sum_{\substack{i=1 \\ i \neq {j}}}^{m}}{{p_i}{{{x_i}_{\text{min}'}}}}$ as well, we need to consider different cases in the following in order to proceed the proof.

\vspace{3mm}
\noindent\textbf{Case 2.1}: 
\newline
First, we consider the case that the RHS of \eqref{eq:sum_pi_xi_LB_case2} is greater or equal to the sum of the equivalent lower limits, i.e.,
\begin{equation}
\label{eq:case_2_1_assumption}
\frac{1-\beta}{\beta} {p_{\kappa}}{x_{\kappa}} - {b_j} \ge {\sum_{\substack{i=1 \\ i \neq {j}}}^{m}}{{p_i}{{x_i}_{\text{min}'}}}.
\end{equation}
In such a case, the equality of the RHS of \eqref{eq:sum_pi_xi_LB_case2} can hold. Since, based on \eqref{eq:x_j_case2}, in order to minimize $x_j$, we need to minimize ${\sum_{\substack{k=1 \\ k \neq {j}}}^{m}} {{p_k}{x_k}}$, from the RHS of \eqref{eq:sum_pi_xi_LB_case2}, which is
\begin{equation}
\label{eq:sum_pi_xi_case2_1}
{\sum_{\substack{k=1 \\ k \neq {j}}}^{m}}{{p_k}{x_k}} = \frac{1-\beta}{\beta} {p_{\kappa}}{x_{\kappa}} - {b_j},
\end{equation}
and thus we need to minimize ${p_{\kappa}}{x_{\kappa}}$. By substituting the LHS of \eqref{eq:sum_pi_xi_case2_1} into \eqref{eq:x_j_case2} and \eqref{eq:u_v_alpha}, we obtain
\begin{equation} 
\label{eq:p_j_x_j_case2}
{p_j}{x_j} 
= {p_{\kappa}}{x_{\kappa}} + {b_j} 
\end{equation} 
and
\begin{equation}
\label{eq:alpha_k_noj_case2_1}
{\alpha_k} = {\frac{1}{b_k}}({p_k}{x_k} - {p_{\kappa}}{x_{\kappa}}), \forall k = 1, \cdots, m, k \neq j,
\end{equation}
respectively. Based on \eqref{eq:alpha_k_noj_case2_1}, similar to Case 1, since ${\alpha_k} \ge 0$ and $b_k \le 0$, we have ${p_{\kappa}}{x_{\kappa}} \ge {p_k}{x_k}$, $\forall k$, i.e., ${p_{\kappa}}{x_{\kappa}} = \underset{k}{\max} {p_k}{x_k}$. Moreover, in order to minimize $x_j$, we need to minimize ${p_{\kappa}}{x_{\kappa}}$. 
To minimize ${p_{\kappa}}{x_{\kappa}}$, since ${p_{\kappa}}{x_{\kappa}} = \underset{k}{\max} {p_k}{x_k}$, and ${p_k}{x_k} \ge {p_k}{{x_k}_{\text{min}}}$, $\forall k$ (including $\kappa$), the minimal ${p_{\kappa}}{x_{\kappa}}$, i.e., ${p_{\kappa}}{{x_{\kappa}}_{\text{min}}}$, is therefore the largest effective lower limit ${p_k}{{x_k}_{\text{min}}}$ over all $k$, i.e., ${p_{\kappa}}{{x_{\kappa}}_{\text{min}}} = \underset{k}{\max} {p_k}{{x_k}_{\text{min}}} = {p_{\pi}}{{x_{\pi}}_{\text{min}}}$ by definition, and thus we get $\kappa = \pi$ and ${x_{\pi}} = {{x_{\pi}}_{\text{min}}}$. Substituting which into \eqref{eq:p_j_x_j_case2}, we thus obtain the minimum of $x_j$
\begin{equation} 
\label{eq:p_j_x_j_case2_1_2}
\begin{split}
{x_j^{\dagger}}
&= \frac{1}{p_j} \Big\{ {p_{\pi}}{{x_{\pi}}_{\text{min}}} + {b_j}  \Big\}\\
&= \frac{1}{p_j} \Big\{ {p_{\pi}}{x_{\pi}}_{\text{min}} + {p_j - \beta \sum_{k=1}^{m} {p_k} } \Big\} \triangleq {x_j^{\dagger p}}.
\end{split}
\end{equation} 
For the rest of $k$, $k \neq j, \pi$, according to \eqref{eq:sum_pi_xi_case2_1}, we have
\begin{equation}
\label{eq:sum_pi_xi_case2_1_2}
\begin{split}
{\sum_{\substack{k=1 \\ k \neq {j, \pi}}}^{m}}{{p_k}{x_k}}
&= \frac{1-2\beta}{\beta} {p_{\pi}}{{x_{\pi}}_{\text{min}}} - b_j \\
&= \frac{1-2\beta}{\beta} {p_{\pi}}{{x_{\pi}}_{\text{min}}} - p_j + \beta \sum_{k=1}^{m} {p_k}.
\end{split}
\end{equation}
If (S1)$\simnot$(S4) are true, ${x_j^\dagger} \le {x_j}_{\text{max}}$. In addition, based on \eqref{eq:p_j_x_j_case2}, we have ${p_j}{x_j} = {p_{\pi}}{{x_{\pi}}_{\text{min}}} + {b_j} \le {p_{\pi}}{{x_{\pi}}_{\text{min}}}$, and thus from \eqref{eq:x_i_max}, we get ${x_j}_{\text{max}} = {x_j}_{\text{max}'}$. 
Therefore, based on \eqref{eq:p_j_x_j_case2_1_2}, the minimum $\beta$ such that \eqref{eq:OPT_1} has feasible solution, for Case 2.1, is
\begin{equation}
\label{eq:beta_case2_1}
\begin{split}
\beta 
&= \frac{ {p_{\pi}}{{x_{\pi}}_{\text{min}}} + {p_j}{(1 - {x_j^{\dagger}})}}{ \sum_{k=1}^{m} {p_k} } \\
&\ge \frac{ {p_{\pi}}{{x_{\pi}}_{\text{min}}} + {p_j}{(1 - {x_j}_{\text{max}})}}{ \sum_{k=1}^{m} {p_k} } \\
&= \frac{ {p_{\pi}}{{x_{\pi}}_{\text{min}}} + {p_j}{{y_j}_{\text{min}}}}{ \sum_{k=1}^{m} {p_k} } \triangleq {{\beta_p}_j}.
\end{split}
\end{equation}

We next prove the converse. If ${x_{\kappa}} = {x_{\pi}} = {{x_{\pi}}_{\text{min}}}$, and \eqref{eq:p_j_x_j_case2_1_2}, \eqref{eq:sum_pi_xi_case2_1_2}, and \eqref{eq:beta_case2_1} hold, from \eqref{eq:x_j_2}, we have
\begin{equation}
\label{eq:p_j_x_j_2_case_2_1}
{p_j}{x_j} 
= {\frac{1}{1-\beta}} {\Big[ \beta {\sum_{\substack{k=1 \\ k \neq {j}}}^{m}}{{p_k}{x_k}} + {{\alpha}_j}{b_j} \Big]} \text{,}
\end{equation} 
and, since ${\sum_{\substack{k=1 \\ k \neq {j}}}^{m}}{{p_k}{x_k}} \ge {\sum_{\substack{k=1 \\ k \neq {j}}}^{m}}{{p_k}{{{x_k}_{\text{min}'}}}}$, from \eqref{eq:sum_pi_xi_case2_1_2}, we have
\begin{equation}
\label{eq:sum_pi_xi_case2_1_3}
{\sum_{\substack{k=1 \\ k \neq {j}}}^{m}}{{p_k}{x_k}}
= \frac{1-\beta}{\beta} {p_{\pi}}{{x_{\pi}}_{\text{min}}} - b_j
\ge {\sum_{\substack{k=1 \\ k \neq {j}}}^{m}}{{p_k}{{{x_k}_{\text{min}'}}}}.
\end{equation}
Since \eqref{eq:sum_pi_xi_case2_1_2} and \eqref{eq:sum_pi_xi_case2_1_3} hold when ${x_j} = {x_j^{\dagger}}$, i.e., \eqref{eq:p_j_x_j_case2_1_2} holds, by substituting the LHS of \eqref{eq:sum_pi_xi_case2_1_3} into \eqref{eq:p_j_x_j_2_case_2_1}, we get
\begin{equation} 
\label{eq:p_j_x_j_case2_1_3}
{p_j}{x_j^{\dagger}}
= {p_{\pi}}{{x_{\pi}}_{\text{min}}} + \Big( \frac{{\alpha_j - \beta}}{1-\beta} \Big){b_j}.
\end{equation} 
Comparing \eqref{eq:p_j_x_j_case2_1_3} with \eqref{eq:p_j_x_j_case2_1_2}, we have $ \frac{{\alpha_j - \beta}}{1-\beta} = 1$, and therefore $\alpha_j = 1$. Based on \eqref{eq:alpha_j_UB_short}, we thus obtain $R^{*}_j \ge 1$ and finish the proof of the converse.

\vspace{0.2em}
\textbf{We summarize Case 2.1 in the following:}
\begin{alignat*}{2}
     & R^{*}_j \ge 1 \text{ and } \frac{1-\beta}{\beta} {p_{\pi}}{{x_{\pi}}_{\text{min}}} - {b_j} \ge {\sum_{\substack{k=1 \\ k \neq {j}}}^{m}}{{p_k}{{x_k}_{\text{min}'}}} \\
\iff & {\sum_{\substack{k=1 \\ k \neq {j}}}^{m}}{{p_k}{{x_k}_{\text{max}'}}} \ge \frac{1-\beta}{\beta} {p_{\pi}}{{x_{\pi}}_{\text{min}}} - {b_j} \ge {\sum_{\substack{k=1 \\ k \neq {j}}}^{m}}{{p_k}{{x_k}_{\text{min}'}}}  \\
\iff & {x_{\pi}} = {{x_{\pi}}_{\text{min}}}\text{ and } {\sum_{\substack{k=1 \\ k \neq {j, \pi}}}^{m}}{{p_k}{x_k}} = \frac{1-2\beta}{\beta} {p_{\pi}}{{x_{\pi}}_{\text{min}}} - p_j + \beta \sum_{k=1}^{m} {p_k} \\
\iff & {x_j^{\dagger}} = \frac{1}{p_j} \Big\{ {p_{\pi}}{x_{\pi}}_{\text{min}} + {p_j - \beta \sum_{k=1}^{m} {p_k} } \Big\} \triangleq {x_j^{\dagger p}} \\
\iff &\beta \ge \frac{ {p_{\pi}}{{x_{\pi}}_{\text{min}}} + {p_j}{{y_j}_{\text{min}}}}{ \sum_{k=1}^{m} {p_k} } \triangleq {{\beta_p}_j}.
\end{alignat*}

\vspace{3mm}
\noindent\textbf{Case 2.2}: 
\newline
Next, we consider the case that 
\begin{equation}
\label{eq:case_2_2_assumption}
\frac{1-\beta}{\beta} {p_{\pi}}{x_{\pi}}_{\text{min}} - {b_j} \le {\sum_{\substack{k=1 \\ k \neq {j}}}^{m}}{{p_k}{{x_k}_{\text{min}'}}}.
\end{equation}
Recall that $y_k \triangleq 1 - x_k$, ${{y_k}_{\text{max}}} \triangleq 1 - {{x_k}_{\text{min}}}$, ${{y_k}_{\text{min}}} \triangleq 1 - {{x_k}_{\text{max}}}$, and ${{\alpha}_k}' \triangleq 1 - {\alpha}_k$, $\forall k$. Since from \eqref{eq:alpha_j_UB_case2}, ${{\alpha}_j}' = 1 - {\alpha}_j = 0$, based on \eqref{eq:alpha_k_apostrophe}, we have
\begin{equation}
\label{eq:alpha_k_apostrophe_case_2_2}
{{\alpha}_k}'
= {\frac{\beta}{1-\beta}} {\frac{1}{b_k}} { \Big[ {\frac{1-\beta}{\beta}} {p_k}{y_k} - {\sum_{\substack{i=1 \\ i \neq {j}}}^{m}}{{p_i}{y_i}} \Big] }, \forall k = 1, \cdots, m, k \neq j \text{.}
\end{equation} 
Moreover, by substituting $x_k = 1 - y_k$, $\forall k$, into \eqref{eq:x_j_case2}, we get
\begin{equation}
\label{eq:x_j_case2_2}
{p_j}{x_j} 
= {p_j} - {\frac{\beta}{1-\beta}} {\sum_{\substack{k=1 \\ k \neq {j}}}^{m}}{{p_k}{y_k}} \text{,}
\end{equation} 
or equivalently,
\begin{equation}
\label{eq:y_j_case2_2}
{p_j}{y_j} 
= {\frac{\beta}{1-\beta}} {\sum_{\substack{k=1 \\ k \neq {j}}}^{m}}{{p_k}{y_k}} \text{,}
\end{equation} 
and
\begin{equation}
\label{eq:beta_ineq_case_2_2}
\beta = \frac{{p_j}{y_j}}{{\sum_{\substack{k=1}}^{m}}{{p_k}{y_k}} } = \frac{{p_j}{y_j}}{{{p_j}{y_j}} + {\sum_{\substack{k=1 \\ k \neq j}}^{m}}{{p_k}{y_k}}}.
\end{equation}

Substituting the RHS of \eqref{eq:y_j_case2_2} into \eqref{eq:alpha_k_apostrophe_case_2_2}, we obtain
\begin{equation}
\label{eq:alpha_k_apostrophe_case_noj}
{{\alpha}_k}' = {\frac{1}{b_k}}({p_k}{y_k} - {p_j}{y_j}), \forall k = 1, \cdots, m, k \neq j,
\end{equation}
Since ${\alpha_k}' \ge 0$ and $b_k \le 0$, we have ${p_j}{y_j} \ge {p_k}{y_k}$, $\forall k = 1, \cdots, m$, $k \neq j$, i.e., ${p_j}{y_j} = \underset{k}{\max} {p_k}{y_k}$. 
Moreover, from \eqref{eq:x_j_case2_2}, since $\beta$ is non-negative and $x_k \ge 0$ for all $k$, in order to minimize $x_j$, we need to maximize ${\sum_{\substack{k=1 \\ k \neq {j}}}^{m}}{{p_k}{y_k}}$. In addition, based on \eqref{eq:beta_ineq_case_2_2}, in Case 2.2, $\beta$ achieves its minimum when (i) minimizing ${p_j}{y_j}$ and (ii) maximizing ${\sum_{\substack{k=1 \\ k \neq j}}^{m}}{{p_k}{y_k}}$.

To minimize ${p_j}{y_j}$, since ${p_j}{y_j} = \underset{k}{\max} {p_k}{y_k}$, and ${p_k}{y_k} \ge {p_k}{{y_k}_{\text{min}}}$, $\forall k$ (including $j$), the minimal ${p_j}{y_j}$, i.e., ${p_j}{{y_j}_{\text{min}}}$, is therefore the largest effective lower limit ${p_k}{{y_k}_{\text{min}}}$ over all $k$, i.e., ${p_j}{{y_j}_{\text{min}}} = \underset{k}{\max} {p_k}{{y_k}_{\text{min}}} = {p_{\theta}}{{y_{\theta}}_{\text{min}}}$ by definition, and thus we get $j = \theta$ and ${y_{\theta}} = {{y_{\theta}}_{\text{min}}}$.

To maximize ${\sum_{\substack{k=1 \\ k \neq {j}}}^{m}}{{p_k}{y_k}}$, we need to find the maximum of each ${y_k}$. Since $0 \le {\alpha_k}' \le 1$, by substituting ${p_j}{y_j} = {p_{\theta}}{{y_{\theta}}_{\text{min}}}$ into \eqref{eq:alpha_k_apostrophe_case_noj}, we have ${p_{\theta}}{{y_{\theta}}_{\text{min}}} + {b_k} \le {p_k}{y_k} \le {p_{\theta}}{{y_{\theta}}_{\text{min}}}$, $\forall k = 1, \cdots, m$, $k \neq \theta$. 
By definitions, ${{y_k}_{\text{max}'}} \triangleq \min \{ {{y_k}_{\text{max}}}, { \frac{p_{\theta}}{p_k} {{y_{\theta}}_{\text{min}}} } \}$ and ${{y_k}_{\text{min}'}} \triangleq \max \{ {{y_k}_{\text{min}}}, { \frac{p_{\theta}}{p_k} {{y_{\theta}}_{\text{min}}} } + {b_k} \}$. 
Combining with the constraints ${{y_k}_{\text{min}}} \le {y_k} \le {{y_k}_{\text{max}}}$, we obtain ${{y_k}_{\text{min}'}} \le {y_k} \le {{y_k}_{\text{max}'}}$, $\forall k = 1, \cdots, m$, and thus ${\sum_{\substack{k=1 \\ k \neq {j}}}^{m}}{{p_k}{y_k}}$ is maximized when ${y_k} = {{y_k}_{\text{max}'}}$, $\forall k$, $k \neq j$.

By substituting the $y_k$'s we obtained above into \eqref{eq:x_j_case2_2}, based on which, the optimal objective value ${x_j^{\dagger}}$, the minimum of $x_j$, is thus
\begin{equation}
\label{eq:x_j_opt1_case2_2}
{x_j^{\dagger}} = \frac{1}{p_j} \Big\{ {p_j} - {\frac{\beta}{1-\beta}} {\sum_{\substack{k=1 \\ k \neq {j}}}^{m}}{{p_k}{{y_k}_{\text{max}'}}} \Big\} \triangleq {x_j^{\dagger0}}.
\end{equation}
If (S1)$\simnot$(S4) are true, ${x_j^\dagger} \le {x_j}_{\text{max}}$. Based on \eqref{eq:beta_ineq_case_2_2}, the minimum $\beta$ such that \eqref{eq:OPT_1} has feasible solution, for Case 2.2, is
\begin{equation}
\label{eq:beta_case2_2}
\beta \ge \frac{{p_{\theta}}{{y_{\theta}}_{\text{min}}}}{{{p_{\theta}}{{y_{\theta}}_{\text{min}}}} + {\sum_{\substack{k=1 \\ k \neq {\theta}}}^{m}}{{p_k}{{y_k}_{\text{max}'}}}} \triangleq \beta_0.
\end{equation}

We next prove the converse. Before proving the converse, we first need the following lemma.
\begin{lem}
\label{lem:sum_ineq}
If ${p_{\theta}}{y_{\theta}}_{\text{min}} \le {p_k}{y_k} - {b_k}$, $\forall k = 1, \ldots, m$, $k \neq \theta$, we have ${p_k}{{x_k}_{\text{min}'}} \le {p_k}{\big( 1-{{y_k}_{\text{max}'}} \big)}$, $\forall k = 1, \ldots, m$, $k \neq \theta$.
\end{lem}
\begin{proof}
Recall that $\forall k = 1, \ldots, m$, by definitions we have
\begin{alignat}{2}
{p_k}{{x_k}_{\text{min}'}} &\triangleq \max \{ {p_{\pi}}{x_{\pi}}_{\text{min}} + {b_k}, {p_k}{{x_k}_{\text{min}}} \}, \label{eq:case2_2_t1} \\
{p_k}{\big( 1-{{y_k}_{\text{max}'}} \big)} &= {p_k} - {p_k}{{{y_k}_{\text{max}'}}} \nonumber \\
&\triangleq \max \{ {p_k} - {p_{\theta}}{y_{\theta}}_{\text{min}}, {p_k} - {p_k}{{y_k}_{\text{max}}} \} \nonumber \\
&= \max \{ {p_k} - {p_{\theta}}{y_{\theta}}_{\text{min}}, {p_k}{{x_k}_{\text{min}}} \} \label{eq:case2_2_t2}.
\end{alignat}
Given the conditions that ${p_{\theta}}{y_{\theta}}_{\text{min}} \le {p_k}{y_k} - {b_k}$, $\forall k = 1, \ldots, m$, $k \neq \theta$, from which we get
\begin{equation}
\label{eq:pi_theta_ineq}
\begin{split}
&{p_{\theta}}{y_{\theta}}_{\text{min}} \le {p_k}{y_k} - {b_k}, \forall k = 1, \ldots, m, k \neq \theta, \\
\implies &{p_{\theta}}{y_{\theta}}_{\text{min}} \le {p_{\pi}}{y_{\pi}} - {b_{\pi}} \\
\implies &{p_{\theta}}{y_{\theta}}_{\text{min}} \le {p_{\pi}} \big( 1 - {x_{\pi}} \big) - {p_{\pi}} + {\beta \sum_{k=1}^{m} {p_k} } \\
\implies &{p_{\pi}}{x_{\pi}} = {p_{\pi}}{x_{\pi}}_{\text{min}} \le -{p_{\theta}}{y_{\theta}}_{\text{min}} + {\beta \sum_{k=1}^{m} {p_k} } \\
\implies &{p_{\pi}}{x_{\pi}}_{\text{min}} + {b_k} \le - {p_{\theta}}{y_{\theta}}_{\text{min}} + {\beta \sum_{k=1}^{m} {p_k} } + {p_k} - {\beta \sum_{k=1}^{m} {p_k} } \\
\implies &{p_{\pi}}{x_{\pi}}_{\text{min}} + {b_k} \le {p_k} - {p_{\theta}}{y_{\theta}}_{\text{min}}.
\end{split}
\end{equation}
Therefore, since the above inequalities hold for all $k$ except $k = \theta$, by comparing the RHS of \eqref{eq:case2_2_t1} and \eqref{eq:case2_2_t2}, we obtain that ${p_k}{{x_k}_{\text{min}'}} \le {p_k}{\big( 1-{{y_k}_{\text{max}'}} \big)}$, $\forall k = 1, \ldots, m$, $k \neq \theta$. We thus finish the proof.
\end{proof}

Now we start proving the converse. If ${y_j} \triangleq {1-{x_j}} = {y_{\theta}} = {{y_{\theta}}_{\text{min}}}$, ${y_k} \triangleq {1-{x_k}} = {{y_k}_{\text{max}'}}$, $\forall k$, $k \neq j$, \eqref{eq:x_j_opt1_case2_2} and the equality in \eqref{eq:beta_case2_2} hold, since from \eqref{eq:alpha_j_UB_case2} we have $\alpha_j = 1$, or equivalently, ${\alpha_j}' = 0$, by substituting the above ${y_k}$'s and ${\alpha_j}'$ into \eqref{eq:alpha_k_apostrophe}, we obtain
\begin{equation}
\label{eq:alpha_k_apostrophe_case2_2}
{{\alpha}_k}'
= {\frac{\beta}{1-\beta}} {\frac{1}{b_k}} { \Big[ {\frac{1-\beta}{\beta}} {p_k}{y_k} - {\sum_{\substack{i=1 \\ i \neq {j}}}^{m}}{{p_i}{{{y_i}_{\text{max}'}}}} \Big] }, \forall k, k \neq j \text{,}
\end{equation} 
and by substituting the ${x_k}$'s transformed from the above ${y_k}$'s (including $k = j = \theta$) into \eqref{eq:x_j_case2}, we have
\begin{equation}
\label{eq:term_A_value}
{\sum_{\substack{k=1 \\ k \neq {\theta}}}^{m}} {p_k}{x_k} = {\sum_{\substack{k=1 \\ k \neq {\theta}}}^{m}} {p_k}{\big( 1-{{y_k}_{\text{max}'}} \big)} = \frac{1-\beta}{\beta} {p_{\theta}}{x_{\theta}} - \frac{1}{\beta} {b_{\theta}}.
\end{equation}
From \eqref{eq:alpha_k_apostrophe_case2_2}, since $j = \theta$, by substituting \eqref{eq:beta_case2_2} into \eqref{eq:alpha_k_apostrophe_case2_2}, we obtain
\begin{equation}
\label{eq:alpha_k_apostrophe_case_noj_converse}
{{\alpha}_k}' = {\frac{1}{b_k}}({p_k}{y_k} - {{p_{\theta}}{{y_{\theta}}_{\text{min}}}}), \forall k = 1, \cdots, m, k \neq \theta.
\end{equation}
Since $0 \le {\alpha_k}' \le 1$, from \eqref{eq:alpha_k_apostrophe_case_noj_converse}, we obtain ${p_{\theta}}{{y_{\theta}}_{\text{min}}} + {b_k} \le {p_k}{y_k} \le {p_{\theta}}{{y_{\theta}}_{\text{min}}}$, $\forall k = 1, \cdots, m$, $k \neq \theta$. Therefore, based on Lemma \ref{lem:sum_ineq}, we have ${p_k}{{x_k}_{\text{min}'}} \le {p_k}{\big( 1-{{y_k}_{\text{max}'}} \big)}$, $\forall k = 1, \ldots, m$, $k \neq \theta$.

Recall that based on \eqref{eq:case2_2_t2}, for each $k$, $k \neq \theta$, ${p_k}{\big( 1-{{y_k}_{\text{max}'}} \big)}$ is the maximum of ${p_k} - {p_{\theta}}{y_{\theta}}_{\text{min}}$ and ${p_k}{{x_k}_{\text{min}}}$. Define $\Phi$ the set of $k$'s yielding ${p_k}{\big( 1-{{y_k}_{\text{max}'}} \big)} = {p_k}{{x_k}_{\text{min}}} \ge {p_k} - {p_{\theta}}{y_{\theta}}_{\text{min}}$, $k \in \Phi$, and define $\Omega$ the complement of $\Phi$, i.e., the set of $k$'s yielding ${p_k}{\big( 1-{{y_k}_{\text{max}'}} \big)} = {p_k} - {p_{\theta}}{y_{\theta}}_{\text{min}} > {p_k}{{x_k}_{\text{min}}}$, $k \in \Omega$. In addition, let $\phi \triangleq |\Phi|$ and $\omega \triangleq |\Omega|$. Note that $\phi + \omega = m-1$. We have the following lemma.

\begin{lem}
\label{lem:omega_upper_limit}
If ${y_{\theta}} = {{y_{\theta}}_{\text{min}}}$, ${y_k} = {{y_k}_{\text{max}'}}$, $\forall k$, $k \neq \theta$, and the equality in \eqref{eq:beta_case2_2} holds, we have $\omega \le \frac{1-\beta}{\beta}$.
\end{lem}
\begin{proof}
If ${y_{\theta}} = {{y_{\theta}}_{\text{min}}}$, ${y_k} = {{y_k}_{\text{max}'}}$, $\forall k$, $k \neq \theta$, and the equality in \eqref{eq:beta_case2_2} holds, from \eqref{eq:beta_case2_2} we have
\begin{equation}
\label{eq:beta_case2_2_converse}
\frac{\beta}{1-\beta} = \frac{{p_{\theta}}{{y_{\theta}}_{\text{min}}}}{{\sum_{\substack{k=1 \\ k \neq {\theta}}}^{m}}{{p_k}{{y_k}_{\text{max}'}}}}.
\end{equation}
Note that since for those $k$'s in $\Omega$, we have ${p_k}{\big( 1-{{y_k}_{\text{max}'}} \big)} = {p_k} - {p_{\theta}}{y_{\theta}}_{\text{min}}$, or equivalently, ${p_k}{{{y_k}_{\text{max}'}}} = {p_{\theta}}{y_{\theta}}_{\text{min}}$, and for those $k$'s in $\Phi$, we have ${p_k}{\big( 1-{{y_k}_{\text{max}'}} \big)} = {p_k}{{x_k}_{\text{min}}}$, or equivalently, ${p_k}{{{y_k}_{\text{max}'}}} = {p_k}{{{y_k}_{\text{max}}}}$, \eqref{eq:beta_case2_2_converse} thus becomes
\begin{equation}
\label{eq:beta_case2_2_converse_2}
\frac{{p_{\theta}}{{y_{\theta}}_{\text{min}}}}{{\sum_{\substack{k=1 \\ k \neq {\theta}}}^{m}}{{p_k}{{y_k}_{\text{max}'}}}} = \frac{{p_{\theta}}{{y_{\theta}}_{\text{min}}}}{\omega {{p_{\theta}}{{y_{\theta}}_{\text{min}}}} + {\sum_{k \in \Phi}{{p_k}{{y_k}_{\text{max}}}}}} \le \frac{1}{\omega}.
\end{equation}
We thus finish the proof.
\end{proof}

Define ${\bf{X}} \triangleq {\sum_{\substack{k=1 \\ k \neq {\theta}}}^{m}} {p_k}{\big( 1-{{y_k}_{\text{max}'}} \big)}$ and ${\bf{Y}} \triangleq {\sum_{\substack{k=1 \\ k \neq {\theta}}}^{m}} {p_k}{{x_k}_{\text{min}'}}$. Based on Lemma \ref{lem:sum_ineq}, we have ${\bf{X}} \ge {\bf{Y}}$. In addition, by definitions of $\Phi$ and $\Omega$, we have
\begin{equation}
\label{eq:term_A}
\begin{split}
{\bf{X}} 
&\triangleq {\sum_{\substack{k=1 \\ k \neq {\theta}}}^{m}} {p_k}{\big( 1-{{y_k}_{\text{max}'}} \big)} \\
&= {\sum_{k \in \Phi}{{p_k}{{x_k}_{\text{min}}}}} + {\sum_{k \in \Omega}{\big( {p_k} - {{p_{\theta}}{{y_{\theta}}_{\text{min}}}} \big)} } \\
&= \frac{1-\beta}{\beta} {p_{\theta}}{x_{\theta}} - \frac{1}{\beta} {b_{\theta}} \text{.  (Based on \eqref{eq:term_A_value})}
\end{split}
\end{equation}

Similarly to the $\Phi$ and $\Omega$ in ${\bf{X}} $, define $\Phi$' the set of $k$'s yielding ${p_k}{{x_k}_{\text{min}'}} = {p_k}{{x_k}_{\text{min}}} \ge {p_{\pi}}{x_{\pi}}_{\text{min}} + {b_k}$, $k \in \Phi$', for ${\bf{Y}}$. Since based on \eqref{eq:pi_theta_ineq} in Lemma \ref{lem:sum_ineq}, we have ${p_{\pi}}{x_{\pi}}_{\text{min}} + {b_k} \le {p_k} - {p_{\theta}}{y_{\theta}}_{\text{min}}$ for all $k$ except $\theta$. Therefore, for those $k$'s belonging to $\Phi$ (in ${\bf{X}})$, based on \eqref{eq:case2_2_t2}, we get ${p_k}{{x_k}_{\text{min}}} \ge {p_k} - {p_{\theta}}{y_{\theta}}_{\text{min}} \ge {p_{\pi}}{x_{\pi}}_{\text{min}} + {b_k}$, and based on \eqref{eq:case2_2_t1}, we find that \emph{those $k$'s belonging to $\Phi$ (in ${\bf{X}}$) also belong to $\Phi$' (in ${\bf{Y}})$, i.e., $\Phi \subseteq \Phi \text{'}$}. Therefore, ${\bf{Y}}$ can be interpreted by $\Phi$ as follow.
\begin{equation}
\label{eq:term_B}
\begin{split}
{\bf{Y}} 
&\triangleq {\sum_{\substack{k=1 \\ k \neq {\theta}}}^{m}} {p_k}{{x_k}_{\text{min}'}} \\
&= {\sum_{k \in \Phi}{{p_k}{{x_k}_{\text{min}}}}} + {\sum_{k \in \Omega}{ \max \Big\{ {p_{\pi}}{x_{\pi}}_{\text{min}} + {b_k}, \text{ } {p_k}{{x_k}_{\text{min}}} \Big\} } }.
\end{split}
\end{equation}

In addition, similarly to ${\bf{Y}}$, we define ${\bf{Z}}$ as follow.
\begin{equation}
\label{eq:term_B_p}
\begin{split}
{\bf{Z}} 
&\triangleq {\sum_{k \in \Phi}{{p_k}{{x_k}_{\text{min}}}}} + {\sum_{k \in \Omega}{\big( {p_{\pi}}{x_{\pi}}_{\text{min}} + {b_k} \big)} }.
\end{split}
\end{equation}
Clearly, the RHS of ${\bf{Z}}$ is not greater than the RHS of ${\bf{Y}}$, and thus we have ${\bf{X}} \ge {\bf{Y}} \ge {\bf{Z}}$. Define ${\bf{W}} \triangleq {\bf{X}} - {\bf{Z}}$, based on \eqref{eq:term_A} and \eqref{eq:term_B_p}, we have
\begin{equation}
\label{eq:term_C}
\begin{split}
{\bf{W}} 
&\triangleq {\bf{X}} - {\bf{Z}} \\
&= \sum_{k \in \Omega} \Big[ {\big( {p_k} - {{p_{\theta}}{{y_{\theta}}_{\text{min}}}} \big)} - {\big( {p_{\pi}}{x_{\pi}}_{\text{min}} + {b_k} \big)} \Big] \\
&= \sum_{k \in \Omega} \Big[ {\big( {p_k} - {{p_{\theta}}{{y_{\theta}}_{\text{min}}}} \big)} - {\big( {p_{\pi}}{x_{\pi}}_{\text{min}} + {{p_k - \beta \sum_{i=1}^{m} {p_i} }} \big)} \Big] \\
&= \sum_{k \in \Omega} \Big[ { {\beta \sum_{i=1}^{m} {p_i} } - {{p_{\theta}}{{y_{\theta}}_{\text{min}}}} - { {p_{\pi}}{x_{\pi}}_{\text{min}} } } \Big] \\
&= \omega \Big[ { {\beta \sum_{i=1}^{m} {p_i} } - {{p_{\theta}}{{y_{\theta}}_{\text{min}}}} - { {p_{\pi}}{x_{\pi}}_{\text{min}} } } \Big] \text{.}
\end{split}
\end{equation}
Define ${\bf{C}} \triangleq { {\beta \sum_{i=1}^{m} {p_i} } - {{p_{\theta}}{{y_{\theta}}_{\text{min}}}} - { {p_{\pi}}{x_{\pi}}_{\text{min}} } }$, the constant term in \eqref{eq:term_C}. Since ${\bf{X}} \ge {\bf{Z}}$, we have ${\bf{W}} = \omega {\bf{C}} \ge 0$, and because $\omega$ is the cardinality of $\Omega$, it is non-negative, and thus ${\bf{C}} \ge 0$. Since ${\bf{C}} \ge 0$, and from Lemma \ref{lem:omega_upper_limit}, $\omega \le \frac{1-\beta}{\beta}$, we thus have 
\begin{equation}
\label{eq:WXYZ}
{\bf{Y}} \ge {\bf{Z}} = {\bf{X}} - \big( {\bf{X}} - {\bf{Z}} \big) = {\bf{X}} - {\bf{W}} = {\bf{X}} - \omega {\bf{C}} \ge {\bf{X}} - \frac{1-\beta}{\beta} {\bf{C}}.
\end{equation}
Note that since ${{y_{\theta}}} = {{y_{\theta}}_{\text{min}}}$, the RHS of \eqref{eq:WXYZ} becomes
\begin{equation}
\begin{split}
&{\bf{X}} - \frac{1-\beta}{\beta} {\bf{C}} \\
= & \Big[ \frac{1-\beta}{\beta} {p_{\theta}}{x_{\theta}} - \frac{1}{\beta} {b_{\theta}} \Big] - \frac{1-\beta}{\beta} {\bf{C}} \\
= & \frac{1-\beta}{\beta} {p_{\theta}}{x_{\theta}} - \frac{1}{\beta} {b_{\theta}} - {(1-\beta) \sum_{i=1}^{m} {p_i} } + \frac{1-\beta}{\beta} {{p_{\theta}}{{y_{\theta}}_{\text{min}}}} + \frac{1-\beta}{\beta} { {p_{\pi}}{x_{\pi}}_{\text{min}} } \\
= & \frac{1-\beta}{\beta} {p_{\theta}} ({x_{\theta}} + {y_{\theta}}) - \frac{1}{\beta} {b_{\theta}} - {(1-\beta) \sum_{i=1}^{m} {p_i} } + \frac{1-\beta}{\beta} { {p_{\pi}}{x_{\pi}}_{\text{min}} } \\
= & \frac{1-\beta}{\beta} {p_{\theta}} - \frac{1-\beta}{\beta} {b_{\theta}} - {(1-\beta) \sum_{i=1}^{m} {p_i} } + \frac{1-\beta}{\beta} { {p_{\pi}}{x_{\pi}}_{\text{min}} } - {b_{\theta}} \\
= & \frac{1-\beta}{\beta} {b_{\theta}} - \frac{1-\beta}{\beta} {b_{\theta}} + \frac{1-\beta}{\beta} { {p_{\pi}}{x_{\pi}}_{\text{min}} } - {b_{\theta}} \\
= & \frac{1-\beta}{\beta} { {p_{\pi}}{x_{\pi}}_{\text{min}} } - {b_{\theta}}.
\end{split}
\end{equation}
Therefore, we obtain
\begin{equation}
\label{eq:case2_2_converse_final}
{\sum_{\substack{k=1 \\ k \neq {\theta}}}^{m}} {p_k}{{x_k}_{\text{min}'}} = {\bf{Y}} \ge {\bf{X}} - \frac{1-\beta}{\beta} {\bf{C}} = \frac{1-\beta}{\beta} { {p_{\pi}}{x_{\pi}}_{\text{min}} } - {b_{\theta}}.
\end{equation}
Since $j = \theta$, by replacing $\theta$ in \eqref{eq:case2_2_converse_final} by $j$, we obtain \eqref{eq:case_2_2_assumption} and finish the proof of the converse.

\vspace{0.2em}
\textbf{We summarize Case 2.2 in the following:}
\begin{alignat*}{2}
     & R^{*}_j \ge 1 \text{ and } \frac{1-\beta}{\beta} {p_{\pi}}{{x_{\pi}}_{\text{min}}} - {b_j} \le {\sum_{\substack{k=1 \\ k \neq {j}}}^{m}}{{p_k}{{x_k}_{\text{min}'}}} \\
\iff & {y_{\theta}} = {{y_{\theta}}_{\text{min}}}\text{ and } {y_k} = {{y_k}_{\text{max}'}}, \forall k = 1, \cdots, m, k \neq \theta \\
\iff & {x_j^{\dagger}} = \frac{1}{p_j} \Big\{ {p_j} - {\frac{\beta}{1-\beta}} {\sum_{\substack{k=1 \\ k \neq {j}}}^{m}}{{p_k}{{y_k}_{\text{max}'}}} \Big\} \triangleq {x_j^{\dagger0}} \\
\iff &\beta \ge \frac{{p_{\theta}}{{y_{\theta}}_{\text{min}}}}{{{p_{\theta}}{{y_{\theta}}_{\text{min}}}} + {\sum_{\substack{k=1 \\ k \neq {\theta}}}^{m}}{{p_k}{{y_k}_{\text{max}'}}}} \triangleq \beta_0.
\end{alignat*}

\vspace{1em}
Therefore, if (S1)~(S4) are true, \eqref{eq:OPT_1} has feasible solutions for arbitrary ${x_k}_{\text{min}}$ and ${x_k}_{\text{max}}$, $\forall k = 1, \ldots, m$, $k \neq j$, which implies \eqref{eq:OPT_1} has feasible solutions for all the cases (Case 1, Case 2.1, and Case 2.2), which requires $\beta$ to be greater than the minimum $\beta$ in each of the above cases, i.e., $\beta \ge \max \{ {{\beta}_0}, {{\beta}_1}, {{\beta_p}_j} \}$, where
\begin{alignat*}{2}
{\beta_0} &= \frac{ {p_{\theta}}{{y_{\theta}}_{\text{min}}} }{ { {p_{\theta}}{{y_{\theta}}_{\text{min}}} } + {\sum_{\substack{k=1 \\ k \neq {\theta}}}^{m}} {{p_k} {{y_k}_{\text{max}'}}} } \text{,} \\
{\beta_1} &= \frac{ {p_{\pi}}{{x_{\pi}}_{\text{min}}} }{ { {p_{\pi}}{{x_{\pi}}_{\text{min}}} } + {\sum_{\substack{k=1 \\ k \neq {\pi}}}^{m}} {{p_k} {{x_k}_{\text{max}'}}} } \text{,}  \\
{{\beta_p}_j} &= \frac{ { {p_{\pi}}{{x_{\pi}}_{\text{min}}} } + { {p_{j}}{{y_{j}}_{\text{min}}} } }{ \sum_{k=1}^{m} {p_k} } \text{,} 
\end{alignat*}
and the corresponding optimal objective value ${x_j^{\dagger}}$ and its corresponding optimal solutions are

\begin{alignat*}{2}
\beta = {{\beta}_0} 
\iff &{x_j^{\dagger}} = \frac{1}{p_j} \Big\{ {p_j} - \frac{\beta}{1-\beta} \sum_{\substack{k=1 \\ k \neq j}}^{m} {p_k}{{y_k}_{\text{max}'}} \Big\} \triangleq {x_j^{\dagger0}} \\
\iff &{y_j} = {y_{\theta}} = {{y_{\theta}}_{\text{min}}} \\
&{y_k} = {{y_k}_{\text{max}'}}, {\forall k = 1, \cdots, m}, \text{ } {k \neq {\theta}} \text{,} \\
\beta = {{\beta}_1} 
\iff &{x_j^{\dagger}} = \frac{1}{p_j} \Big\{ \frac{1-\beta}{\beta}{p_{\pi}}{x_{\pi}}_{\text{min}} - \sum_{\substack{k=1 \\ k \neq j, {\pi}}}^{m} {p_k}{{x_k}_{\text{max}'}} \Big\} \triangleq {x_j^{\dagger1}} \\
\iff &{x_{\pi}} = {{x_{\pi}}_{\text{min}}} \\
&{x_k} = {{x_k}_{\text{max}'}}, {\forall k = 1, \cdots, m}, \text{ } {k \neq {j, \pi}} \text{,} \\
\beta = {{\beta_p}_j} 
\iff &{x_j^{\dagger}} = \frac{1}{p_j} \Big\{ {p_{\pi}}{x_{\pi}}_{\text{min}} + {p_j - \beta \sum_{k=1}^{m} {p_k} } \Big\} \triangleq {x_j^{\dagger p}} \\
\iff &{x_{\pi}} = {{x_{\pi}}_{\text{min}}}  \\
&{\sum_{\substack{k=1 \\ k \neq {j, \pi}}}^{m}}{{p_k}{x_k}} = \frac{1-2\beta}{\beta} {p_{\pi}}{{x_{\pi}}_{\text{min}}} - p_j + \beta \sum_{k=1}^{m} {p_k} \text{.}
\end{alignat*}
We thus finish the proof of Lemma \ref{lem:x_dagger}.


\end{document}